\documentclass{article} 
\usepackage{iclr2025_conference,times}

\usepackage{amsmath,amsfonts,bm}

\def\eqref#1{equation~\ref{#1}}

\def\1{\bm{1}}

\DeclareMathAlphabet{\mathsfit}{\encodingdefault}{\sfdefault}{m}{sl}
\SetMathAlphabet{\mathsfit}{bold}{\encodingdefault}{\sfdefault}{bx}{n}

\usepackage{hyperref}
\usepackage{url}
\usepackage{booktabs}
\usepackage{latexsym}
\usepackage{graphicx}
\usepackage{subcaption}
\usepackage{multirow}
\usepackage{multicol}
\usepackage{array}
\usepackage{arydshln}
\usepackage{amsmath}
\usepackage{cleveref}
\usepackage{amssymb}
\usepackage{pifont}
\usepackage{algorithm}
\usepackage{algpseudocode}

\title{\ours: Human Response-Guided Evaluation of Instruction Following in Language Models}

\newcommand{\affilsup}[1]{\rlap{\textsuperscript{\normalfont#1}}}
\author{Xinxi Lyu\affilsup{1} \qquad
    Yizhong Wang\affilsup{1,2}  \qquad
    \textbf{Hannaneh Hajishirzi}\affilsup{1,2} \qquad Pradeep Dasigi\affilsup{1} \\
    $^1$Allen Institute for AI \qquad
    $^2$University of Washington
}

\newcommand{\ours}{HREF}
\newcommand{\ourmethod}{HREF}
\newcommand{\ourrb}{RoBERTa embedding}
\newcommand{\winrate}{expected win rate}
\newcommand{\cmark}{\textcolor{green}{\textbf{\ding{51}}}}
\newcommand{\xmark}{\textcolor{red}{\textbf{\ding{55}}}}%

\iclrfinalcopy 
\begin{document}

\maketitle

\begin{abstract}
Evaluating instruction following in Language Models has heavily relied on using LLMs as judges. In this work, we reevaluate the common choices for automatic evaluation setup, e.g.: choice of model and prompting strategy, on a wide range of instruction-following tasks. We experiment with methods that leverage human-written responses and observe that they enhance the reliability of automatic evaluations across a wide range of tasks, resulting in up to a 3.2\% improvement in agreement with human judges. We also show that human-written responses offer an orthogonal perspective to model-generated responses in following instructions and should be used as additional context when comparing model responses. Based on these observations, we develop a new evaluation benchmark, Human Response-Guided Evaluation of Instruction Following (\ours). It contains 4,258 human-written instructions spanning 11 task categories and we use the most reliable evaluation setup for each category among the choices we consider. To prevent test-set leakage, we keep a portion of our evaluation dataset hidden. We publicly release a separate development set, code\footnote{\url{https://github.com/allenai/href}} to evaluate models on it, and host a live leaderboard\footnote{\url{https://huggingface.co/spaces/allenai/href}} for publicly available models on our hidden evaluation set.

\end{abstract}

\newcommand{\oursb}{65.3}
\newcommand{\oursoq}{60.5}
\newcommand{\ourscq}{73.0}
\newcommand{\ourse}{75.6}
\newcommand{\oursg}{70.3}
\newcommand{\oursr}{75.4}
\newcommand{\ourss}{68.0}
\newcommand{\oursc}{68.1}
\newcommand{\oursf}{71.5}
\newcommand{\oursm}{66.0}
\newcommand{\oursro}{73.8}
\newcommand{\ourso}{69.4}

\newcommand{\ourllsb}{59.9}
\newcommand{\ourllsoq}{60.5}
\newcommand{\ourllscq}{70.3}
\newcommand{\ourllse}{69.9}
\newcommand{\ourllsg}{64.1}
\newcommand{\ourllsr}{70.4}
\newcommand{\ourllss}{61.0}
\newcommand{\ourllsc}{61.9}
\newcommand{\ourllsf}{71.5}
\newcommand{\ourllsm}{61.5}
\newcommand{\ourllsro}{70.0}
\newcommand{\ourllso}{65.0}

\newcommand{\ourgb}{60.1}
\newcommand{\ourgoq}{60.5}
\newcommand{\ourgcq}{70.3}
\newcommand{\ourge}{69.9}
\newcommand{\ourgg}{66.6}
\newcommand{\ourgr}{70.4}
\newcommand{\ourgs}{61.0}
\newcommand{\ourgc}{61.9}
\newcommand{\ourgf}{71.5}
\newcommand{\ourgm}{64.8}
\newcommand{\ourgro}{72.3}
\newcommand{\ourgo}{65.6}

\newcommand{\ourgtb}{65.5}
\newcommand{\ourgtoq}{60.5}
\newcommand{\ourgtcq}{70.3}
\newcommand{\ourgte}{75.6}
\newcommand{\ourgtg}{64.1}
\newcommand{\ourgtr}{72.5}
\newcommand{\ourgts}{64.4}
\newcommand{\ourgtc}{68.1}
\newcommand{\ourgtf}{71.5}
\newcommand{\ourgtm}{64.5}
\newcommand{\ourgtro}{70.5}
\newcommand{\ourgto}{67.4}

\newcommand{\hub}{62.6}
\newcommand{\huoq}{59.5}
\newcommand{\hucq}{71.9}
\newcommand{\hue}{73.8}
\newcommand{\hug}{66.3}
\newcommand{\hur}{69.9}
\newcommand{\hus}{63.4}
\newcommand{\huc}{69.8}
\newcommand{\huf}{71.0}
\newcommand{\hum}{62.8}
\newcommand{\huro}{74.3}
\newcommand{\huo}{67.0}

\newcommand{\lgb}{50.1}
\newcommand{\lgoq}{33.3}
\newcommand{\lgcq}{28.3}
\newcommand{\lge}{38.1}
\newcommand{\lgg}{49.1}
\newcommand{\lgr}{52.6}
\newcommand{\lgs}{40.5}
\newcommand{\lgc}{43.3}
\newcommand{\lgf}{39.0}
\newcommand{\lgm}{54.8}
\newcommand{\lgro}{45.8}
\newcommand{\lgo}{43.1}

\newcommand{\shb}{43.8}
\newcommand{\shoq}{61.6}
\newcommand{\shcq}{65.6}
\newcommand{\she}{57.6}
\newcommand{\shg}{49.0}
\newcommand{\shr}{45.5}
\newcommand{\shs}{56.6}
\newcommand{\shc}{52.9}
\newcommand{\shf}{57.8}
\newcommand{\shm}{42.8}
\newcommand{\shro}{53.5}
\newcommand{\sho}{53.2}

\newcommand{\rab}{46.3}
\newcommand{\raoq}{49.5}
\newcommand{\racq}{41.1}
\newcommand{\rae}{47.7}
\newcommand{\rag}{54.9}
\newcommand{\rar}{47.3}
\newcommand{\ras}{50.1}
\newcommand{\rac}{50.0}
\newcommand{\raf}{45.3}
\newcommand{\ram}{44.8}
\newcommand{\raro}{50.0}
\newcommand{\rao}{48.2}

\newcommand{\gtb}{60.0}
\newcommand{\gtoq}{49.1}
\newcommand{\gtcq}{54.8}
\newcommand{\gte}{67.8}
\newcommand{\gtg}{66.6}
\newcommand{\gtr}{68.6}
\newcommand{\gts}{53.3}
\newcommand{\gtc}{52.1}
\newcommand{\gtf}{56.3}
\newcommand{\gtm}{60.0}
\newcommand{\gtro}{69.0}
\newcommand{\gto}{59.4}

\newcommand{\gtwmb}{58.9}
\newcommand{\gtwmoq}{54.1}
\newcommand{\gtwmcq}{50.2}
\newcommand{\gtwme}{71.3}
\newcommand{\gtwmg}{64.4}
\newcommand{\gtwmr}{70.9}
\newcommand{\gtwms}{52.5}
\newcommand{\gtwmc}{57.5}
\newcommand{\gtwmf}{70.5}
\newcommand{\gtwmm}{62.5}
\newcommand{\gtwmro}{55.3}
\newcommand{\gtwmo}{60.2}

\newcommand{\gtwb}{60.1}
\newcommand{\gtwoq}{53.3}
\newcommand{\gtwcq}{63.1}
\newcommand{\gtwe}{69.3}
\newcommand{\gtwg}{65.6}
\newcommand{\gtwr}{70.1}
\newcommand{\gtws}{59.0}
\newcommand{\gtwc}{57.5}
\newcommand{\gtwf}{59.8}
\newcommand{\gtwm}{64.8}
\newcommand{\gtwro}{72.3}
\newcommand{\gtwo}{62.6}

\newcommand{\gttb}{65.5}
\newcommand{\gttoq}{57.1}
\newcommand{\gttcq}{63.6}
\newcommand{\gtte}{72.5}
\newcommand{\gttg}{63.9}
\newcommand{\gttr}{72.5}
\newcommand{\gtts}{61.9}
\newcommand{\gttc}{62.5}
\newcommand{\gttf}{61.5}
\newcommand{\gttm}{64.5}
\newcommand{\gttro}{69.8}
\newcommand{\gtto}{64.8}

\newcommand{\gttwmb}{67.1}
\newcommand{\gttwmoq}{62.1}
\newcommand{\gttwmcq}{65.0}
\newcommand{\gttwme}{71.7}
\newcommand{\gttwmg}{64.4}
\newcommand{\gttwmr}{74.0}
\newcommand{\gttwms}{62.9}
\newcommand{\gttwmc}{66.2}
\newcommand{\gttwmf}{69.3}
\newcommand{\gttwmm}{62.8}
\newcommand{\gttwmro}{57.5}
\newcommand{\gttwmo}{65.9}

\newcommand{\gttwb}{65.5}
\newcommand{\gttwoq}{58.8}
\newcommand{\gttwcq}{67.8}
\newcommand{\gttwe}{75.6}
\newcommand{\gttwg}{63.1}
\newcommand{\gttwr}{72.5}
\newcommand{\gttws}{64.4}
\newcommand{\gttwc}{68.1}
\newcommand{\gttwf}{64.3}
\newcommand{\gttwm}{62.5}
\newcommand{\gttwro}{70.5}
\newcommand{\gttwo}{66.4}

\newcommand{\lsb}{59.9}
\newcommand{\lsoq}{55.5}
\newcommand{\lscq}{49.5}
\newcommand{\lse}{64.8}
\newcommand{\lsg}{56.1}
\newcommand{\lsr}{64.3}
\newcommand{\lss}{57.5}
\newcommand{\lsc}{57.8}
\newcommand{\lsf}{63.8}
\newcommand{\lsm}{56.3}
\newcommand{\lsro}{64.5}
\newcommand{\lso}{58.7}

\newcommand{\lswmb}{60.0}
\newcommand{\lswmoq}{51.4}
\newcommand{\lswmcq}{45.9}
\newcommand{\lswme}{63.5}
\newcommand{\lswmg}{51.8}
\newcommand{\lswmr}{63.8}
\newcommand{\lswms}{54.9}
\newcommand{\lswmc}{54.1}
\newcommand{\lswmf}{61.5}
\newcommand{\lswmm}{51.3}
\newcommand{\lswmro}{70.0}
\newcommand{\lswmo}{56.5}

\newcommand{\lswb}{56.5}
\newcommand{\lswoq}{53.1}
\newcommand{\lswcq}{47.2}
\newcommand{\lswe}{65.8}
\newcommand{\lswg}{53.1}
\newcommand{\lswr}{63.0}
\newcommand{\lsws}{54.3}
\newcommand{\lswc}{55.5}
\newcommand{\lswf}{68.5}
\newcommand{\lswm}{52.5}
\newcommand{\lswro}{63.8}
\newcommand{\lswo}{56.8}

\newcommand{\lmb}{65.3}
\newcommand{\lmoq}{58.5}
\newcommand{\lmcq}{64.8}
\newcommand{\lme}{69.7}
\newcommand{\lmg}{64.1}
\newcommand{\lmr}{73.4}
\newcommand{\lms}{68.0}
\newcommand{\lmc}{66.0}
\newcommand{\lmf}{69.0}
\newcommand{\lmm}{63.5}
\newcommand{\lmro}{66.8}
\newcommand{\lmo}{66.2}

\newcommand{\lmwmb}{65.0}
\newcommand{\lmwmoq}{55.4}
\newcommand{\lmwmcq}{69.2}
\newcommand{\lmwme}{75.6}
\newcommand{\lmwmg}{67.8}
\newcommand{\lmwmr}{76.3}
\newcommand{\lmwms}{66.1}
\newcommand{\lmwmc}{68.9}
\newcommand{\lmwmf}{70.5}
\newcommand{\lmwmm}{68.3}
\newcommand{\lmwmro}{65.0}
\newcommand{\lmwmo}{67.7}

\newcommand{\lmwb}{62.5}
\newcommand{\lmwoq}{55.3}
\newcommand{\lmwcq}{73.0}
\newcommand{\lmwe}{75.6}
\newcommand{\lmwg}{70.3}
\newcommand{\lmwr}{75.4}
\newcommand{\lmws}{66.9}
\newcommand{\lmwc}{68.1}
\newcommand{\lmwf}{64.0}
\newcommand{\lmwm}{66.0}
\newcommand{\lmwro}{73.8}
\newcommand{\lmwo}{67.9}

\newcommand{\bemb}{64.1}
\newcommand{\bemoq}{66.5}
\newcommand{\bemcq}{64.8}
\newcommand{\beme}{76.4}
\newcommand{\bemg}{68.5}
\newcommand{\bemr}{67.6}
\newcommand{\bems}{60.1}
\newcommand{\bemc}{64.6}
\newcommand{\bemf}{71.3}
\newcommand{\bemm}{66.0}
\newcommand{\bemro}{65.8}
\newcommand{\bemo}{66.4}

\newcommand{\beb}{48.4}
\newcommand{\beoq}{60.5}
\newcommand{\becq}{68.8}
\newcommand{\bee}{69.9}
\newcommand{\beg}{64.1}
\newcommand{\ber}{64.8}
\newcommand{\bes}{61.0}
\newcommand{\bec}{61.9}
\newcommand{\bef}{71.5}
\newcommand{\bem}{55.5}
\newcommand{\bero}{70.0}
\newcommand{\beo}{62.5}

\newcommand{\romb}{61.3}
\newcommand{\romoq}{66.8}
\newcommand{\romcq}{67.7}
\newcommand{\rome}{73.4}
\newcommand{\romg}{63.3}
\newcommand{\romr}{69.0}
\newcommand{\roms}{57.5}
\newcommand{\romc}{59.0}
\newcommand{\romf}{67.3}
\newcommand{\romm}{67.5}
\newcommand{\romro}{64.0}
\newcommand{\romo}{64.7}

\newcommand{\rob}{52.4}
\newcommand{\rooq}{57.3}
\newcommand{\rocq}{70.3}
\newcommand{\roe}{65.0}
\newcommand{\rog}{58.0}
\newcommand{\ror}{70.4}
\newcommand{\ros}{55.0}
\newcommand{\roc}{50.0}
\newcommand{\rof}{70.3}
\newcommand{\rom}{61.5}
\newcommand{\roro}{61.0}
\newcommand{\roo}{60.3}

\newcommand{\ppb}{47.9}
\newcommand{\ppoq}{51.5}
\newcommand{\ppcq}{57.1}
\newcommand{\ppe}{51.8}
\newcommand{\ppg}{48.8}
\newcommand{\ppr}{42.8}
\newcommand{\pps}{52.9}
\newcommand{\ppc}{42.1}
\newcommand{\ppf}{54.9}
\newcommand{\ppm}{52.8}
\newcommand{\ppro}{59.7}
\newcommand{\ppo}{50.3}

\newcommand{\llamadbhabinstructb}{48.6}
\newcommand{\llamadbhabinstructoq}{84.3}
\newcommand{\llamadbhabinstructcq}{55.2}
\newcommand{\llamadbhabinstructe}{45.0}
\newcommand{\llamadbhabinstructg}{45.5}
\newcommand{\llamadbhabinstructr}{44.6}
\newcommand{\llamadbhabinstructs}{43.3}
\newcommand{\llamadbhabinstructc}{54.5}
\newcommand{\llamadbhabinstructro}{50.3}
\newcommand{\llamadbhabinstructm}{49.8}
\newcommand{\llamadbhabinstructf}{57.1}
\newcommand{\llamadbhabinstructo}{49.8}
\newcommand{\llamadbhabinstructrk}{1}
\newcommand{\mistrallargeinstructceahb}{54.5}
\newcommand{\mistrallargeinstructceahoq}{58.8}
\newcommand{\mistrallargeinstructceahcq}{35.1}
\newcommand{\mistrallargeinstructceahe}{39.6}
\newcommand{\mistrallargeinstructceahg}{51.4}
\newcommand{\mistrallargeinstructceahr}{50.5}
\newcommand{\mistrallargeinstructceahs}{45.3}
\newcommand{\mistrallargeinstructceahc}{44.5}
\newcommand{\mistrallargeinstructceahro}{48.5}
\newcommand{\mistrallargeinstructceahm}{59.1}
\newcommand{\mistrallargeinstructceahf}{26.0}
\newcommand{\mistrallargeinstructceaho}{47.6}
\newcommand{\mistrallargeinstructceahrk}{1}
\newcommand{\qwencfhcbinstructb}{51.4}
\newcommand{\qwencfhcbinstructoq}{71.6}
\newcommand{\qwencfhcbinstructcq}{30.2}
\newcommand{\qwencfhcbinstructe}{40.8}
\newcommand{\qwencfhcbinstructg}{47.7}
\newcommand{\qwencfhcbinstructr}{46.0}
\newcommand{\qwencfhcbinstructs}{36.4}
\newcommand{\qwencfhcbinstructc}{43.5}
\newcommand{\qwencfhcbinstructro}{45.6}
\newcommand{\qwencfhcbinstructm}{57.5}
\newcommand{\qwencfhcbinstructf}{23.4}
\newcommand{\qwencfhcbinstructo}{44.8}
\newcommand{\qwencfhcbinstructrk}{3}
\newcommand{\qwenbfbbabchatb}{47.7}
\newcommand{\qwenbfbbabchatoq}{80.4}
\newcommand{\qwenbfbbabchatcq}{31.4}
\newcommand{\qwenbfbbabchate}{30.4}
\newcommand{\qwenbfbbabchatg}{43.9}
\newcommand{\qwenbfbbabchatr}{39.0}
\newcommand{\qwenbfbbabchats}{37.1}
\newcommand{\qwenbfbbabchatc}{49.5}
\newcommand{\qwenbfbbabchatro}{38.8}
\newcommand{\qwenbfbbabchatm}{51.7}
\newcommand{\qwenbfbbabchatf}{39.8}
\newcommand{\qwenbfbbabchato}{43.0}
\newcommand{\qwenbfbbabchatrk}{3}
\newcommand{\llamadbtuludhabdpob}{51.3}
\newcommand{\llamadbtuludhabdpooq}{56.9}
\newcommand{\llamadbtuludhabdpocq}{35.9}
\newcommand{\llamadbtuludhabdpoe}{39.1}
\newcommand{\llamadbtuludhabdpog}{46.1}
\newcommand{\llamadbtuludhabdpor}{43.7}
\newcommand{\llamadbtuludhabdpos}{27.7}
\newcommand{\llamadbtuludhabdpoc}{41.5}
\newcommand{\llamadbtuludhabdporo}{43.2}
\newcommand{\llamadbtuludhabdpom}{51.5}
\newcommand{\llamadbtuludhabdpof}{25.4}
\newcommand{\llamadbtuludhabdpoo}{42.8}
\newcommand{\llamadbtuludhabdpork}{3}
\newcommand{\llamadbtuludhabb}{50.6}
\newcommand{\llamadbtuludhaboq}{58.3}
\newcommand{\llamadbtuludhabcq}{35.1}
\newcommand{\llamadbtuludhabe}{38.1}
\newcommand{\llamadbtuludhabg}{44.4}
\newcommand{\llamadbtuludhabr}{44.1}
\newcommand{\llamadbtuludhabs}{28.7}
\newcommand{\llamadbtuludhabc}{42.8}
\newcommand{\llamadbtuludhabro}{42.1}
\newcommand{\llamadbtuludhabm}{50.2}
\newcommand{\llamadbtuludhabf}{26.0}
\newcommand{\llamadbtuludhabo}{42.3}
\newcommand{\llamadbtuludhabrk}{3}
\newcommand{\mistralsmallinstructceajb}{47.0}
\newcommand{\mistralsmallinstructceajoq}{67.6}
\newcommand{\mistralsmallinstructceajcq}{32.2}
\newcommand{\mistralsmallinstructceaje}{36.4}
\newcommand{\mistralsmallinstructceajg}{42.4}
\newcommand{\mistralsmallinstructceajr}{44.9}
\newcommand{\mistralsmallinstructceajs}{40.1}
\newcommand{\mistralsmallinstructceajc}{39.3}
\newcommand{\mistralsmallinstructceajro}{39.7}
\newcommand{\mistralsmallinstructceajm}{54.7}
\newcommand{\mistralsmallinstructceajf}{25.8}
\newcommand{\mistralsmallinstructceajo}{42.2}
\newcommand{\mistralsmallinstructceajrk}{3}
\newcommand{\llamadbibinstructb}{46.7}
\newcommand{\llamadbibinstructoq}{79.4}
\newcommand{\llamadbibinstructcq}{40.6}
\newcommand{\llamadbibinstructe}{32.4}
\newcommand{\llamadbibinstructg}{36.3}
\newcommand{\llamadbibinstructr}{36.7}
\newcommand{\llamadbibinstructs}{32.9}
\newcommand{\llamadbibinstructc}{43.0}
\newcommand{\llamadbibinstructro}{29.9}
\newcommand{\llamadbibinstructm}{35.8}
\newcommand{\llamadbibinstructf}{39.3}
\newcommand{\llamadbibinstructo}{38.5}
\newcommand{\llamadbibinstructrk}{8}
\newcommand{\yibfdebchatb}{49.7}
\newcommand{\yibfdebchatoq}{71.6}
\newcommand{\yibfdebchatcq}{25.2}
\newcommand{\yibfdebchate}{27.0}
\newcommand{\yibfdebchatg}{36.2}
\newcommand{\yibfdebchatr}{30.1}
\newcommand{\yibfdebchats}{21.5}
\newcommand{\yibfdebchatc}{35.8}
\newcommand{\yibfdebchatro}{31.8}
\newcommand{\yibfdebchatm}{41.1}
\newcommand{\yibfdebchatf}{28.8}
\newcommand{\yibfdebchato}{35.4}
\newcommand{\yibfdebchatrk}{9}
\newcommand{\qwenchcbinstructb}{42.4}
\newcommand{\qwenchcbinstructoq}{71.6}
\newcommand{\qwenchcbinstructcq}{24.8}
\newcommand{\qwenchcbinstructe}{32.4}
\newcommand{\qwenchcbinstructg}{31.7}
\newcommand{\qwenchcbinstructr}{29.1}
\newcommand{\qwenchcbinstructs}{15.3}
\newcommand{\qwenchcbinstructc}{40.0}
\newcommand{\qwenchcbinstructro}{38.4}
\newcommand{\qwenchcbinstructm}{26.4}
\newcommand{\qwenchcbinstructf}{47.2}
\newcommand{\qwenchcbinstructo}{35.0}
\newcommand{\qwenchcbinstructrk}{9}
\newcommand{\olmocbbcebdbinstructb}{37.7}
\newcommand{\olmocbbcebdbinstructoq}{59.8}
\newcommand{\olmocbbcebdbinstructcq}{30.9}
\newcommand{\olmocbbcebdbinstructe}{20.3}
\newcommand{\olmocbbcebdbinstructg}{34.8}
\newcommand{\olmocbbcebdbinstructr}{36.0}
\newcommand{\olmocbbcebdbinstructs}{32.2}
\newcommand{\olmocbbcebdbinstructc}{44.0}
\newcommand{\olmocbbcebdbinstructro}{25.9}
\newcommand{\olmocbbcebdbinstructm}{44.6}
\newcommand{\olmocbbcebdbinstructf}{28.4}
\newcommand{\olmocbbcebdbinstructo}{34.7}
\newcommand{\olmocbbcebdbinstructrk}{9}
\newcommand{\phidmediumekinstructb}{27.7}
\newcommand{\phidmediumekinstructoq}{81.4}
\newcommand{\phidmediumekinstructcq}{34.9}
\newcommand{\phidmediumekinstructe}{24.3}
\newcommand{\phidmediumekinstructg}{26.2}
\newcommand{\phidmediumekinstructr}{28.1}
\newcommand{\phidmediumekinstructs}{24.0}
\newcommand{\phidmediumekinstructc}{45.3}
\newcommand{\phidmediumekinstructro}{37.2}
\newcommand{\phidmediumekinstructm}{17.2}
\newcommand{\phidmediumekinstructf}{60.2}
\newcommand{\phidmediumekinstructo}{33.3}
\newcommand{\phidmediumekinstructrk}{9}
\newcommand{\llamadbtuludibdpob}{40.9}
\newcommand{\llamadbtuludibdpooq}{63.7}
\newcommand{\llamadbtuludibdpocq}{26.0}
\newcommand{\llamadbtuludibdpoe}{19.6}
\newcommand{\llamadbtuludibdpog}{36.6}
\newcommand{\llamadbtuludibdpor}{33.7}
\newcommand{\llamadbtuludibdpos}{22.8}
\newcommand{\llamadbtuludibdpoc}{33.1}
\newcommand{\llamadbtuludibdporo}{27.9}
\newcommand{\llamadbtuludibdpom}{45.0}
\newcommand{\llamadbtuludibdpof}{19.9}
\newcommand{\llamadbtuludibdpoo}{33.2}
\newcommand{\llamadbtuludibdpork}{9}
\newcommand{\llamadbtuludibb}{41.6}
\newcommand{\llamadbtuludiboq}{58.8}
\newcommand{\llamadbtuludibcq}{23.5}
\newcommand{\llamadbtuludibe}{18.3}
\newcommand{\llamadbtuludibg}{35.6}
\newcommand{\llamadbtuludibr}{34.0}
\newcommand{\llamadbtuludibs}{21.0}
\newcommand{\llamadbtuludibc}{32.3}
\newcommand{\llamadbtuludibro}{28.0}
\newcommand{\llamadbtuludibm}{41.6}
\newcommand{\llamadbtuludibf}{21.9}
\newcommand{\llamadbtuludibo}{32.7}
\newcommand{\llamadbtuludibrk}{9}
\newcommand{\olmocbbcehbinstructb}{38.1}
\newcommand{\olmocbbcehbinstructoq}{47.5}
\newcommand{\olmocbbcehbinstructcq}{19.6}
\newcommand{\olmocbbcehbinstructe}{16.8}
\newcommand{\olmocbbcehbinstructg}{32.3}
\newcommand{\olmocbbcehbinstructr}{30.1}
\newcommand{\olmocbbcehbinstructs}{17.6}
\newcommand{\olmocbbcehbinstructc}{26.9}
\newcommand{\olmocbbcehbinstructro}{18.3}
\newcommand{\olmocbbcehbinstructm}{36.4}
\newcommand{\olmocbbcehbinstructf}{15.8}
\newcommand{\olmocbbcehbinstructo}{27.7}
\newcommand{\olmocbbcehbinstructrk}{15}
\newcommand{\tulucdpohabb}{17.5}
\newcommand{\tulucdpohaboq}{60.8}
\newcommand{\tulucdpohabcq}{26.5}
\newcommand{\tulucdpohabe}{21.0}
\newcommand{\tulucdpohabg}{22.9}
\newcommand{\tulucdpohabr}{21.5}
\newcommand{\tulucdpohabs}{11.9}
\newcommand{\tulucdpohabc}{30.8}
\newcommand{\tulucdpohabro}{23.2}
\newcommand{\tulucdpohabm}{18.1}
\newcommand{\tulucdpohabf}{55.6}
\newcommand{\tulucdpohabo}{25.8}
\newcommand{\tulucdpohabrk}{15}
\newcommand{\llamachabchathfb}{22.3}
\newcommand{\llamachabchathfoq}{61.3}
\newcommand{\llamachabchathfcq}{30.0}
\newcommand{\llamachabchathfe}{26.0}
\newcommand{\llamachabchathfg}{19.2}
\newcommand{\llamachabchathfr}{17.1}
\newcommand{\llamachabchathfs}{21.8}
\newcommand{\llamachabchathfc}{34.3}
\newcommand{\llamachabchathfro}{19.1}
\newcommand{\llamachabchathfm}{13.2}
\newcommand{\llamachabchathff}{48.5}
\newcommand{\llamachabchathfo}{24.5}
\newcommand{\llamachabchathfrk}{17}
\newcommand{\mistralhbinstructvadb}{24.1}
\newcommand{\mistralhbinstructvadoq}{60.8}
\newcommand{\mistralhbinstructvadcq}{21.3}
\newcommand{\mistralhbinstructvade}{14.1}
\newcommand{\mistralhbinstructvadg}{19.1}
\newcommand{\mistralhbinstructvadr}{18.5}
\newcommand{\mistralhbinstructvads}{16.8}
\newcommand{\mistralhbinstructvadc}{27.1}
\newcommand{\mistralhbinstructvadro}{17.3}
\newcommand{\mistralhbinstructvadm}{25.5}
\newcommand{\mistralhbinstructvadf}{46.5}
\newcommand{\mistralhbinstructvado}{24.3}
\newcommand{\mistralhbinstructvadrk}{17}
\newcommand{\llamadbtuludhabsftb}{2.8}
\newcommand{\llamadbtuludhabsftoq}{95.1}
\newcommand{\llamadbtuludhabsftcq}{37.6}
\newcommand{\llamadbtuludhabsfte}{28.2}
\newcommand{\llamadbtuludhabsftg}{13.0}
\newcommand{\llamadbtuludhabsftr}{11.7}
\newcommand{\llamadbtuludhabsfts}{4.0}
\newcommand{\llamadbtuludhabsftc}{36.6}
\newcommand{\llamadbtuludhabsftro}{35.3}
\newcommand{\llamadbtuludhabsftm}{10.5}
\newcommand{\llamadbtuludhabsftf}{60.4}
\newcommand{\llamadbtuludhabsfto}{23.6}
\newcommand{\llamadbtuludhabsftrk}{17}
\newcommand{\wizardlmbdbvbcb}{16.9}
\newcommand{\wizardlmbdbvbcoq}{63.2}
\newcommand{\wizardlmbdbvbccq}{22.8}
\newcommand{\wizardlmbdbvbce}{17.1}
\newcommand{\wizardlmbdbvbcg}{15.8}
\newcommand{\wizardlmbdbvbcr}{14.7}
\newcommand{\wizardlmbdbvbcs}{7.7}
\newcommand{\wizardlmbdbvbcc}{33.6}
\newcommand{\wizardlmbdbvbcro}{9.7}
\newcommand{\wizardlmbdbvbcm}{11.3}
\newcommand{\wizardlmbdbvbcf}{43.9}
\newcommand{\wizardlmbdbvbco}{19.6}
\newcommand{\wizardlmbdbvbcrk}{20}
\newcommand{\llamacbdbchathfb}{16.8}
\newcommand{\llamacbdbchathfoq}{52.5}
\newcommand{\llamacbdbchathfcq}{21.8}
\newcommand{\llamacbdbchathfe}{19.8}
\newcommand{\llamacbdbchathfg}{15.9}
\newcommand{\llamacbdbchathfr}{14.2}
\newcommand{\llamacbdbchathfs}{19.8}
\newcommand{\llamacbdbchathfc}{27.4}
\newcommand{\llamacbdbchathfro}{11.5}
\newcommand{\llamacbdbchathfm}{9.1}
\newcommand{\llamacbdbchathff}{42.0}
\newcommand{\llamacbdbchathfo}{19.4}
\newcommand{\llamacbdbchathfrk}{20}
\newcommand{\tuluvcfppobdbufmeanhabufrmb}{37.7}
\newcommand{\tuluvcfppobdbufmeanhabufrmoq}{21.6}
\newcommand{\tuluvcfppobdbufmeanhabufrmcq}{6.4}
\newcommand{\tuluvcfppobdbufmeanhabufrme}{10.9}
\newcommand{\tuluvcfppobdbufmeanhabufrmg}{16.4}
\newcommand{\tuluvcfppobdbufmeanhabufrmr}{15.5}
\newcommand{\tuluvcfppobdbufmeanhabufrms}{16.1}
\newcommand{\tuluvcfppobdbufmeanhabufrmc}{12.4}
\newcommand{\tuluvcfppobdbufmeanhabufrmro}{11.7}
\newcommand{\tuluvcfppobdbufmeanhabufrmm}{21.9}
\newcommand{\tuluvcfppobdbufmeanhabufrmf}{21.4}
\newcommand{\tuluvcfppobdbufmeanhabufrmo}{19.0}
\newcommand{\tuluvcfppobdbufmeanhabufrmrk}{20}
\newcommand{\tulucdpobdbb}{9.3}
\newcommand{\tulucdpobdboq}{68.6}
\newcommand{\tulucdpobdbcq}{17.8}
\newcommand{\tulucdpobdbe}{13.6}
\newcommand{\tulucdpobdbg}{13.3}
\newcommand{\tulucdpobdbr}{15.6}
\newcommand{\tulucdpobdbs}{9.2}
\newcommand{\tulucdpobdbc}{24.9}
\newcommand{\tulucdpobdbro}{10.9}
\newcommand{\tulucdpobdbm}{10.2}
\newcommand{\tulucdpobdbf}{50.2}
\newcommand{\tulucdpobdbo}{18.4}
\newcommand{\tulucdpobdbrk}{20}
\newcommand{\vicunabdbvbfb}{4.1}
\newcommand{\vicunabdbvbfoq}{81.4}
\newcommand{\vicunabdbvbfcq}{26.5}
\newcommand{\vicunabdbvbfe}{13.9}
\newcommand{\vicunabdbvbfg}{10.2}
\newcommand{\vicunabdbvbfr}{11.9}
\newcommand{\vicunabdbvbfs}{7.9}
\newcommand{\vicunabdbvbfc}{25.6}
\newcommand{\vicunabdbvbfro}{10.6}
\newcommand{\vicunabdbvbfm}{5.5}
\newcommand{\vicunabdbvbff}{55.8}
\newcommand{\vicunabdbvbfo}{17.4}
\newcommand{\vicunabdbvbfrk}{20}
\newcommand{\llamadbtuludibsftb}{1.5}
\newcommand{\llamadbtuludibsftoq}{87.7}
\newcommand{\llamadbtuludibsftcq}{31.9}
\newcommand{\llamadbtuludibsfte}{17.6}
\newcommand{\llamadbtuludibsftg}{6.8}
\newcommand{\llamadbtuludibsftr}{7.4}
\newcommand{\llamadbtuludibsfts}{0.7}
\newcommand{\llamadbtuludibsftc}{31.3}
\newcommand{\llamadbtuludibsftro}{18.7}
\newcommand{\llamadbtuludibsftm}{4.0}
\newcommand{\llamadbtuludibsftf}{53.9}
\newcommand{\llamadbtuludibsfto}{17.0}
\newcommand{\llamadbtuludibsftrk}{25}
\newcommand{\llamachbchathfb}{16.4}
\newcommand{\llamachbchathfoq}{41.7}
\newcommand{\llamachbchathfcq}{21.3}
\newcommand{\llamachbchathfe}{12.9}
\newcommand{\llamachbchathfg}{13.3}
\newcommand{\llamachbchathfr}{12.9}
\newcommand{\llamachbchathfs}{9.7}
\newcommand{\llamachbchathfc}{19.2}
\newcommand{\llamachbchathfro}{7.3}
\newcommand{\llamachbchathfm}{7.5}
\newcommand{\llamachbchathff}{33.5}
\newcommand{\llamachbchathfo}{15.6}
\newcommand{\llamachbchathfrk}{25}
\newcommand{\vicunahbvbfb}{2.9}
\newcommand{\vicunahbvbfoq}{75.5}
\newcommand{\vicunahbvbfcq}{20.3}
\newcommand{\vicunahbvbfe}{10.6}
\newcommand{\vicunahbvbfg}{8.2}
\newcommand{\vicunahbvbfr}{7.0}
\newcommand{\vicunahbvbfs}{2.7}
\newcommand{\vicunahbvbfc}{24.1}
\newcommand{\vicunahbvbfro}{6.7}
\newcommand{\vicunahbvbfm}{4.2}
\newcommand{\vicunahbvbff}{52.8}
\newcommand{\vicunahbvbfo}{14.3}
\newcommand{\vicunahbvbfrk}{27}
\newcommand{\tulucdpohbb}{4.9}
\newcommand{\tulucdpohboq}{55.9}
\newcommand{\tulucdpohbcq}{14.6}
\newcommand{\tulucdpohbe}{9.9}
\newcommand{\tulucdpohbg}{10.0}
\newcommand{\tulucdpohbr}{9.9}
\newcommand{\tulucdpohbs}{6.9}
\newcommand{\tulucdpohbc}{18.4}
\newcommand{\tulucdpohbro}{5.4}
\newcommand{\tulucdpohbm}{6.0}
\newcommand{\tulucdpohbf}{46.5}
\newcommand{\tulucdpohbo}{13.8}
\newcommand{\tulucdpohbrk}{27}
\newcommand{\olmohbsfthfb}{1.2}
\newcommand{\olmohbsfthfoq}{80.4}
\newcommand{\olmohbsfthfcq}{13.4}
\newcommand{\olmohbsfthfe}{6.9}
\newcommand{\olmohbsfthfg}{6.7}
\newcommand{\olmohbsfthfr}{4.4}
\newcommand{\olmohbsfthfs}{1.5}
\newcommand{\olmohbsfthfc}{22.4}
\newcommand{\olmohbsfthfro}{4.7}
\newcommand{\olmohbsfthfm}{2.4}
\newcommand{\olmohbsfthff}{50.0}
\newcommand{\olmohbsfthfo}{12.1}
\newcommand{\olmohbsfthfrk}{29}
\newcommand{\mpthbchatb}{0.5}
\newcommand{\mpthbchatoq}{73.0}
\newcommand{\mpthbchatcq}{12.1}
\newcommand{\mpthbchate}{3.7}
\newcommand{\mpthbchatg}{5.0}
\newcommand{\mpthbchatr}{3.9}
\newcommand{\mpthbchats}{1.5}
\newcommand{\mpthbchatc}{21.1}
\newcommand{\mpthbchatro}{4.1}
\newcommand{\mpthbchatm}{1.3}
\newcommand{\mpthbchatf}{47.4}
\newcommand{\mpthbchato}{10.8}
\newcommand{\mpthbchatrk}{29}
\newcommand{\koalabdbhfb}{1.0}
\newcommand{\koalabdbhfoq}{70.1}
\newcommand{\koalabdbhfcq}{15.8}
\newcommand{\koalabdbhfe}{8.4}
\newcommand{\koalabdbhfg}{4.3}
\newcommand{\koalabdbhfr}{5.0}
\newcommand{\koalabdbhfs}{1.0}
\newcommand{\koalabdbhfc}{16.4}
\newcommand{\koalabdbhfro}{5.2}
\newcommand{\koalabdbhfm}{2.2}
\newcommand{\koalabdbhff}{39.8}
\newcommand{\koalabdbhfo}{10.4}
\newcommand{\koalabdbhfrk}{29}
\newcommand{\gpteallbdbsnoozyb}{0.7}
\newcommand{\gpteallbdbsnoozyoq}{90.2}
\newcommand{\gpteallbdbsnoozycq}{3.7}
\newcommand{\gpteallbdbsnoozye}{6.2}
\newcommand{\gpteallbdbsnoozyg}{6.4}
\newcommand{\gpteallbdbsnoozyr}{5.7}
\newcommand{\gpteallbdbsnoozys}{1.0}
\newcommand{\gpteallbdbsnoozyc}{11.7}
\newcommand{\gpteallbdbsnoozyro}{5.1}
\newcommand{\gpteallbdbsnoozym}{1.3}
\newcommand{\gpteallbdbsnoozyf}{35.9}
\newcommand{\gpteallbdbsnoozyo}{9.9}
\newcommand{\gpteallbdbsnoozyrk}{32}
\newcommand{\olmohbahceinstructhfb}{8.9}
\newcommand{\olmohbahceinstructhfoq}{36.3}
\newcommand{\olmohbahceinstructhfcq}{5.7}
\newcommand{\olmohbahceinstructhfe}{5.2}
\newcommand{\olmohbahceinstructhfg}{5.9}
\newcommand{\olmohbahceinstructhfr}{5.1}
\newcommand{\olmohbahceinstructhfs}{5.7}
\newcommand{\olmohbahceinstructhfc}{8.0}
\newcommand{\olmohbahceinstructhfro}{5.0}
\newcommand{\olmohbahceinstructhfm}{7.3}
\newcommand{\olmohbahceinstructhff}{22.7}
\newcommand{\olmohbahceinstructhfo}{8.8}
\newcommand{\olmohbahceinstructhfrk}{32}
\newcommand{\dollyvcbcbb}{0.3}
\newcommand{\dollyvcbcboq}{75.5}
\newcommand{\dollyvcbcbcq}{12.4}
\newcommand{\dollyvcbcbe}{7.4}
\newcommand{\dollyvcbcbg}{2.8}
\newcommand{\dollyvcbcbr}{2.0}
\newcommand{\dollyvcbcbs}{0.0}
\newcommand{\dollyvcbcbc}{6.7}
\newcommand{\dollyvcbcbro}{3.3}
\newcommand{\dollyvcbcbm}{0.9}
\newcommand{\dollyvcbcbf}{39.6}
\newcommand{\dollyvcbcbo}{8.6}
\newcommand{\dollyvcbcbrk}{32}
\newcommand{\koalahbhfb}{0.4}
\newcommand{\koalahbhfoq}{65.7}
\newcommand{\koalahbhfcq}{8.2}
\newcommand{\koalahbhfe}{7.9}
\newcommand{\koalahbhfg}{3.3}
\newcommand{\koalahbhfr}{3.4}
\newcommand{\koalahbhfs}{0.0}
\newcommand{\koalahbhfc}{9.7}
\newcommand{\koalahbhfro}{3.1}
\newcommand{\koalahbhfm}{1.3}
\newcommand{\koalahbhff}{38.7}
\newcommand{\koalahbhfo}{8.6}
\newcommand{\koalahbhfrk}{32}
\newcommand{\dollyvchbb}{0.2}
\newcommand{\dollyvchboq}{70.6}
\newcommand{\dollyvchbcq}{9.4}
\newcommand{\dollyvchbe}{5.0}
\newcommand{\dollyvchbg}{2.0}
\newcommand{\dollyvchbr}{1.3}
\newcommand{\dollyvchbs}{0.0}
\newcommand{\dollyvchbc}{6.2}
\newcommand{\dollyvchbro}{3.6}
\newcommand{\dollyvchbm}{0.2}
\newcommand{\dollyvchbf}{42.0}
\newcommand{\dollyvchbo}{8.2}
\newcommand{\dollyvchbrk}{32}
\newcommand{\oasstsftbpythiabcbb}{0.2}
\newcommand{\oasstsftbpythiabcboq}{62.3}
\newcommand{\oasstsftbpythiabcbcq}{2.0}
\newcommand{\oasstsftbpythiabcbe}{1.5}
\newcommand{\oasstsftbpythiabcbg}{1.7}
\newcommand{\oasstsftbpythiabcbr}{0.6}
\newcommand{\oasstsftbpythiabcbs}{0.0}
\newcommand{\oasstsftbpythiabcbc}{3.2}
\newcommand{\oasstsftbpythiabcbro}{2.4}
\newcommand{\oasstsftbpythiabcbm}{0.9}
\newcommand{\oasstsftbpythiabcbf}{23.2}
\newcommand{\oasstsftbpythiabcbo}{5.1}
\newcommand{\oasstsftbpythiabcbrk}{37}

\newcommand{\mistrallargeinstructceahbd}{67.5}
\newcommand{\mistrallargeinstructceahoqd}{70.7}
\newcommand{\mistrallargeinstructceahcqd}{23.3}
\newcommand{\mistrallargeinstructceahed}{35.7}
\newcommand{\mistrallargeinstructceahgd}{60.3}
\newcommand{\mistrallargeinstructceahrd}{58.6}
\newcommand{\mistrallargeinstructceahsd}{44.0}
\newcommand{\mistrallargeinstructceahcd}{71.9}
\newcommand{\mistrallargeinstructceahod}{60.3}
\newcommand{\mistrallargeinstructceahrkd}{1}
\newcommand{\gpteturbocaceaeajbd}{59.2}
\newcommand{\gpteturbocaceaeajoqd}{74.1}
\newcommand{\gpteturbocaceaeajcqd}{26.7}
\newcommand{\gpteturbocaceaeajed}{21.4}
\newcommand{\gpteturbocaceaeajgd}{54.4}
\newcommand{\gpteturbocaceaeajrd}{64.3}
\newcommand{\gpteturbocaceaeajsd}{44.0}
\newcommand{\gpteturbocaceaeajcd}{43.8}
\newcommand{\gpteturbocaceaeajod}{56.0}
\newcommand{\gpteturbocaceaeajrkd}{1}
\newcommand{\gpteocaceafbdbd}{58.3}
\newcommand{\gpteocaceafbdoqd}{82.8}
\newcommand{\gpteocaceafbdcqd}{23.3}
\newcommand{\gpteocaceafbded}{7.1}
\newcommand{\gpteocaceafbdgd}{54.7}
\newcommand{\gpteocaceafbdrd}{55.7}
\newcommand{\gpteocaceafbdsd}{34.0}
\newcommand{\gpteocaceafbdcd}{37.5}
\newcommand{\gpteocaceafbdod}{55.3}
\newcommand{\gpteocaceafbdrkd}{1}
\newcommand{\gptebbagpreviewbd}{58.3}
\newcommand{\gptebbagpreviewoqd}{75.9}
\newcommand{\gptebbagpreviewcqd}{20.0}
\newcommand{\gptebbagpreviewed}{21.4}
\newcommand{\gptebbagpreviewgd}{53.7}
\newcommand{\gptebbagpreviewrd}{47.1}
\newcommand{\gptebbagpreviewsd}{44.0}
\newcommand{\gptebbagpreviewcd}{50.0}
\newcommand{\gptebbagpreviewod}{54.4}
\newcommand{\gptebbagpreviewrkd}{1}
\newcommand{\llamadbhabinstructbd}{52.5}
\newcommand{\llamadbhabinstructoqd}{65.5}
\newcommand{\llamadbhabinstructcqd}{63.3}
\newcommand{\llamadbhabinstructed}{14.3}
\newcommand{\llamadbhabinstructgd}{49.8}
\newcommand{\llamadbhabinstructrd}{55.7}
\newcommand{\llamadbhabinstructsd}{66.0}
\newcommand{\llamadbhabinstructcd}{53.1}
\newcommand{\llamadbhabinstructod}{53.7}
\newcommand{\llamadbhabinstructrkd}{5}
\newcommand{\qwencfhcbinstructbd}{63.3}
\newcommand{\qwencfhcbinstructoqd}{72.4}
\newcommand{\qwencfhcbinstructcqd}{30.0}
\newcommand{\qwencfhcbinstructed}{21.4}
\newcommand{\qwencfhcbinstructgd}{49.1}
\newcommand{\qwencfhcbinstructrd}{47.1}
\newcommand{\qwencfhcbinstructsd}{50.0}
\newcommand{\qwencfhcbinstructcd}{53.1}
\newcommand{\qwencfhcbinstructod}{53.1}
\newcommand{\qwencfhcbinstructrkd}{6}
\newcommand{\mistralsmallinstructceajbd}{59.2}
\newcommand{\mistralsmallinstructceajoqd}{72.4}
\newcommand{\mistralsmallinstructceajcqd}{26.7}
\newcommand{\mistralsmallinstructceajed}{21.4}
\newcommand{\mistralsmallinstructceajgd}{46.3}
\newcommand{\mistralsmallinstructceajrd}{50.0}
\newcommand{\mistralsmallinstructceajsd}{44.0}
\newcommand{\mistralsmallinstructceajcd}{50.0}
\newcommand{\mistralsmallinstructceajod}{50.8}
\newcommand{\mistralsmallinstructceajrkd}{6}
\newcommand{\qwenbfbbabchatbd}{53.3}
\newcommand{\qwenbfbbabchatoqd}{81.9}
\newcommand{\qwenbfbbabchatcqd}{20.0}
\newcommand{\qwenbfbbabchated}{35.7}
\newcommand{\qwenbfbbabchatgd}{42.5}
\newcommand{\qwenbfbbabchatrd}{44.3}
\newcommand{\qwenbfbbabchatsd}{34.0}
\newcommand{\qwenbfbbabchatcd}{56.2}
\newcommand{\qwenbfbbabchatod}{48.6}
\newcommand{\qwenbfbbabchatrkd}{8}
\newcommand{\llamadbibinstructbd}{53.3}
\newcommand{\llamadbibinstructoqd}{58.6}
\newcommand{\llamadbibinstructcqd}{40.0}
\newcommand{\llamadbibinstructed}{14.3}
\newcommand{\llamadbibinstructgd}{37.4}
\newcommand{\llamadbibinstructrd}{41.4}
\newcommand{\llamadbibinstructsd}{44.0}
\newcommand{\llamadbibinstructcd}{34.4}
\newcommand{\llamadbibinstructod}{42.8}
\newcommand{\llamadbibinstructrkd}{9}
\newcommand{\qwenchcbinstructbd}{50.0}
\newcommand{\qwenchcbinstructoqd}{77.6}
\newcommand{\qwenchcbinstructcqd}{20.0}
\newcommand{\qwenchcbinstructed}{50.0}
\newcommand{\qwenchcbinstructgd}{30.8}
\newcommand{\qwenchcbinstructrd}{24.3}
\newcommand{\qwenchcbinstructsd}{34.0}
\newcommand{\qwenchcbinstructcd}{31.2}
\newcommand{\qwenchcbinstructod}{39.4}
\newcommand{\qwenchcbinstructrkd}{9}
\newcommand{\yibfdebchatbd}{51.7}
\newcommand{\yibfdebchatoqd}{60.3}
\newcommand{\yibfdebchatcqd}{30.0}
\newcommand{\yibfdebchated}{57.1}
\newcommand{\yibfdebchatgd}{32.0}
\newcommand{\yibfdebchatrd}{32.9}
\newcommand{\yibfdebchatsd}{20.0}
\newcommand{\yibfdebchatcd}{46.9}
\newcommand{\yibfdebchatod}{38.8}
\newcommand{\yibfdebchatrkd}{9}
\newcommand{\phidmediumekinstructbd}{29.2}
\newcommand{\phidmediumekinstructoqd}{77.6}
\newcommand{\phidmediumekinstructcqd}{40.0}
\newcommand{\phidmediumekinstructed}{35.7}
\newcommand{\phidmediumekinstructgd}{28.0}
\newcommand{\phidmediumekinstructrd}{21.4}
\newcommand{\phidmediumekinstructsd}{24.0}
\newcommand{\phidmediumekinstructcd}{46.9}
\newcommand{\phidmediumekinstructod}{35.3}
\newcommand{\phidmediumekinstructrkd}{9}
\newcommand{\gptdfturbobd}{19.2}
\newcommand{\gptdfturbooqd}{81.0}
\newcommand{\gptdfturbocqd}{40.0}
\newcommand{\gptdfturboed}{7.1}
\newcommand{\gptdfturbogd}{22.0}
\newcommand{\gptdfturbord}{10.0}
\newcommand{\gptdfturbosd}{12.0}
\newcommand{\gptdfturbocd}{43.8}
\newcommand{\gptdfturbood}{29.2}
\newcommand{\gptdfturborkd}{13}
\newcommand{\tulucdpohabbd}{21.7}
\newcommand{\tulucdpohaboqd}{79.3}
\newcommand{\tulucdpohabcqd}{26.7}
\newcommand{\tulucdpohabed}{14.3}
\newcommand{\tulucdpohabgd}{19.2}
\newcommand{\tulucdpohabrd}{21.4}
\newcommand{\tulucdpohabsd}{16.0}
\newcommand{\tulucdpohabcd}{31.2}
\newcommand{\tulucdpohabod}{28.3}
\newcommand{\tulucdpohabrkd}{13}
\newcommand{\mistralhbinstructvadbd}{22.5}
\newcommand{\mistralhbinstructvadoqd}{67.2}
\newcommand{\mistralhbinstructvadcqd}{26.7}
\newcommand{\mistralhbinstructvaded}{7.1}
\newcommand{\mistralhbinstructvadgd}{21.0}
\newcommand{\mistralhbinstructvadrd}{25.7}
\newcommand{\mistralhbinstructvadsd}{14.0}
\newcommand{\mistralhbinstructvadcd}{37.5}
\newcommand{\mistralhbinstructvadod}{28.0}
\newcommand{\mistralhbinstructvadrkd}{13}
\newcommand{\llamachabchathfbd}{20.0}
\newcommand{\llamachabchathfoqd}{60.3}
\newcommand{\llamachabchathfcqd}{53.3}
\newcommand{\llamachabchathfed}{21.4}
\newcommand{\llamachabchathfgd}{17.5}
\newcommand{\llamachabchathfrd}{21.4}
\newcommand{\llamachabchathfsd}{14.0}
\newcommand{\llamachabchathfcd}{46.9}
\newcommand{\llamachabchathfod}{26.2}
\newcommand{\llamachabchathfrkd}{16}
\newcommand{\wizardlmbdbvbcbd}{16.7}
\newcommand{\wizardlmbdbvbcoqd}{69.0}
\newcommand{\wizardlmbdbvbccqd}{26.7}
\newcommand{\wizardlmbdbvbced}{14.3}
\newcommand{\wizardlmbdbvbcgd}{14.5}
\newcommand{\wizardlmbdbvbcrd}{7.1}
\newcommand{\wizardlmbdbvbcsd}{8.0}
\newcommand{\wizardlmbdbvbccd}{34.4}
\newcommand{\wizardlmbdbvbcod}{22.3}
\newcommand{\wizardlmbdbvbcrkd}{16}
\newcommand{\tuluvcfppobdbufmeanhabufrmbd}{45.0}
\newcommand{\tuluvcfppobdbufmeanhabufrmoqd}{34.5}
\newcommand{\tuluvcfppobdbufmeanhabufrmcqd}{0.0}
\newcommand{\tuluvcfppobdbufmeanhabufrmed}{7.1}
\newcommand{\tuluvcfppobdbufmeanhabufrmgd}{14.3}
\newcommand{\tuluvcfppobdbufmeanhabufrmrd}{17.1}
\newcommand{\tuluvcfppobdbufmeanhabufrmsd}{14.0}
\newcommand{\tuluvcfppobdbufmeanhabufrmcd}{15.6}
\newcommand{\tuluvcfppobdbufmeanhabufrmod}{20.9}
\newcommand{\tuluvcfppobdbufmeanhabufrmrkd}{16}
\newcommand{\vicunabdbvbfbd}{10.0}
\newcommand{\vicunabdbvbfoqd}{77.6}
\newcommand{\vicunabdbvbfcqd}{16.7}
\newcommand{\vicunabdbvbfed}{21.4}
\newcommand{\vicunabdbvbfgd}{8.9}
\newcommand{\vicunabdbvbfrd}{10.0}
\newcommand{\vicunabdbvbfsd}{20.0}
\newcommand{\vicunabdbvbfcd}{43.8}
\newcommand{\vicunabdbvbfod}{20.8}
\newcommand{\vicunabdbvbfrkd}{16}
\newcommand{\tulucdpobdbbd}{6.7}
\newcommand{\tulucdpobdboqd}{65.5}
\newcommand{\tulucdpobdbcqd}{10.0}
\newcommand{\tulucdpobdbed}{21.4}
\newcommand{\tulucdpobdbgd}{12.4}
\newcommand{\tulucdpobdbrd}{18.6}
\newcommand{\tulucdpobdbsd}{10.0}
\newcommand{\tulucdpobdbcd}{31.2}
\newcommand{\tulucdpobdbod}{19.9}
\newcommand{\tulucdpobdbrkd}{16}
\newcommand{\llamacbdbchathfbd}{19.2}
\newcommand{\llamacbdbchathfoqd}{48.3}
\newcommand{\llamacbdbchathfcqd}{10.0}
\newcommand{\llamacbdbchathfed}{7.1}
\newcommand{\llamacbdbchathfgd}{13.3}
\newcommand{\llamacbdbchathfrd}{7.1}
\newcommand{\llamacbdbchathfsd}{16.0}
\newcommand{\llamacbdbchathfcd}{31.2}
\newcommand{\llamacbdbchathfod}{19.0}
\newcommand{\llamacbdbchathfrkd}{21}
\newcommand{\llamachbchathfbd}{17.5}
\newcommand{\llamachbchathfoqd}{41.4}
\newcommand{\llamachbchathfcqd}{26.7}
\newcommand{\llamachbchathfed}{21.4}
\newcommand{\llamachbchathfgd}{12.4}
\newcommand{\llamachbchathfrd}{10.0}
\newcommand{\llamachbchathfsd}{10.0}
\newcommand{\llamachbchathfcd}{21.9}
\newcommand{\llamachbchathfod}{17.7}
\newcommand{\llamachbchathfrkd}{21}
\newcommand{\tulucdpohbbd}{3.3}
\newcommand{\tulucdpohboqd}{67.2}
\newcommand{\tulucdpohbcqd}{16.7}
\newcommand{\tulucdpohbed}{21.4}
\newcommand{\tulucdpohbgd}{8.4}
\newcommand{\tulucdpohbrd}{4.3}
\newcommand{\tulucdpohbsd}{10.0}
\newcommand{\tulucdpohbcd}{21.9}
\newcommand{\tulucdpohbod}{16.4}
\newcommand{\tulucdpohbrkd}{21}
\newcommand{\vicunahbvbfbd}{3.3}
\newcommand{\vicunahbvbfoqd}{75.9}
\newcommand{\vicunahbvbfcqd}{23.3}
\newcommand{\vicunahbvbfed}{14.3}
\newcommand{\vicunahbvbfgd}{5.6}
\newcommand{\vicunahbvbfrd}{4.3}
\newcommand{\vicunahbvbfsd}{4.0}
\newcommand{\vicunahbvbfcd}{25.0}
\newcommand{\vicunahbvbfod}{16.0}
\newcommand{\vicunahbvbfrkd}{21}
\newcommand{\gpteallbdbsnoozybd}{1.7}
\newcommand{\gpteallbdbsnoozyoqd}{77.6}
\newcommand{\gpteallbdbsnoozycqd}{23.3}
\newcommand{\gpteallbdbsnoozyed}{7.1}
\newcommand{\gpteallbdbsnoozygd}{3.3}
\newcommand{\gpteallbdbsnoozyrd}{2.9}
\newcommand{\gpteallbdbsnoozysd}{0.0}
\newcommand{\gpteallbdbsnoozycd}{12.5}
\newcommand{\gpteallbdbsnoozyod}{14.0}
\newcommand{\gpteallbdbsnoozyrkd}{21}
\newcommand{\olmohbsfthfbd}{0.0}
\newcommand{\olmohbsfthfoqd}{70.7}
\newcommand{\olmohbsfthfcqd}{13.3}
\newcommand{\olmohbsfthfed}{7.1}
\newcommand{\olmohbsfthfgd}{3.3}
\newcommand{\olmohbsfthfrd}{2.9}
\newcommand{\olmohbsfthfsd}{4.0}
\newcommand{\olmohbsfthfcd}{25.0}
\newcommand{\olmohbsfthfod}{13.1}
\newcommand{\olmohbsfthfrkd}{26}
\newcommand{\olmohbahceinstructhfbd}{15.0}
\newcommand{\olmohbahceinstructhfoqd}{46.6}
\newcommand{\olmohbahceinstructhfcqd}{0.0}
\newcommand{\olmohbahceinstructhfed}{0.0}
\newcommand{\olmohbahceinstructhfgd}{7.0}
\newcommand{\olmohbahceinstructhfrd}{5.7}
\newcommand{\olmohbahceinstructhfsd}{4.0}
\newcommand{\olmohbahceinstructhfcd}{9.4}
\newcommand{\olmohbahceinstructhfod}{12.9}
\newcommand{\olmohbahceinstructhfrkd}{26}
\newcommand{\dollyvchbbd}{1.7}
\newcommand{\dollyvchboqd}{67.2}
\newcommand{\dollyvchbcqd}{16.7}
\newcommand{\dollyvchbed}{7.1}
\newcommand{\dollyvchbgd}{1.9}
\newcommand{\dollyvchbrd}{1.4}
\newcommand{\dollyvchbsd}{0.0}
\newcommand{\dollyvchbcd}{6.2}
\newcommand{\dollyvchbod}{11.3}
\newcommand{\dollyvchbrkd}{26}
\newcommand{\koalabdbhfbd}{0.0}
\newcommand{\koalabdbhfoqd}{62.1}
\newcommand{\koalabdbhfcqd}{10.0}
\newcommand{\koalabdbhfed}{14.3}
\newcommand{\koalabdbhfgd}{2.3}
\newcommand{\koalabdbhfrd}{0.0}
\newcommand{\koalabdbhfsd}{4.0}
\newcommand{\koalabdbhfcd}{21.9}
\newcommand{\koalabdbhfod}{11.2}
\newcommand{\koalabdbhfrkd}{26}
\newcommand{\mpthbchatbd}{0.0}
\newcommand{\mpthbchatoqd}{56.9}
\newcommand{\mpthbchatcqd}{13.3}
\newcommand{\mpthbchated}{0.0}
\newcommand{\mpthbchatgd}{3.3}
\newcommand{\mpthbchatrd}{0.0}
\newcommand{\mpthbchatsd}{0.0}
\newcommand{\mpthbchatcd}{28.1}
\newcommand{\mpthbchatod}{10.8}
\newcommand{\mpthbchatrkd}{26}
\newcommand{\dollyvcbcbbd}{0.0}
\newcommand{\dollyvcbcboqd}{62.1}
\newcommand{\dollyvcbcbcqd}{13.3}
\newcommand{\dollyvcbcbed}{0.0}
\newcommand{\dollyvcbcbgd}{0.9}
\newcommand{\dollyvcbcbrd}{2.9}
\newcommand{\dollyvcbcbsd}{0.0}
\newcommand{\dollyvcbcbcd}{12.5}
\newcommand{\dollyvcbcbod}{10.0}
\newcommand{\dollyvcbcbrkd}{26}
\newcommand{\koalahbhfbd}{0.0}
\newcommand{\koalahbhfoqd}{55.2}
\newcommand{\koalahbhfcqd}{16.7}
\newcommand{\koalahbhfed}{7.1}
\newcommand{\koalahbhfgd}{0.0}
\newcommand{\koalahbhfrd}{0.0}
\newcommand{\koalahbhfsd}{0.0}
\newcommand{\koalahbhfcd}{18.8}
\newcommand{\koalahbhfod}{8.8}
\newcommand{\koalahbhfrkd}{32}
\newcommand{\oasstsftbpythiabcbbd}{0.0}
\newcommand{\oasstsftbpythiabcboqd}{63.8}
\newcommand{\oasstsftbpythiabcbcqd}{0.0}
\newcommand{\oasstsftbpythiabcbed}{0.0}
\newcommand{\oasstsftbpythiabcbgd}{0.0}
\newcommand{\oasstsftbpythiabcbrd}{0.0}
\newcommand{\oasstsftbpythiabcbsd}{0.0}
\newcommand{\oasstsftbpythiabcbcd}{0.0}
\newcommand{\oasstsftbpythiabcbod}{8.6}
\newcommand{\oasstsftbpythiabcbrkd}{32}

\section{Introduction}\label{sec:intro}

Automatic evaluations of instruction following abilities in Large Language Models (LLMs) has recently received significant attention~\citep{zheng2023judging,alpaca_eval,lin2024wildbench,li2024crowdsourced,chiang2024chatbot}. To make evaluation efficient and enable rapid iteration over modeling choices during development, prior work has approximated human judgments of the model response quality using by using powerful language models a judge (\textbf{LLM-as-a-Judge}). Although model judges have been shown to exhibit biases due to superficial features, such as the length of responses, prior work has indicated that such biases can be addressed~\citep{dubois2024length} to improve the reliability of these judgments. However, the analysis of such biases and the corresponding debiasing techniques developed in prior work are based on a distribution of tasks that is not representative of the full range of applications of instruction-tuned language models.

In this work, we reevaluate various choices for automatic evaluation on a wider range of instruction following tasks (Section~\ref{sec:experimental-setup}). We choose a task distribution closely aligned with those typically used to train instruction-tuned models~\citep{ouyang2022training}, and measure the agreement between human and model judges by comparing LLM-as-a-Judge and embedding-based similarity approaches. We experiment with using human-written reference responses in the process---by including them as additional context in the LLM-as-a-Judge or by measuring embedding similarity between model responses and human responses---and observe that they enhance the reliability of automatic evaluation across many tasks, resulting in up to a 3.2\% improvement in agreement with human judges (Section~\ref{subsec:model-generated}). Our analysis also provides insights into how human-written responses are helpful for evaluating instruction following. We discovered that human-written responses often offer an orthogonal perspective to model-generated responses and should be used as a complementary reference when comparing model responses.

Based on these observations, we develop a new evaluation benchmark with 4,258 human-written prompts and reference responses spanning 11 task-categories. We use a composite evaluation setup that uses the most reliable evaluation method for each task-category. Given the reliance on human-written responses, we name this benchmark Human Reference-guided Evaluation of instruction Following (\textbf{\ours}). Our new benchmark additionally addresses two important limitations in existing instruction-following evaluations: \textit{test-set leakage} and \textit{a limited focus on individual tasks}.

\vspace{-0.2cm}
\paragraph{Test-set leakage.} A consequence of the open availability of the existing instruction following evaluation sets is that these datasets can (often inadvertently) end up in the post-training datasets. For instance, \citet{lambert2024t} show that datasets containing real user conversations with language models, like LMSys-Chat 1M contain significant portions of AlpacaEval data in them. Training on such contaminated datasets can lead to inflated model performance on these benchmarks. To deal with this issue, we create separate development and test splits of \ours, and keep the test split private.

\vspace{-0.2cm}
\paragraph{Limited focus on individual tasks.} Prior instruction-following evaluations either focus on a small set of tasks~\citep{alpaca_eval,zheng2023judging} or use a relatively small sample of real user interactions with language models~\citep{lin2024wildbench} where some tasks are under-represented\footnote{WildBench has task categories identified post-hoc, and the smallest category has only 16 instances.}. As a result, both these approaches result in evaluation datasets that provide limited actionable insights about the model development process at the individual task level, e.g., which skills to upsample in the training datasets. In contrast, we take a \textit{task-centric} view of data curation with \ours. We start with a taxonomy of 11 task categories used in \citet{ouyang2022training} and collect more than 100 human-written instruction-response pairs for each task category. We apply a task-specific evaluation method and report the result for each task category separately in order to provide a reliable evaluation and deliver insights about the tasks the developers should focus on.

We study the impact of our design choices in \ours, including the size of the evaluation set, the choice of the judge model and baseline model, and our prompt template in Section~\ref{sec:discussion}. We build a leaderboard that uses the private test split of \ours.


\begin{figure}
\centering \footnotesize
\resizebox{0.95\columnwidth}{!}{\includegraphics[trim={0cm 0cm 0cm 0cm},clip]{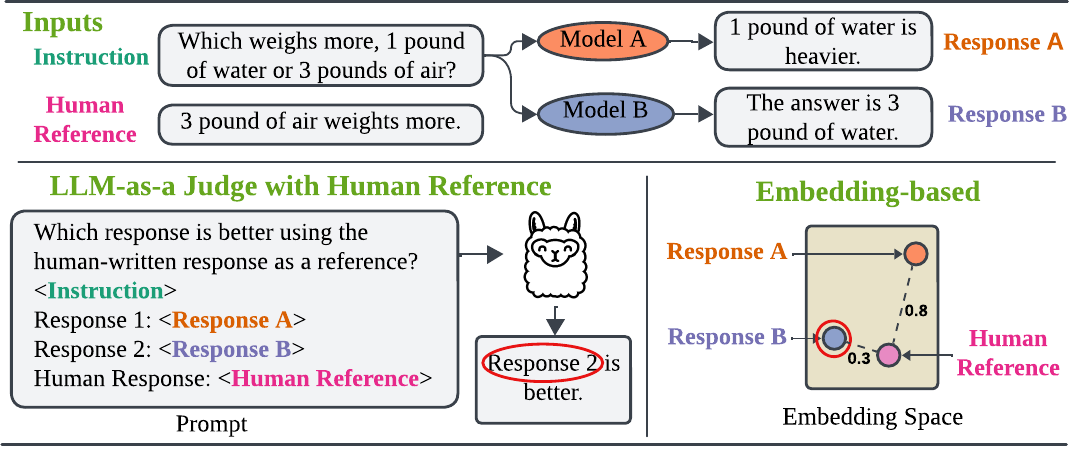}}
\caption{
An overview of our composite method leverage the human-written response to judge between two responses given an instruction. The example and the prompt shown in the figure are not exact. See details of these methods in Section~\ref{subsec:methods}.
}\label{fig:teaser}
\end{figure}


\section{Empirical Basis for the Evaluation Setup}\label{sec:experimental-setup}
\begin{table*}[t!]
    \centering  \small
    \setlength{\tabcolsep}{4.5pt}
    \begin{tabular}{p{0.15\linewidth} p{0.75\linewidth}}
    \toprule
        \textbf{Task} & \textbf{Example} \\
    \midrule
        Brainstorming &  What should I look for when buying a car?  \\
    \midrule
        Open QA & Do tomatoes contain MSG?\\
    \midrule
        Closed QA & What was the first type of anesthesia used in surgery? On October 16, 1846, the first successful public demonstration of the use of ether for surgical anesthesia was performed, making pain-free surgery possible. \\
    \midrule
        Extraction & I want to know the sizes ticks come in in a numbered list. Here is the texting I am talking about: Ticks come in three sizes depending on their life stage. They can be the size of a grain of sand, a poppy seed, or an apple seed.\\
    \midrule
        Generation & Write a poem about tacos. \\
    \midrule
        Rewriting & Rewrite the sentence in active voice. My vegetable garden was eaten by a donkey. \\
    \midrule
        Summarization & Summarize this in one sentence: SOCIETY FOR HUMAN RIGHTS 1924 After being temporarily committed to a mental institution because of his sexual orientation, Henry Gerber, a German immigrant and World War I Army enlistee, establishes the Society for Human Rights, the first American homosexual rights organization. \\
    \midrule
        Classification & If a poem is titled "Hide and Seek" does that sound childish or mature?\\
    \midrule
        Fact Checking & What were the top 5 cited papers in AI from 2022? \\
    \midrule
        Multi-Document Synthesis & According to these reviews from Yelp and Google Maps, determine whether or not Moonshadows is an ideal location for a date night during my upcoming trip to Malibu. Yelp Reviews: \textbackslash n 1. ... \textbackslash n 2. ... \textbackslash n 3. ... \textbackslash n 4. ... \\
    \midrule
        Reasoning Over Numerical Data & What is the increase in sales from Jan to Feb? \textbackslash n$|$ Month      $|$ Sales      $|$ Expenses $|$ \textbackslash n Jan             $|$   2,894    $|$   2,582     $|$ \textbackslash n Feb             $|$ 3,820      $|$ 3,517     $|$ \textbackslash n March        $|$ 2,009      $|$ 1,796      $|$ \\
    \bottomrule
    \end{tabular}\vspace{-.1em}
    \caption{
    \textbf Examples of instructions in each of the 11 task categories.
 }\label{tab:task-defination}
\end{table*}

In this section, we describe our experimental settings to evaluate various choices in the LLM-as-a-judge setup used for instruction following evaluations. We also explore how human-written responses can be utilized to improve the reliability of such evaluations.
Specifically, we construct a dataset for evaluating the evaluation methods, and collect human annotations on the pairwise preference between response pairs (Section~\ref{subsec:human-dataset-construction}). We introduce three new automatic evaluation methods that leverage human-written responses (Section~\ref{subsec:methods}), and show the results of the experiment comparing them in Section~\ref{sec:result}.

\subsection{Human Agreement Set Construction}\label{subsec:human-dataset-construction}
We compare evaluation methods based on how well they agree with human judgments. To enable such a comparison, we built a dataset with human annotated preferences comparing pairs of responses sampled from a diverse set of models. We refer to this dataset as the \textbf{human agreement set} and it is a subset of the final dataset described in Section~\ref{sec:preference-annotator}.

\subsubsection{Instructions and Responses Collection}\label{subsec:human-agreement-instruction-collection}
We construct a dataset of instructions, each associated with a human-written response, two candidate model responses, and multiple human judgment annotations indicating which model response is preferred\footnote{Data Link:  \url{https://huggingface.co/datasets/allenai/href_preference}.}.

\vspace{-0.2cm}\paragraph{Task Selection.}  Prior benchmarks for evaluating instruction following include sets of instructions that are representative of real user interactions with publicly hosted language models. While evaluating on such datasets can inform how the model would perform in practice, the input distributions tend to be heavily skewed towards a small set of tasks as shown by~\citep{lin2024wildbench, chiang2024chatbot, li2024crowdsourced}. Consequently, the decisions regarding the evaluation setup, though based on rigorous human agreement experiments, may be biased towards a small number of tasks. 
In contrast, we begin with a taxonomy of 11 instruction-following tasks and build a dataset of instructions specifically targeting these tasks. Specifically, we select 8 tasks from the InstructGPT taxonomy~\citep{ouyang2022training}---Brainstorming, Open QA, Closed QA, Extraction, Generation, Rewriting, Summarization, Classification, and 3 additional tasks focused on scientific text understanding---Fact Checking, Multi-Document Synthesis, and Reasoning Over Numerical Data. See Table~\ref{tab:task-defination} for examples on instructions in each category.

\vspace{-0.2cm} \paragraph{Instruction Set.} We sample instructions and human-written responses for 8 of the tasks from the No Robots dataset~\citep{no_robots}. We sample data primarily from the test set, and for tasks that are not well represented in the test set, we additionally sample from the training set.  For the remaining 3 scientific text understanding tasks, we hire human experts to write instructions and associated responses. We ended up with 438 pairs where all 11 categories are reasonably represented (See Figure~\ref{fig:dataset-distribution}).

\vspace{-0.2cm} \paragraph{Model Pool.} In order to ensure the diversity of the responses, we build a model pool with 32 LLMs with sizes ranging from 7B to over 100B from more than 10 different model families.  
See the full list of models in Appendix~\ref{app:human-set-model-pool}.

\begin{figure}
\centering \footnotesize
\resizebox{0.7\columnwidth}{!}{\includegraphics[trim={0.5cm 0.5cm 0.5cm 0.5cm},clip]{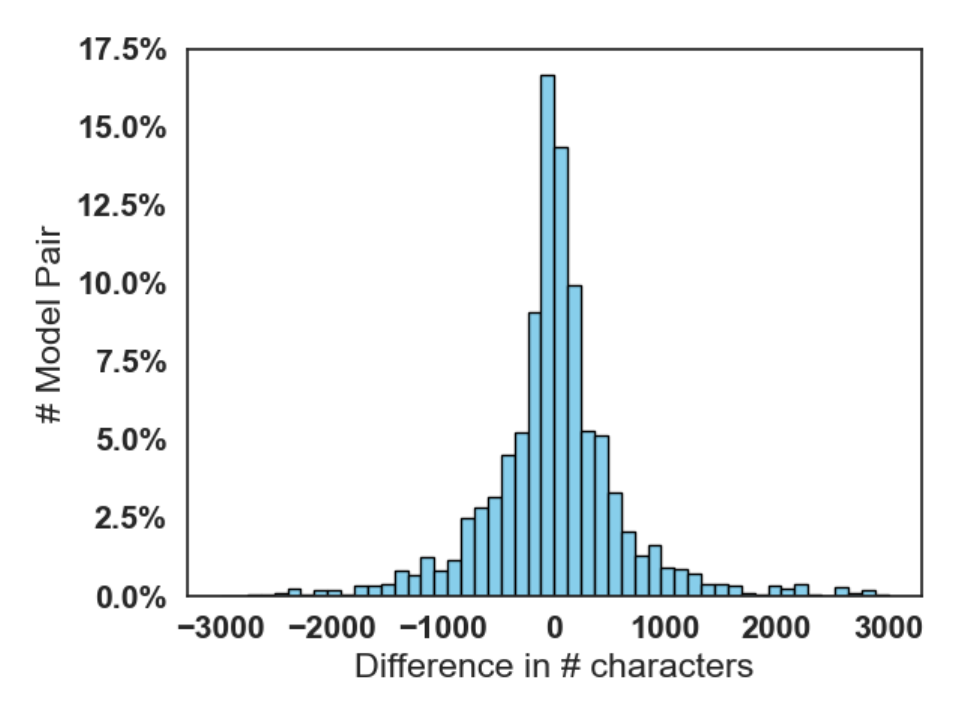}}
\caption{
\textbf{The distribution of length difference between sampled model responses and the base line model responses}. The distribution is symmetrical.
}\label{fig:length-distribution}
\end{figure}

\vspace{-0.2cm} \paragraph{Response Sampling.} For each instruction, we sample responses from four distinct models. We create instances of pairwise comparison, i.e., comparing two model responses for the same instruction, by pairing each of the four model responses with that from a fixed \textit{baseline model}, Llama-3.1-405B-Instruct-FP8. To avoid response length-related bias (e.g., a positive correlation between response length and quality), we divide all model responses for each instruction into two groups based on whether they are longer or shorter than the baseline model responses. We then randomly sample two response from each of the two groups. To ensure high-quality of the response and to avoid repetitions in generation, we use a decoding temperature of 1.0 for all the models. 
Figure~\ref{fig:length-distribution} shows the resulting distribution of the length difference between sampled model responses and baseline model responses. The symmetrical distribution shows that both the shorter and the longer responses are roughly equally sampled.

\subsubsection{Collecting Human Annotations}\label{subsec:annotation-collection}
We collected four human preference annotations for each instance (instruction + model response pair) in our human agreement set following the procedure described below. Importantly, the annotators are shown only the instructions and the two model responses per each instance, and not the human-written responses.

\vspace{-0.2cm} \paragraph{Annotator Selection.} We recruited native English speakers from the U.S., the U.K., and Canada, who have  a Bachelor's degrees or above, and a prior approval rating over 99\% from Prolific~\citep{Prolific_2014}. We further screened annotators using a qualification test that required them to correctly annotate at least 9 out of 10 instances with easily distinguishable model response pairs. We assign the qualification task to 50 participants, and recruited 16 of them as our final group of annotators and paid them \$16 / hour.

\vspace{-0.2cm} \paragraph{Annotation Guidelines and Interface.}
We used the annotation guidelines from \citet{alpaca_eval} with the following modifications: We slightly modified checklist of judging aspects, included two example annotations, and importantly allowed the annotators to choose ``tie'' when both the model responses are indistinguishable in quality. See the full set of guidelines in Appendix~\ref{app:human-guideline} and details of our website for collecting the annotations in Appendix~\ref{fig:website}. To avoid potential bias in order of the responses, we randomly swap the two responses.


\vspace{-0.2cm} \paragraph{Statistics.}
We collected 4 annotations for each of 1,752 instances . The annotators spent around 180s with a standard deviation of 79s on each annotation on average with a tie rate of 4.9\%.

\begin{table*}[t!]
    \centering  \small
    \setlength{\tabcolsep}{4.5pt}
    \begin{tabular}{p{0.19\linewidth} c c c c c c c c c c c c}
    \toprule
        \multirow{2}{*}{Method} & \multicolumn{8}{c}{General Category} & \multicolumn{3}{c}{Science Category}  & \multirow{2}{*}{All} \\
        \cmidrule(lr){2-9} \cmidrule(lr){10-12} &
         Brn & OQA & CQA & Ext & Gen & Rew & Sum & Cls & FC & MDS & RND  \\
    \midrule
        \textit{Heuristics} \\
        Random & \rab & \raoq & \racq & \rae & \rag & \rar & \ras & \rac & \raf & \ram & \raro & \rao \\
        Shorter & \shb & \shoq & \shcq & \she & \shg & \shr & \shs & \shc & \shf & \shm & \shro & \sho \\
        Longer & \lgb & \lgoq & \lgcq & \lge & \lgg & \lgr & \lgs & \lgc & \lgf & \lgm & \lgro & \lgo \\
        Rouge & \rob & \rooq & \rocq & \roe & \rog & \ror & \ros & \roc & \rof & \rom & \roro & \roo \\
    \midrule
        \multicolumn{8}{l}{\textit{LLM-as-a-Judge}} \\
        GPT4 & \gtb & \gtoq & \gtcq & \gte & \gtg & \gtr & \gts & \gtc & \gtf & \gtm & \gtro & \gto \\
        GPT4-Tb & \gttb & \gttoq & \gttcq & \gtte & \gttg & \gttr & \gtts & \gttc & \gttf & \gttm & \gttro & \gtto \\
        Llama-7B & \lsb & \lsoq & \lscq & \lse & \lsg & \lsr & \lss & \lsc & \lsf & \lsm & \lsro & \lso \\
        Llama-70B & \textbf{\lmb} & \lmoq & \lmcq & \lme & \lmg & \lmr & \textbf{\lms} & \lmc & \lmf & \lmm & \lmro & \lmo \\
    \midrule
        \multicolumn{8}{l}{\textit{LLM-as-a-Judge with human response}} \\
        GPT4 & \gtwb & \gtwoq & \gtwcq & \gtwe & \gtwg & \gtwr & \gtws & \gtwc & \gtwf & \gtwm & \gtwro & \gtwo \\
        GPT4-Tb& \gttwb & \gttwoq & \gttwcq & \gttwe & \gttwg & \gttwr & \gttws & \gttwc & \gttwf & \gttwm & \gttwro & \gttwo \\
        Llama-7B & \lswb & \lswoq & \lswcq & \lswe & \lswg & \lswr & \lsws & \lswc & \lswf & \lswm & \lswro & \lswo \\
        Llama-70B & \lmwb & \lmwoq & \textbf{\lmwcq} & \textbf{\lmwe} & \textbf{\lmwg} & \textbf{\lmwr} & \lmws & \textbf{\lmwc} & \lmwf & \textbf{\lmwm} & \textbf{\lmwro} & \lmwo \\
    \midrule
        \multicolumn{8}{l}{\textit{Embedding-Based}} \\
        RoBERTa-Large & \beb & \textbf{\beoq} & \becq & \bee & \beg & \ber & \bes & \bec & \textbf{\bef} & \bem & \bero & \beo \\
    \midrule
        \multicolumn{8}{l}{\textit{Perplexity-based}} \\
        Perplexity & \ppb & \ppoq & \ppcq & \ppe & \ppg & \ppr & \pps & \ppc & \ppf & \ppm & \ppro & \ppo \\
    \midrule
        \multicolumn{8}{l}{\textit{Composite}} \\
        GPT4 & \ourgb & \ourgoq & \ourgcq & \ourge & \ourgg & \ourgr & \ourgs & \ourgc & \ourgf & \ourgm & \ourgro & \ourgo \\
        GPT4-Tb & \ourgtb & \ourgtoq & \ourgtcq & \ourgte & \ourgtg & \ourgtr & \ourgts & \ourgtc & \ourgtf & \ourgtm & \ourgtro & \ourgto \\
        Llama-7B & \ourllsb & \ourllsoq & \ourllscq & \ourllse & \ourllsg & \ourllsr & \ourllss & \ourllsc & \ourllsf & \ourllsm & \ourllsro & \ourllso \\
        \textbf{Llama-70B (Ours)} & \oursb & \oursoq & \ourscq & \ourse & \oursg & \oursr & \ourss & \oursc & \oursf & \oursm & \oursro & \textbf{\ourso} \\
    \midrule
        Human & \hub & \huoq & \hucq & \hue & \hug & \hur & \hus & \huc & \huf & \hum & \huro & \huo \\
    \bottomrule
    \end{tabular}\vspace{-.1em}
    \caption{
    \textbf{Human Agreement Rates of Different Evaluation Methods on 11 Categories.} All numbers are average LOO agreement rates in \%. Bold numbers are the highest numbers with Llama-3.1-70B-Instruct for each categories, and we choose their corresponding methods to form the final composite method. When calculating Perplexity, we omit some instances in the human agreement datasets where the perplexity are not available with OpenAI models. Brn $\rightarrow$ Brainstorming; OQA $\rightarrow$ Open QA; CQA $\rightarrow$ Closed QA; Ext $\rightarrow$ Extraction; Gen $\rightarrow$ Generation; Rew $\rightarrow$ Rewriting; Sum $\rightarrow$ Summarization; Cls $\rightarrow$ Classification; FC $\rightarrow$ Fact Checking / Attributed QA; MDS $\rightarrow$ Multi-Document Synthesis; RND $\rightarrow$ Reasoning Over Numerical Data. 
 }\label{tab:agreement-results}
\end{table*}

\subsection{Evaluation Methods}\label{subsec:methods}
We evaluate a set of \textit{pairwise} evaluation methods~\citep{zheng2023judging}, i.e., those that select the better response between two candidate model responses, based on their agreement with the human judgments we collected.

\textbf{LLM-as-a-Judge} involves prompting a powerful LLM to judge the better response between a pair of responses from two models. This is the most common method used by prior work. We experiment with the choice of the judge model, and consider Llama-3.1-7B-Instruct, Llama-3.1-70B-Instruct~\citep{dubey2024llama}, GPT-4, and GPT-4-Turbo~\citep{achiam2023gpt}in our experiments. See Appendix~\ref{app:llm-template} for the prompt template we use. Note that we allow the models to judge 'tie' between the two model responses.


\textbf{LLM-as-a-Judge with human response} is similar to LLM-as-a-Judge except that it embeds human-written response into the prompt and instructs the judge to refer to it. See Appendix~\ref{app:llm-template} for the prompt template we use.  We experiment with the same set of four judges in this setting as well.

\textbf{Embedding-based} methods compute the similarity between the text embeddings of a model response and a human-written response, using the resulting score to select the response with the higher similarity. We use RoBERTa-Large~\citep{DBLP:journals/corr/abs-1907-11692} as the embedding model.

\textbf{Perplexity-based} method calculates the perplexities of the human-written  answer conditioned on the instruction for both the models, and selects the model with lower perplexity.

\textbf{Heuristic-based} methods include uniformly selecting one of the two responses (Random), naively preferring the shorter \slash\ longer response (Shorter \slash\ Longer), and selecting the response with a higher n-gram overlap (Rouge).



\textbf{Composite} select the best method from LLM-as-a-Judge, LLM-as-a-Judge with human response, and embedding-based methods for each category.




\subsection{Computing Human Agreement}
Following \citet{alpaca_eval}, we use the \textbf{Leave-One-Out (LOO) agreement rate} to evaluate the agreement between a method's output and the four human annotations for each sample. Concretely, we compute the frequency with which the evaluation method’s output matches the mode of each combination of 3 out of 4 human annotations, then average the results across all 4 possible combinations. We report the \textit{human agreement rate} as the average of LOO agreement rate over the all response pairs. To calculate the agreement rate within the human annotator themselves, we treat the remaining annotation as the ``model" prediction for each combination of 3 annotations and perform the same calculation. See Appendix~\ref{app:loo} for more details.

\section{Results}\label{sec:result}

\begin{figure}
\centering \footnotesize
\resizebox{0.8\columnwidth}{!}{\includegraphics[trim={0.0cm 0.5cm 0.0cm 0.5cm},clip]{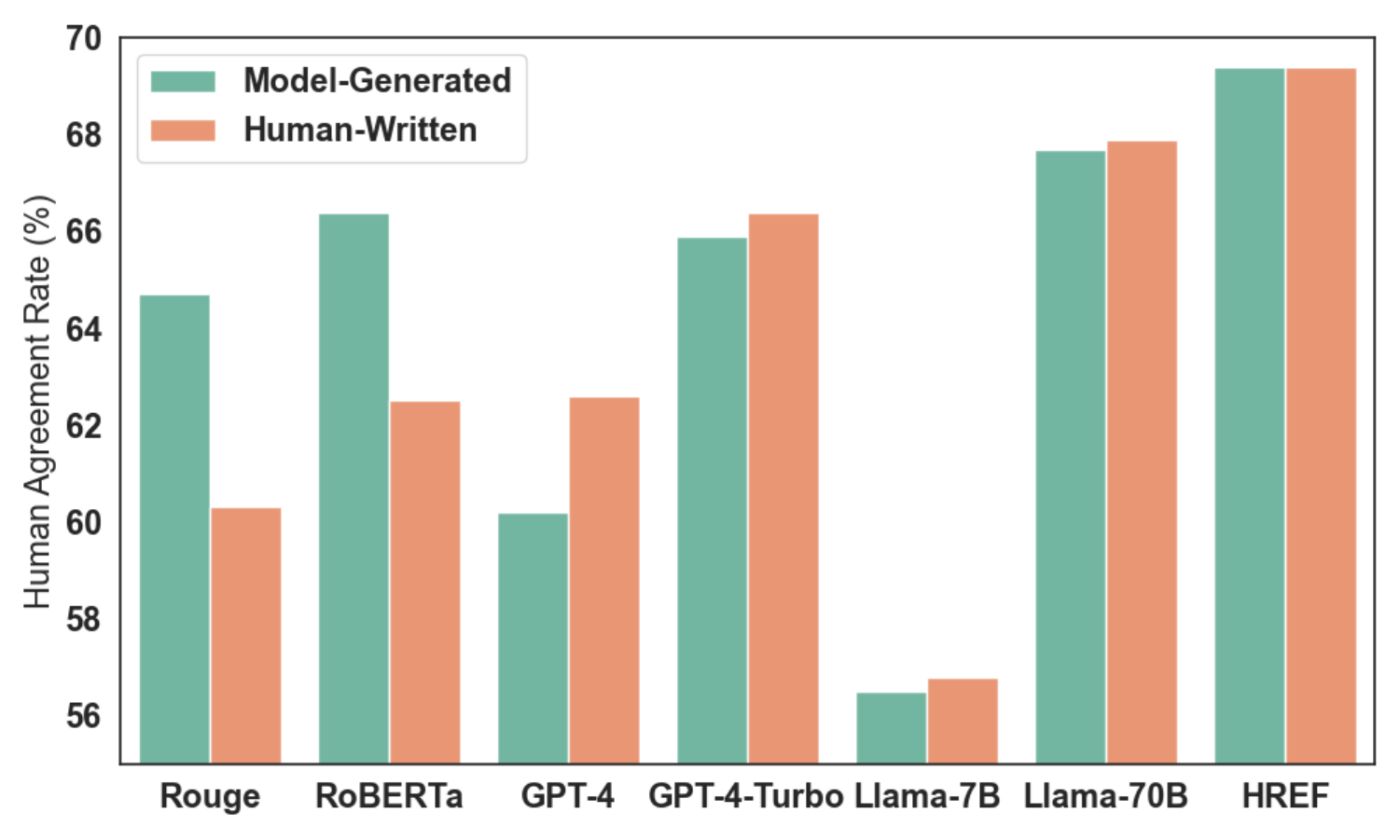}}
\caption{
\textbf{Human Agreement Rate using model-generated v.s. human-written responses.} Human response outperforms model response for LLM-based evaluation methods but underperforms for embedding-based evaluation methods.
}\label{fig:model-reference}
\end{figure}

In this section, we present the results from the experiment described in Section~\ref{sec:experimental-setup}, and we provide additional insights into why human-written responses are helpful in improving the evaluation methods.

\subsection{Main Results}\label{subsec:results}
Table~\ref{tab:agreement-results} presents the results of the human agreement analysis. 

\vspace{-0.2cm} \paragraph{Human agreement rates varies across task categories.} Tasks such as Brainstorming, Open QA, Summarization, and Multi-Document Synthesis, tend to have responses that vary in multiple dimensions, including general content, level of details, tone, etc. 
We observe that both the inner-agreement rate among human annotators and the agreement rates across all evaluation methods are lower within these task categories, indicating that humans apply divergent standards for judging LLM responses and weights various dimension of such open-ended responses differently. Conversely, categories that tends to have easily verifiable answers, including Close QA, Extraction, Classification, and Reasoning Over Numerical Data, appears to have higher agreements. Note that although Rewrite contains many open-ended instructions, a large portion of the instructions are verifiable as they ask for specific tone or format of the response. These findings highlight the importance to evaluate LLMs on specific task categories.

\vspace{-0.2cm} \paragraph{Llama-3.1-70B-Instruct is the best judge.}  Llama-3.1-70B-Instruct outperforms GPT-4 by 6\% and GPT-4-Turbo by 1.5\% without human responses, achieving the closest agreement rate compared to the human. It also outperforms GPT-4 by 4.2\%, GPT-4-Turbo by 1.3\%, and even humans by 0.9\% using human responses on average.

\vspace{-0.2cm} \paragraph{Human-written responses improve agreement with human judgments.} Across all models except Llama-3.1-7B-Instruct, embedding human-written responses into the prompts and using them as additional context frequently improves agreement with human judgments. The performance drop with Llama-3.1-7B-Instruct is likely because LLMs have to reach a certain capability threshold so that they understand how to properly utilize the human-written responses. 
In task categories Close QA, Extraction, Generation, Rewriting, Classification, Multi-
Document Synthesis, and Reasoning Over Numerical Data, using human-written responses brings an increment of 4.8\% on average in agreement with human for using Llama-3.1-70B-Instruct as the judge. 
For OpenQA, Summarization, and Fact Checking, we observe that human-written response improves agreement with human judgement for GPT-4 and GPT-4-Turbo but not for Llama models. This suggests that the capability of properly leverage human-written responses as additional context is inconsistent across different models for these task categories.
We also see that RoBERTa-Large is able to deliver the highest agreement rate with human on Open QA and Fact Checking.
These results show that, despite that the annotators who write the human response and the ones who annotate the preference are two different groups, a human-written response can help improve the judgment by serving as an additional context or a comparable reference. We will talk about the insights around the usefulness of human-written responses in the following Section~\ref{subsec:model-generated}. 

\begin{figure}
\centering \footnotesize
\resizebox{1.0\columnwidth}{!}{\includegraphics[trim={0.0cm 0.5cm 0.0cm 0.5cm},clip]{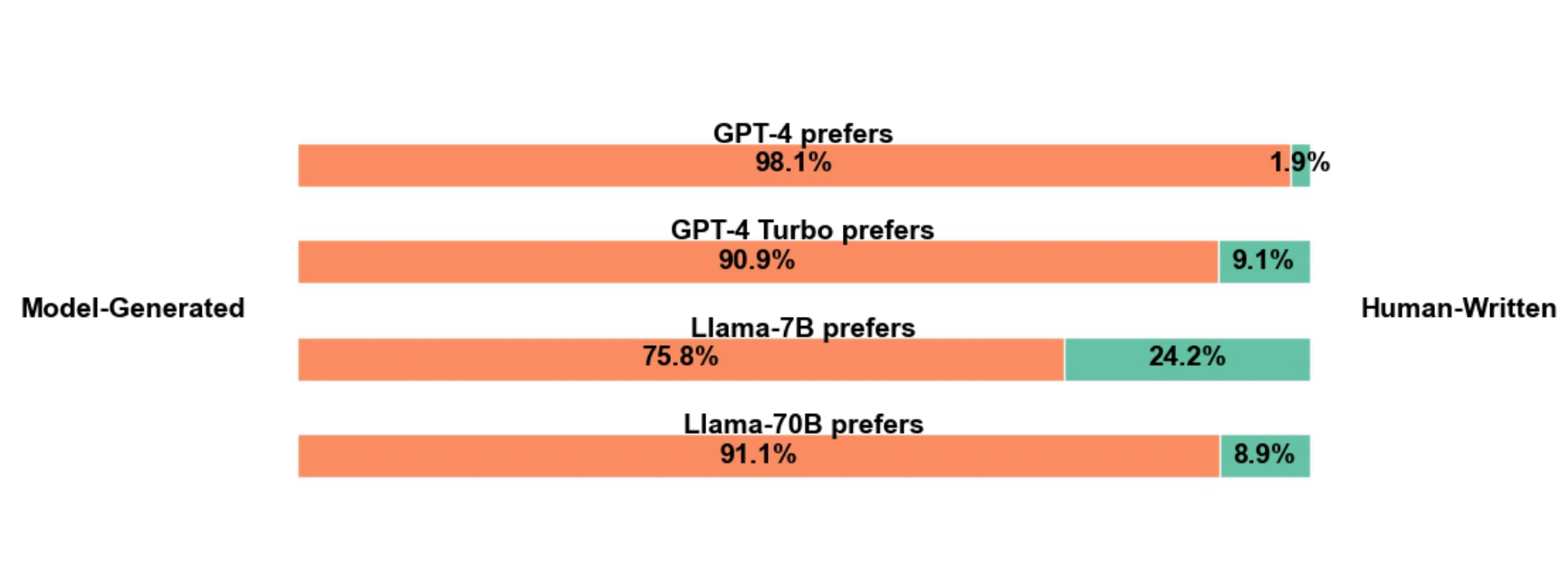}}
\caption{
\textbf{Judges Preference Between Model-Generated Responses v.s. Human-Written Responses}. Model-generated responses are in great favor of all the judges.
}\label{fig:reference-comparision}
\end{figure}

\vspace{-0.2cm} \paragraph{Choosing the best method for each category.} 
With the new set of evaluation methods that leverage human-written responses, we are provided with the option to select the best evaluation methods for each task categories and compose the final composite methods. Overall, the resulting composite method with Llama-3.1-70B-Instruct achieves 1.5\% higher in human agreement rate than only using Llama-3.1-70B-Instruct as a judge with human reference, outperforming human annotators' inner agreement rate by 2.4\%.


\subsection{Analysis: Leveraging Human References }\label{subsec:model-generated}
In order to understand the unique value of human-written responses, we compare them directly against \textit{model-generated response} proposed in \citet{zheng2023judging}.

\vspace{-0.2cm} \paragraph{Human-written responses are more useful than model-generated responses with LLM-as-a-Judge.} We use generate responses from GPT-4-Turbo 
for the instructions in the human agreement set and repeat the experiments in Section~\ref{sec:experimental-setup} with model-generated responses. Figure~\ref{fig:model-reference} demonstrates a comparison between using human-written responses and model-generated responses. We observe that with LLM-as-a-judge methods, human-written responses display higher agreement rates than model-generated responses across all judge models. This demonstrates that \textbf{references written by humans are consistently  more useful than those generated by even the strongest LLMs}. With embedding-based evaluation methods (RoBERTa and Rouge), using model-generated responses display higher agreements than human-written responses. This is due to the fact that model-generated responses are syntactically and stylistically more similar to each other than to human-written ones, likely biasing these simpler evaluation methods.

\vspace{-0.2cm} \paragraph{Why not directly compare against human responses?} We experimented with a setup where we prompt each LLM judge in Section~\ref{sec:experimental-setup} to directly compare model responses with human responses. Figure~\ref{fig:reference-comparision} shows that, surprisingly, all the judge models strongly prefer model responses over human responses despite their judgments being more aligned with those of human annotators when using human responses as additional context. This is likely because that the judge models strongly prefer the stylistic characteristics of model-generated responses. However, humans may prefer the style of human-written responses and other impactful dimensions, such as correctness, which are overlooked by the judge models. This demonstrates that \textbf{human-written responses are much more effective as additional context or additional reference for comparing model responses, rather than serving as the sole reference for direct comparison in evaluating response quality}.




\section{New Benchmark: \ours}\label{sec:preference-annotator}\begin{table}[t!]
    \centering  \small
    \setlength{\tabcolsep}{4.5pt}
    \begin{tabular}{c|ccccccc}
\toprule
\text{Benchmark} & \text{Size} & \text{Eval} & \text{BM} & \text{Judge} & \text{TaskCent} & \text{Private} & \text{HumResp}  \\
\midrule
\text{MT-Bench} & 80 & \text{Score} & --- & \text{gpt4} & \cmark & \xmark & \xmark \\
\text{AlpacaEval 2.0} & 805 & \text{PWC} & \text{gpt4-turbo} & \text{gpt4-turbo} & \xmark & \xmark & \xmark  \\
\text{Chatbot Arena} & --- & \text{PWC} & --- & \text{Human} & \xmark & \cmark & \xmark \\
\text{Arena-Hard} & 500 & \text{PWC} & \text{gpt4-0314} & \text{gpt4-turbo} & \xmark & \xmark & \xmark \\
\text{WildBench} & 1,024 & \text{Score/PWC} &\text{gpt4-turbo} & \text{three models} & \xmark & \xmark & \xmark \\
\textbf{\ours} & 4,258 & \text{PWC} & \text{Llama-405B}  & \text{Llama-70B} & \cmark & \cmark & \cmark \\
    \bottomrule
    
    \end{tabular}\vspace{-.1em}
    \caption{
    \textbf{Benchmark Comparision.} A comparison between the existing instructional LLM evaluation benchmarks and \ours. \textbf{TaskOrit} refers to whether the instructions are task-oriented. \textbf{PWC} refers to the paired comparison. \ours\ has the largest evaluation set, is the only benchmark that uses open-weight models (Llama-3.1 Instruct) as both the baseline model (BM) and the judge, is built with task-centric instruction, is completely private, and uses human-written responses (HumResp) to facilitate preference judgment.
 }\label{tab:benchmark-comparison}
\end{table}

Based on the insights that human-written responses significantly improves the evaluation of LLMs' instruction-following capability, we construct a new evaluation benchmark, \textbf{Human Response-guided Evaluation of instruction Following (\ours)}. See Table~\ref{tab:benchmark-comparison} for an overview of the comparison between \ours\ and similar existing benchmarks.  We release two evaluation sets in addition to the human agreement set we used for experiments described in Section~\ref{sec:experimental-setup}: a private evaluation set and a public development set.

    

\paragraph{Public Development Set}
We adopt a subset of the No Robots~\citep{no_robots} test split as the development set, which contains 430 human-written instruction and response pairs covering 8 out of the same 11 task categories as the evaluation set (See Figure~\ref{fig:dataset-distribution}). The remaining three scientific text understanding tasks are exclusive to in the evaluation set of \ours\ and can be considered held-out tasks. We generate a baseline model response from Llama-3.1-405B-Instruct-FP8 for each instruction. We will later show that the rankings on this set highly correlate with those from the evaluation set in Section~\ref{subsec:dev-transfer} \footnote{Data Link:  \url{https://huggingface.co/datasets/allenai/href}.}.

\subsection{Private Evaluation Set}\label{sec:dataset}

\begin{figure}
\centering \footnotesize
\resizebox{1.0\columnwidth}{!}{\includegraphics[trim={4.5cm 0.5cm 4.2cm 3cm},clip]{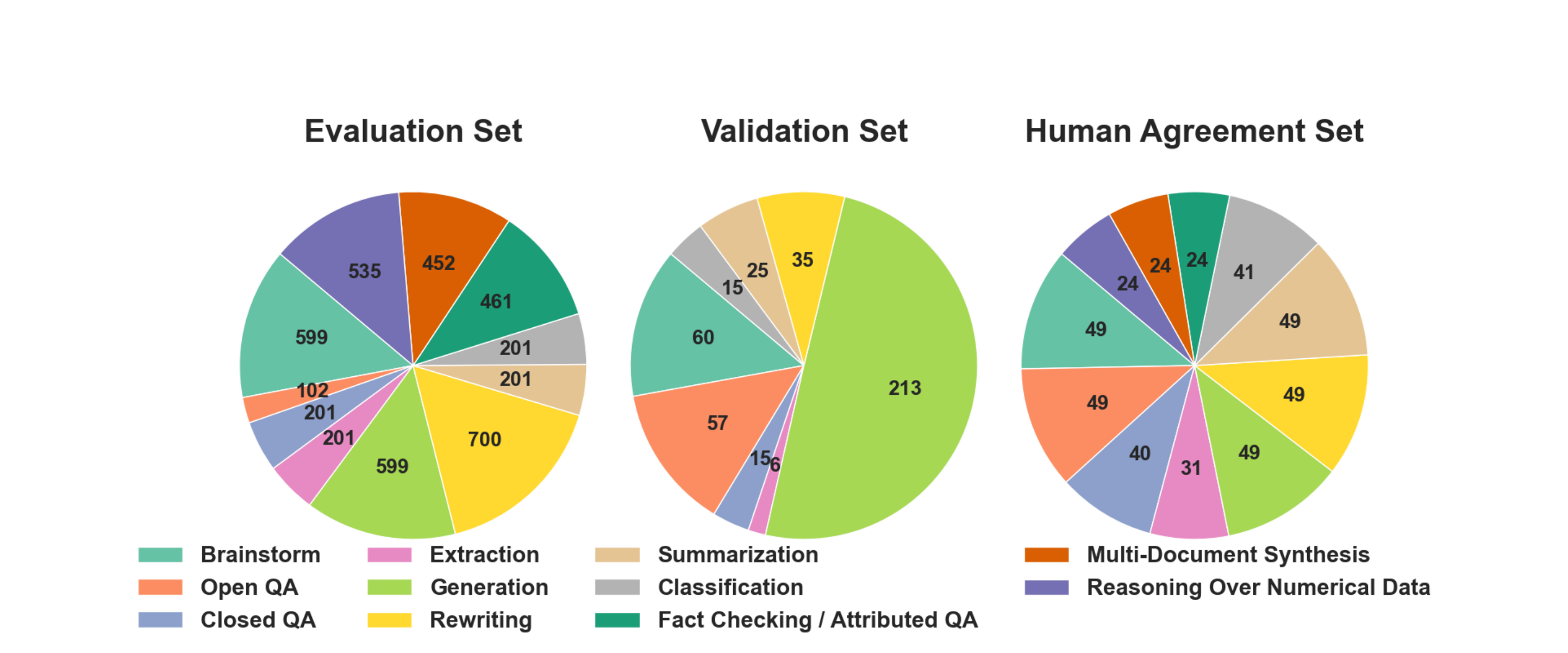}}
\caption{
\textbf{Task Categorical Distribution of the three subsets in \ours.} Left: evaluation set; Middle: development set; Right: human agreement set.
}\label{fig:dataset-distribution}
\end{figure}

\vspace{-0.2cm} \paragraph{Instruction and Human Response Collection.} We hire human experts to write instructions and corresponding responses specifically targetting the taxonomy of tasks shown in Table~\ref{tab:task-defination}. This results in 4,258 high quality instruction-response pairs. Figure~\ref{fig:dataset-distribution} Left shows the resulting distribution of the instructions.

\vspace{-0.2cm} \paragraph{Baseline Response Generation. } We generate a baseline response for each instruction to be compared against by a target model using the open model Llama-3.1-405B-Instruct-FP8 using greedy decoding. We compare this model with other choices for baseline models in Section~\ref{subsec:choice-of-judge-baseline}.


\subsection{Evaluation Details}
\vspace{-0.2cm} \paragraph{Pipeline.} For a target model, we first generate its response to each instruction to compare against the baseline model response using \ourmethod, and consider it a as win if \ourmethod\ either prefers the target model response or selects a tie. To obtain the final \textbf{\winrate}, we compute the frequency of wins for the target model across all data points.

\vspace{-0.2cm} \paragraph{Method Details.} Following the observation from Section~\ref{subsec:choice-of-judge-baseline}, we use the composite method with Llama-3.1-70B-Instruct as the judge model.

\vspace{-0.2cm} \paragraph{Decoding Strategy.} For reproducibility, we choose greedy decoding for these models. We find that this choice does not significantly impact the evaluation results---we find a high correlation (0.98 Spearman and 0.99 Pearson) between the results obtained from using greedy decoding and those obtained from using a temperature of 1.0 on our development set.

\begin{table*}[t!]
    \centering  \small
    \setlength{\tabcolsep}{4.5pt}
    \begin{tabular}{l c c l c c }
    \toprule
        Dataset & Size & \# Category & Annotation & Release \\
    \midrule
        Evaluation & 4,258 & 11 & \xmark & \xmark \\
        Development & 430 &  8 & \xmark & \cmark \\
        Agreement & 438 & 11 & \cmark & \cmark \\
    \bottomrule
    \end{tabular}\vspace{-.1em}
    \caption{
    \textbf{\ours\ Subsets Comparison.} An comparison of important aspects among the three subsets.
 }\label{tab:dataset-overview}
\end{table*}

\vspace{-0.2cm} \paragraph{Prompt Template.}  To reduce the difference between the model judge and human annotations in terms of their annotation criteria, we adopt the prompt template given to the human annotators (See Appendix~\ref{app:human-guideline}) and carefully modify it for LLM prompting (See Appendex~\ref{app:llm-template}). We compare this with other choices of prompts in Section~\ref{subsec:choice-of-prompt}. 
\vspace{-0.2cm} \paragraph{Expected Win Rate.} Because we allow tie in LLM-as-a-Judge both with and without human response, we define \textbf{expected win rate} as the sum of the frequency that our composite method prefers the target model over the baseline model and \textbf{half} the frequency that our composite method selects a tie, over all samples.

Note that we keep the option of judging ties and consider it as a win.

\begin{table*}[t!]
    \centering  \small
    \setlength{\tabcolsep}{2.5pt}
    \begin{tabular}{c p{0.26\linewidth} |  c  c c c c c c c c c c | c}
    \toprule
        \multirow{2}{*}{Rank} & \multirow{2}{*}{Model} &
         \multicolumn{8}{c}{General Category} & \multicolumn{3}{c}{Science Category}  & \multirow{2}{*}{All} \\
        \cmidrule(lr){3-10} \cmidrule(lr){11-13} &
         & Brn & OQA & CQA & Ext & Gen & Rew & Sum & Cls & FC & MDS & RND  \\
    \midrule
        1 & Llama-3.1-70B-Instruct & \llamadbhabinstructb & \llamadbhabinstructoq & \llamadbhabinstructcq & \llamadbhabinstructe & \llamadbhabinstructg & \llamadbhabinstructr & \llamadbhabinstructs & \llamadbhabinstructc & \llamadbhabinstructro & \llamadbhabinstructm & \llamadbhabinstructf & \llamadbhabinstructo \\
        1 & Mistral-Large-Instruct-2407 & \mistrallargeinstructceahb & \mistrallargeinstructceahoq & \mistrallargeinstructceahcq & \mistrallargeinstructceahe & \mistrallargeinstructceahg & \mistrallargeinstructceahr & \mistrallargeinstructceahs & \mistrallargeinstructceahc & \mistrallargeinstructceahro & \mistrallargeinstructceahm & \mistrallargeinstructceahf & \mistrallargeinstructceaho \\
        3 & Qwen2.5-72B-Instruct & \qwencfhcbinstructb & \qwencfhcbinstructoq & \qwencfhcbinstructcq & \qwencfhcbinstructe & \qwencfhcbinstructg & \qwencfhcbinstructr & \qwencfhcbinstructs & \qwencfhcbinstructc & \qwencfhcbinstructro & \qwencfhcbinstructm & \qwencfhcbinstructf & \qwencfhcbinstructo \\
        3 & Qwen1.5-110B-Chat & \qwenbfbbabchatb & \qwenbfbbabchatoq & \qwenbfbbabchatcq & \qwenbfbbabchate & \qwenbfbbabchatg & \qwenbfbbabchatr & \qwenbfbbabchats & \qwenbfbbabchatc & \qwenbfbbabchatro & \qwenbfbbabchatm & \qwenbfbbabchatf & \qwenbfbbabchato \\
        3 & Llama-3.1-Tulu-3-70B-DPO & \llamadbtuludhabdpob & \llamadbtuludhabdpooq & \llamadbtuludhabdpocq & \llamadbtuludhabdpoe & \llamadbtuludhabdpog & \llamadbtuludhabdpor & \llamadbtuludhabdpos & \llamadbtuludhabdpoc & \llamadbtuludhabdporo & \llamadbtuludhabdpom & \llamadbtuludhabdpof & \llamadbtuludhabdpoo \\
        3 & Llama-3.1-Tulu-3-70B & \llamadbtuludhabb & \llamadbtuludhaboq & \llamadbtuludhabcq & \llamadbtuludhabe & \llamadbtuludhabg & \llamadbtuludhabr & \llamadbtuludhabs & \llamadbtuludhabc & \llamadbtuludhabro & \llamadbtuludhabm & \llamadbtuludhabf & \llamadbtuludhabo \\
        3 & Mistral-Small-Instruct-2409 & \mistralsmallinstructceajb & \mistralsmallinstructceajoq & \mistralsmallinstructceajcq & \mistralsmallinstructceaje & \mistralsmallinstructceajg & \mistralsmallinstructceajr & \mistralsmallinstructceajs & \mistralsmallinstructceajc & \mistralsmallinstructceajro & \mistralsmallinstructceajm & \mistralsmallinstructceajf & \mistralsmallinstructceajo \\
        8 & Llama-3.1-8B-Instruct & \llamadbibinstructb & \llamadbibinstructoq & \llamadbibinstructcq & \llamadbibinstructe & \llamadbibinstructg & \llamadbibinstructr & \llamadbibinstructs & \llamadbibinstructc & \llamadbibinstructro & \llamadbibinstructm & \llamadbibinstructf & \llamadbibinstructo \\
        9 & Yi-1.5-34B-Chat & \yibfdebchatb & \yibfdebchatoq & \yibfdebchatcq & \yibfdebchate & \yibfdebchatg & \yibfdebchatr & \yibfdebchats & \yibfdebchatc & \yibfdebchatro & \yibfdebchatm & \yibfdebchatf & \yibfdebchato \\
        9 & Qwen2-72B-Instruct & \qwenchcbinstructb & \qwenchcbinstructoq & \qwenchcbinstructcq & \qwenchcbinstructe & \qwenchcbinstructg & \qwenchcbinstructr & \qwenchcbinstructs & \qwenchcbinstructc & \qwenchcbinstructro & \qwenchcbinstructm & \qwenchcbinstructf & \qwenchcbinstructo \\
        9 & OLMo-2-1124-13B-Instruct & \olmocbbcebdbinstructb & \olmocbbcebdbinstructoq & \olmocbbcebdbinstructcq & \olmocbbcebdbinstructe & \olmocbbcebdbinstructg & \olmocbbcebdbinstructr & \olmocbbcebdbinstructs & \olmocbbcebdbinstructc & \olmocbbcebdbinstructro & \olmocbbcebdbinstructm & \olmocbbcebdbinstructf & \olmocbbcebdbinstructo \\
        9 & Phi-3-medium-4k-instruct & \phidmediumekinstructb & \phidmediumekinstructoq & \phidmediumekinstructcq & \phidmediumekinstructe & \phidmediumekinstructg & \phidmediumekinstructr & \phidmediumekinstructs & \phidmediumekinstructc & \phidmediumekinstructro & \phidmediumekinstructm & \phidmediumekinstructf & \phidmediumekinstructo \\
        9 & Llama-3.1-Tulu-3-8B-DPO & \llamadbtuludibdpob & \llamadbtuludibdpooq & \llamadbtuludibdpocq & \llamadbtuludibdpoe & \llamadbtuludibdpog & \llamadbtuludibdpor & \llamadbtuludibdpos & \llamadbtuludibdpoc & \llamadbtuludibdporo & \llamadbtuludibdpom & \llamadbtuludibdpof & \llamadbtuludibdpoo \\
        9 & Llama-3.1-Tulu-3-8B & \llamadbtuludibb & \llamadbtuludiboq & \llamadbtuludibcq & \llamadbtuludibe & \llamadbtuludibg & \llamadbtuludibr & \llamadbtuludibs & \llamadbtuludibc & \llamadbtuludibro & \llamadbtuludibm & \llamadbtuludibf & \llamadbtuludibo \\
        15 & OLMo-2-1124-7B-Instruct & \olmocbbcehbinstructb & \olmocbbcehbinstructoq & \olmocbbcehbinstructcq & \olmocbbcehbinstructe & \olmocbbcehbinstructg & \olmocbbcehbinstructr & \olmocbbcehbinstructs & \olmocbbcehbinstructc & \olmocbbcehbinstructro & \olmocbbcehbinstructm & \olmocbbcehbinstructf & \olmocbbcehbinstructo \\
        15 & tulu-2-dpo-70b & \tulucdpohabb & \tulucdpohaboq & \tulucdpohabcq & \tulucdpohabe & \tulucdpohabg & \tulucdpohabr & \tulucdpohabs & \tulucdpohabc & \tulucdpohabro & \tulucdpohabm & \tulucdpohabf & \tulucdpohabo \\
        17 & Llama-2-70b-chat-hf & \llamachabchathfb & \llamachabchathfoq & \llamachabchathfcq & \llamachabchathfe & \llamachabchathfg & \llamachabchathfr & \llamachabchathfs & \llamachabchathfc & \llamachabchathfro & \llamachabchathfm & \llamachabchathff & \llamachabchathfo \\
        17 & Mistral-7B-Instruct-v0.3 & \mistralhbinstructvadb & \mistralhbinstructvadoq & \mistralhbinstructvadcq & \mistralhbinstructvade & \mistralhbinstructvadg & \mistralhbinstructvadr & \mistralhbinstructvads & \mistralhbinstructvadc & \mistralhbinstructvadro & \mistralhbinstructvadm & \mistralhbinstructvadf & \mistralhbinstructvado \\
        17 & Llama-3.1-Tulu-3-70B-SFT & \llamadbtuludhabsftb & \llamadbtuludhabsftoq & \llamadbtuludhabsftcq & \llamadbtuludhabsfte & \llamadbtuludhabsftg & \llamadbtuludhabsftr & \llamadbtuludhabsfts & \llamadbtuludhabsftc & \llamadbtuludhabsftro & \llamadbtuludhabsftm & \llamadbtuludhabsftf & \llamadbtuludhabsfto \\
        20 & WizardLM-13B-V1.2 & \wizardlmbdbvbcb & \wizardlmbdbvbcoq & \wizardlmbdbvbccq & \wizardlmbdbvbce & \wizardlmbdbvbcg & \wizardlmbdbvbcr & \wizardlmbdbvbcs & \wizardlmbdbvbcc & \wizardlmbdbvbcro & \wizardlmbdbvbcm & \wizardlmbdbvbcf & \wizardlmbdbvbco \\
        20 & Llama-2-13b-chat-hf & \llamacbdbchathfb & \llamacbdbchathfoq & \llamacbdbchathfcq & \llamacbdbchathfe & \llamacbdbchathfg & \llamacbdbchathfr & \llamacbdbchathfs & \llamacbdbchathfc & \llamacbdbchathfro & \llamacbdbchathfm & \llamacbdbchathff & \llamacbdbchathfo \\
        20 & tulu-v2.5-ppo-13b & \tuluvcfppobdbufmeanhabufrmb & \tuluvcfppobdbufmeanhabufrmoq & \tuluvcfppobdbufmeanhabufrmcq & \tuluvcfppobdbufmeanhabufrme & \tuluvcfppobdbufmeanhabufrmg & \tuluvcfppobdbufmeanhabufrmr & \tuluvcfppobdbufmeanhabufrms & \tuluvcfppobdbufmeanhabufrmc & \tuluvcfppobdbufmeanhabufrmro & \tuluvcfppobdbufmeanhabufrmm & \tuluvcfppobdbufmeanhabufrmf & \tuluvcfppobdbufmeanhabufrmo \\
        20 & tulu-2-dpo-13b & \tulucdpobdbb & \tulucdpobdboq & \tulucdpobdbcq & \tulucdpobdbe & \tulucdpobdbg & \tulucdpobdbr & \tulucdpobdbs & \tulucdpobdbc & \tulucdpobdbro & \tulucdpobdbm & \tulucdpobdbf & \tulucdpobdbo \\
        20 & vicuna-13b-v1.5 & \vicunabdbvbfb & \vicunabdbvbfoq & \vicunabdbvbfcq & \vicunabdbvbfe & \vicunabdbvbfg & \vicunabdbvbfr & \vicunabdbvbfs & \vicunabdbvbfc & \vicunabdbvbfro & \vicunabdbvbfm & \vicunabdbvbff & \vicunabdbvbfo \\
        25 & Llama-3.1-Tulu-3-8B-SFT & \llamadbtuludibsftb & \llamadbtuludibsftoq & \llamadbtuludibsftcq & \llamadbtuludibsfte & \llamadbtuludibsftg & \llamadbtuludibsftr & \llamadbtuludibsfts & \llamadbtuludibsftc & \llamadbtuludibsftro & \llamadbtuludibsftm & \llamadbtuludibsftf & \llamadbtuludibsfto \\
        25 & Llama-2-7b-chat-hf & \llamachbchathfb & \llamachbchathfoq & \llamachbchathfcq & \llamachbchathfe & \llamachbchathfg & \llamachbchathfr & \llamachbchathfs & \llamachbchathfc & \llamachbchathfro & \llamachbchathfm & \llamachbchathff & \llamachbchathfo \\
        27 & vicuna-7b-v1.5 & \vicunahbvbfb & \vicunahbvbfoq & \vicunahbvbfcq & \vicunahbvbfe & \vicunahbvbfg & \vicunahbvbfr & \vicunahbvbfs & \vicunahbvbfc & \vicunahbvbfro & \vicunahbvbfm & \vicunahbvbff & \vicunahbvbfo \\
        27 & tulu-2-dpo-7b & \tulucdpohbb & \tulucdpohboq & \tulucdpohbcq & \tulucdpohbe & \tulucdpohbg & \tulucdpohbr & \tulucdpohbs & \tulucdpohbc & \tulucdpohbro & \tulucdpohbm & \tulucdpohbf & \tulucdpohbo \\
        29 & OLMo-7B-SFT-hf & \olmohbsfthfb & \olmohbsfthfoq & \olmohbsfthfcq & \olmohbsfthfe & \olmohbsfthfg & \olmohbsfthfr & \olmohbsfthfs & \olmohbsfthfc & \olmohbsfthfro & \olmohbsfthfm & \olmohbsfthff & \olmohbsfthfo \\
        29 & mpt-7b-chat & \mpthbchatb & \mpthbchatoq & \mpthbchatcq & \mpthbchate & \mpthbchatg & \mpthbchatr & \mpthbchats & \mpthbchatc & \mpthbchatro & \mpthbchatm & \mpthbchatf & \mpthbchato \\
        29 & koala-13B-HF & \koalabdbhfb & \koalabdbhfoq & \koalabdbhfcq & \koalabdbhfe & \koalabdbhfg & \koalabdbhfr & \koalabdbhfs & \koalabdbhfc & \koalabdbhfro & \koalabdbhfm & \koalabdbhff & \koalabdbhfo \\
        32 & gpt4all-13b-snoozy & \gpteallbdbsnoozyb & \gpteallbdbsnoozyoq & \gpteallbdbsnoozycq & \gpteallbdbsnoozye & \gpteallbdbsnoozyg & \gpteallbdbsnoozyr & \gpteallbdbsnoozys & \gpteallbdbsnoozyc & \gpteallbdbsnoozyro & \gpteallbdbsnoozym & \gpteallbdbsnoozyf & \gpteallbdbsnoozyo \\
        32 & OLMo-7B-0724-Instruct-hf & \olmohbahceinstructhfb & \olmohbahceinstructhfoq & \olmohbahceinstructhfcq & \olmohbahceinstructhfe & \olmohbahceinstructhfg & \olmohbahceinstructhfr & \olmohbahceinstructhfs & \olmohbahceinstructhfc & \olmohbahceinstructhfro & \olmohbahceinstructhfm & \olmohbahceinstructhff & \olmohbahceinstructhfo \\
        32 & dolly-v2-12b & \dollyvcbcbb & \dollyvcbcboq & \dollyvcbcbcq & \dollyvcbcbe & \dollyvcbcbg & \dollyvcbcbr & \dollyvcbcbs & \dollyvcbcbc & \dollyvcbcbro & \dollyvcbcbm & \dollyvcbcbf & \dollyvcbcbo \\
        32 & koala-7B-HF & \koalahbhfb & \koalahbhfoq & \koalahbhfcq & \koalahbhfe & \koalahbhfg & \koalahbhfr & \koalahbhfs & \koalahbhfc & \koalahbhfro & \koalahbhfm & \koalahbhff & \koalahbhfo \\
        32 & dolly-v2-7b & \dollyvchbb & \dollyvchboq & \dollyvchbcq & \dollyvchbe & \dollyvchbg & \dollyvchbr & \dollyvchbs & \dollyvchbc & \dollyvchbro & \dollyvchbm & \dollyvchbf & \dollyvchbo \\
        37 & oasst-sft-1-pythia-12b & \oasstsftbpythiabcbb & \oasstsftbpythiabcboq & \oasstsftbpythiabcbcq & \oasstsftbpythiabcbe & \oasstsftbpythiabcbg & \oasstsftbpythiabcbr & \oasstsftbpythiabcbs & \oasstsftbpythiabcbc & \oasstsftbpythiabcbro & \oasstsftbpythiabcbm & \oasstsftbpythiabcbf & \oasstsftbpythiabcbo \\
    \bottomrule
    \end{tabular}\vspace{-.1em}
    \caption{
    \textbf{\winrate s of all 37 starting models evaluated on the evaluation set of \ours.} All numbers are in \%. ($i$) indicates the ranking. Brn $\rightarrow$ Brainstorm; OQA $\rightarrow$ Open QA; CQA $\rightarrow$ Closed QA; Ext $\rightarrow$ Extraction; Gen $\rightarrow$ Generation; Rew $\rightarrow$ Rewriting; Sum $\rightarrow$ Summarization; Cls $\rightarrow$ Classification; FC $\rightarrow$ Fact Checking / Attributed QA; MDS $\rightarrow$ Multi-Document Synthesis; RND $\rightarrow$ Reasoning Over Numerical Data;  
 }\label{tab:main-test-results}
\end{table*}

\subsection{Results on Current LLMs}
 We evaluate 37 LLMs with a variety of model families and sizes on \ours\ as the initial benchmark. Table~\ref{tab:main-test-results} presents the results ranked by their total \winrate s, along with their \winrate s in each of the 11 categories. See the full table in Appendex~\ref{app:full-eval-results}.

In general, LLMs with larger sizes display higher \winrate s, and such trends holds consistently within the same model family. For example, Llama-2-70b-chat-hf holds a higher \winrate\ than Llama-2-13b-chat-hf on average. Also note that model \winrate s vary across different categories. For example, while Mistral-Large-Instruct-2407 a high average \winrate\ among the models that we evaluate, it performs poolly in Open QA. This demonstrates the importance of focusing into the evaluation on individual task and underscores the advantage of \ours\ in providing task-centric evaluation.  

\vspace{-0.2cm} \paragraph{Correlation with evaluation on the development set.}\label{subsec:dev-transfer} 
We also evaluate the same group of LLMs on our development set with 8 categories (See Section~\ref{sec:dataset}), additionally with several GPT models. See the full results in Appendex~\ref{app:full-eval-results}. We observe similar trends to those seen in the test dataset. To validate that model developers can expect a reasonable transfer of their results from the public development set to the private evaluation set, we calculate the correlation of the \winrate s between these two sets and observe high correlations: a Spearman correlation of 0.98 and a Pearson correlation of 0.99.

\begin{figure}
\centering \footnotesize
\resizebox{1.0\columnwidth}{!}{\includegraphics[trim={0.0cm 0.5cm 1.3cm 0.5cm},clip]{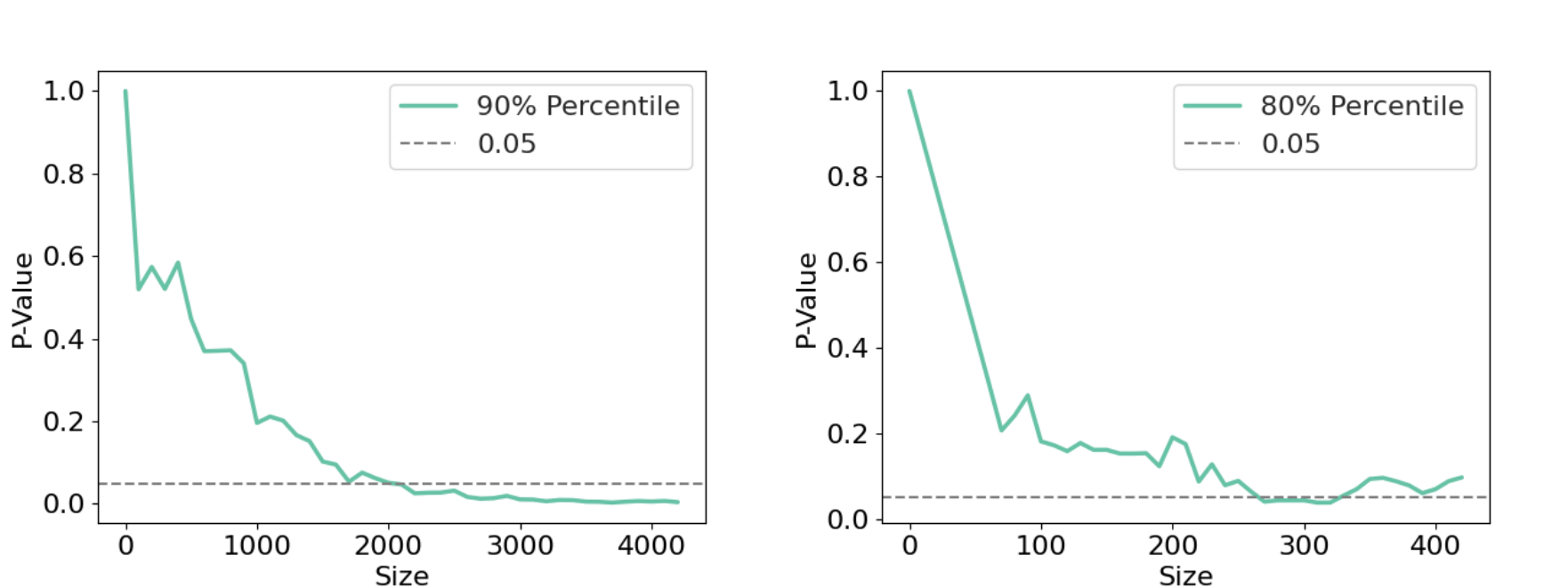}}
\caption{
\textbf{P-values of paired T-test on annotations across 13 models on the evaluation and development set.} Evaluation set on the left; Development set on the right. We show the average and 90th and 80th quantile p-values from doing paired t-test among all model pairs among 13 models with different numbers of annotated samples used. 
}\label{fig:p-values}
\end{figure}

\subsection{Statistical significance}\label{subsec:distinguish}
To ensure the reliability of our evaluation set in distinguishing between models, we evaluate HREF’s capability of statistically distinguishing among a diverse set of models of reasonable size.
Specifically, we sample from a pool of 13 models following ~\citet{alpaca_eval} but use a set of more recent and diverse models \footnote{Qwen1.5-110B-Chat, Mistral-Large-Instruct-2407, Yi-1.5-34B-Chat, tulu-2-dpo-70b, vicuna-13b-v1.5, Qwen2-72B-Instruct, mpt-7b-chat, koala-7B-HF, OLMo-7B-SFT-hf, dolly-v2-12b, Llama-2-7b-chat-hf, oasst-sft-1-pythia-12b, gpt4all-13b-snoozy-t=0.0}. For each pair of models, we apply a paired t-test to evaluate the null hypothesis that the preference predictions from the pair of models have identical expected values, and we measure the resulting p-values.
 We perform this analysis on both of the evaluation set and development set.

\vspace{-0.2cm} \paragraph{Capacity of the development and test sets. } Figure~\ref{fig:p-values} Left shows that with fewer than 2000 samples in the evaluation set, the p-values at 90th quantile falls below 0.05, which suggests that our evaluation set is able to statistically significantly distinguish between 90\% of the model pairs. Similarly, Figure~\ref{fig:p-values} right suggests that our development set is able to statistically significantly distinguish between 80\% of the model pairs.

\vspace{-0.2cm} \paragraph{Relevance of HREF.} As the size of the model pool and the strength of the models in the pool increase, the chance that a model pair will be indistinguishable (t-test with a p-value less than 0.05) will also increase. In other words, a larger evaluation set will be needed to distinguish more and stronger models.
Hence, as the community keeps developing stronger models, we expect \ours, with the largest evaluation set among similar benchmarks, to remain relevant for longer. 


\section{Discussion on Design Choices}\label{sec:discussion}\begin{figure}
\centering \footnotesize
\resizebox{0.7\columnwidth}{!}{\includegraphics[trim={0.5cm 0.5cm 0.35cm 0.35cm},clip]{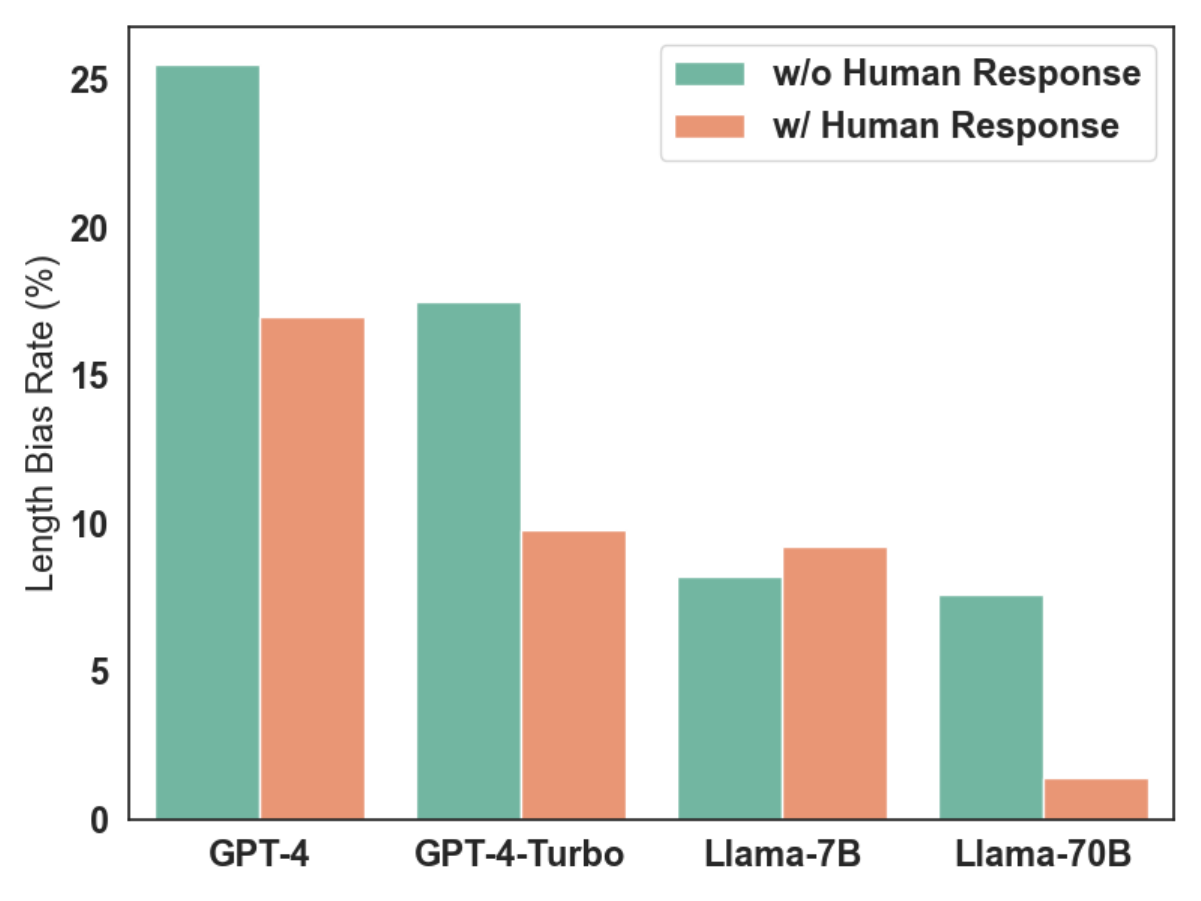}}
\caption{
\textbf{Length Bias Rate of Different LLM Judges}. It is clear that Llama-3.1-70B Instruct has the least length bias, and such bias is further reduced when using human-written responses as additional context. 
}\label{fig:length-difference}
\end{figure}

In this section, we discuss the specific design choices and the advantages they bring to \ours, including the choice of the judge model for LLM-as-a-Judge, and choice of the baseline model, and the choice of the prompt template.

\subsection{Choice of the Judge and Baseline Models}\label{subsec:choice-of-judge-baseline}
Unlike prior work~\citep{alpaca_eval, chiang2024chatbot, zheng2023judging, li2024crowdsourced, lin2024wildbench}, we choose Llama-3.1-70B-Instruct as the LLM judge and Llama-3.1-405B-Instruct-FP8 as our baseline model instead of GPT models. In this section, we discuss the rationale behind such choices.

\vspace{-0.2cm} \paragraph{High Human Agreement Rate with the Judge Model.} Llama-3.1-70B agrees with human judgmements the most on \ours\ as discussed in Section~\ref{sec:result}.

\vspace{-0.2cm} \paragraph{A Less Length-biased Judge Model.} Previous work~\citep{dubois2024length, lin2024wildbench, li2024crowdsourced} has observed that the judge LLMs strongly prefer longer responses and has adopted length normalization methods to account for such bias. We quantify the length bias of various judge models on our human agreement set, by measuring the difference between each judge's frequency of preferring longer responses versus the frequency of preferring shorter responses. We refer to this difference as the \textit{length bias rate}. Since we explicitly control for response length while sampling responses in the human agreement set (see Section~\ref{subsec:human-agreement-instruction-collection}), we expect a model with no length bias to have a length bias rate close to 0\% on our dataset. Figure~\ref{fig:length-difference} shows that Llama-3.1-70B-Instruct has the lowest length bias rate among all the four judge models that we experiment with. The use of human written responses further lowers its length bias rate to 1.4\%. As a result, we chose not to add any length debiasing controls.

\vspace{-0.2cm} \paragraph{A Less Costly Judge Model. } Referring to the price from Lepton AI\footnote{\url{https://www.lepton.ai/pricing}.}, Llama-3.1-70B-Instruct is at least 12.5 times cheaper than GPT-4 Turbo, and 37.5 times cheaper than GPT-4. To minimize the computational requirements of evaluating new models, we restrict the evaluator size to 70B, which requires at most 4 A100 GPUs to run.

\begin{figure}
\centering \footnotesize
\resizebox{0.8\columnwidth}{!}{\includegraphics[trim={0.5cm 0.5cm 0.40cm 0.4cm},clip]{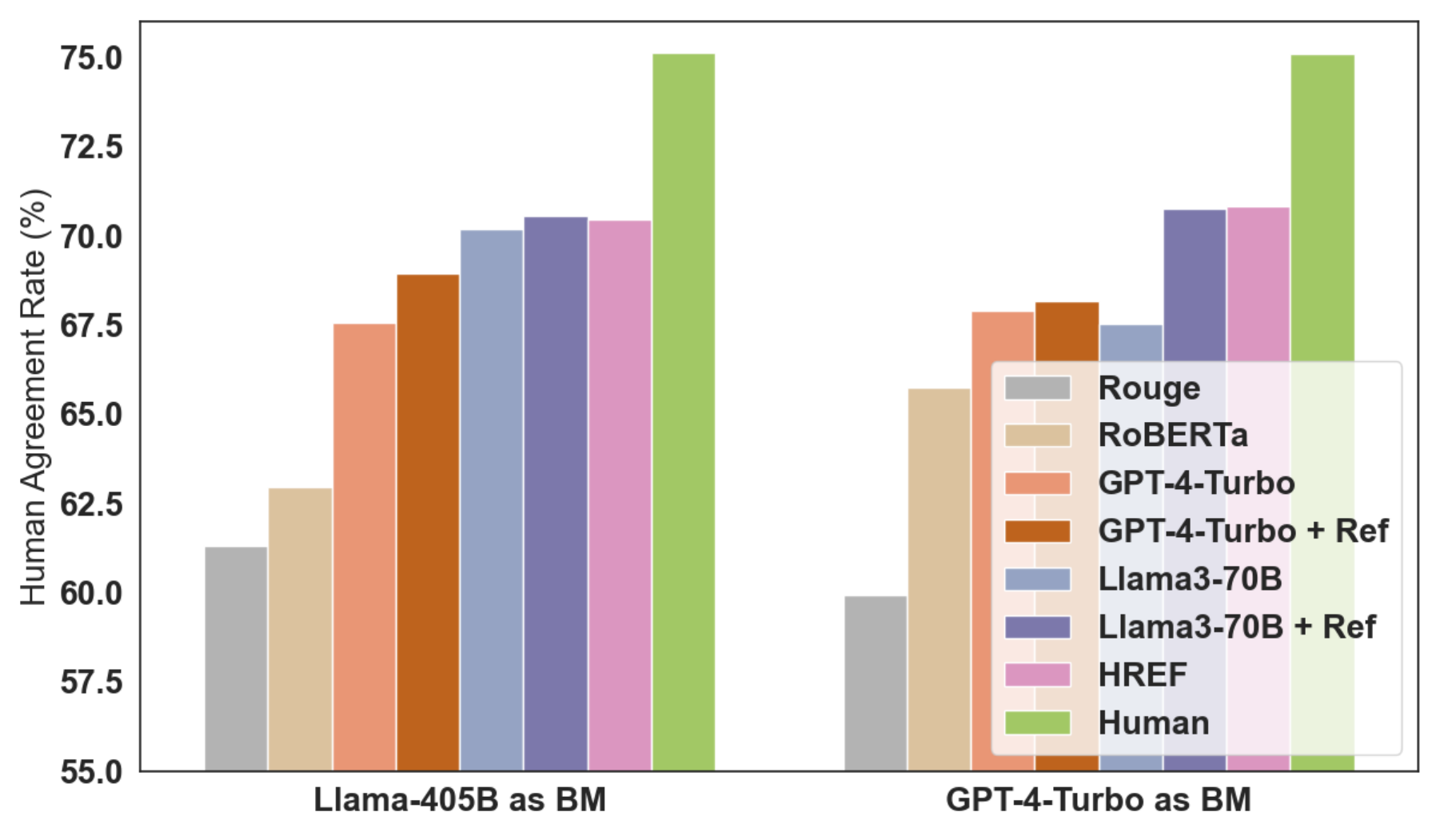}}
\caption{
\textbf{Impact of Changing Baseline Model}. The average human agreement rates of various evaluators using two different baseline models. We observe very similar trends when using Llama-3.1-405B-Instruct-FP8 and GPT-4-Turbo as the baseline model.
}\label{fig:baseline_model}
\end{figure}

\vspace{-0.2cm} \paragraph{Invariant to a Baseline Model Change.}
To analyze the impact of the choice of the baseline model, we conduct the same experiments as in Section~\ref{sec:result} but with GPT-4-Turbo as the baseline model on a subset of 1100 samples (275 instruction) of the human agreement set. Figure~\ref{fig:baseline_model} compares the average human agreement rates of various evaluators using Llama-3.1-405B-Instruct-FP8 and GPT-4-Turbo as the baseline models. We observe similar trends with both the baseline models, indicating that the reliability of the evaluation setup is unaffected by using the open-weight Llama model instead of the closed GPT-4-Turbo model.

\vspace{-0.2cm} \paragraph{Reproducible Evaluations.} Closed API models can be modified internally causing their outputs to change over time or eventually put to retirement, all of which makes evaluations relying on them irreproducible. In contrast, using an open-weight model like Llama-3.1-70B-Instruct, renders \ours\ more transparent and reproducible. 

\vspace{-0.2cm} \paragraph{Feasibility of a Private Test Set.} Using API models as judges requires sharing the instructions and responses with those models, meaning that the test set cannot remain truly private. Moreover, the common practice of synthesizing training datasets from such API models can potentially lead to test set contamination. By using an open-weight models running locally, we can keep our test data truly private to all models.

\subsection{Choice of the Prompt Template}\label{subsec:choice-of-prompt}
Unlike prior work such as AlpacaEval, we directly transform the guidelines we provide to human annotators into the prompt we provide to the judge LLMs, and we show the reasoning behind such choice here. We structure each prompt template into two components: a guideline and a list of demonstration examples. We interchange these components with those from AlpacaEval and compare the 4 resulting prompt templates using Llama-3.1-70B-Instruct with human-written responses on our development set as shown in Table~\ref{tab:prompt-templates}. 
 Table~\ref{tab:prompt-templates} shows that using a different set of examples (Prompt B), dropping the examples (Prompt C), or completely changing the prompt (Prompt D) negatively impacts agreement with human annotators compared to aligning the model prompt with the guidelines provided to human annotators (Prompt A). These results imply that \textbf{ensuring the consistency between the guidelines given to human annotators and the prompts for LLMs effectively improves the agreements between the human annotators and the judge LLMs}, as they are encouraged to judge based on the same criteria.

With these four prompts, we evaluate 33 models on our development set and calculate the Pearson correlation on the resulting scores. As shown in Table~\ref{tab:prompt-templates}, the strong correlation between our prompt (Prompt D) and AlpacaEval's prompt (Prompt A) shows that our prompt reasonably aligns with the prompt used in prior work, and the strong correlation between our prompt and the alternative examples (Prompt B and C) shows that our prompt is not overly dependent or biased towards the specific examples that we select. 

\begin{table}[t!]
    \centering  \small
    \setlength{\tabcolsep}{4.5pt}
    \begin{tabular}{ccccc}
\toprule
 & \text{Guideline} & \text{Examples} & \text{Human Agreement Rate} & \text{r with Prompt D} \\
\midrule
    A & \ours\ & \ours\ & \textbf{68.4}\% & -- \\
    B & \ours\ & AlpacaEval & 66.2\% & 0.98 \\
    C & \ours\ & None &  65.6\% & 0.98 \\
    D &  AlpacaEval & None & 66.3\% & 0.95  \\
\bottomrule
    \end{tabular}\vspace{-.1em}
    \caption{
    \textbf{Prompt Template Comparision.} A overview and comparison among four prompt templates on their guideline, examples, human agreement rate, and correlation with the prompt we use on the development set (Prompt A). Note that the prompt that AlpacaEval uses for LLM does not contain examples, and we adopt the examples they give to human annotators for Prompt A. 
 }\label{tab:prompt-templates}
\end{table}

\section{Related Work}\label{sec:related}

To evaluate the capability of post-trained LLMs in instruction-following, prior work has constructs benchmarks in several ways.

\vspace{-0.2cm} \paragraph{Instruction Source.}
Prior work have chosen to source instructions from real-world users. ChatbotArena~\citep{chiang2024chatbot} is a benchmark that constantly collects instructions from the online community users by directly prompting for the user's inputs. ArenaHard~\citep{li2024crowdsourced} automatically curates instructions from those collected by Chatbot Area. These benchmarks possess sets of instructions that closely matches human's common interest in terms of instruction categories, but they are also heavily skewed towards OpenQA and Generation as a result. Another widely recognized benchmark is AlpacaEval~\citep{alpaca_eval, dubois2024length}, which is consist of synthetically generated instructions generated using human-written template~\citep{wang2022self}. WildBench~\citep{lin2024wildbench} also collect instructions from the user in the wild. MT-Bench, with task-specific instructions created by human experts, is the most similar to our work, but it is restricted by the small size of the instruction size. Our work have collected instructions covering a wider range of tasks with a much larger evaluation set.

\vspace{-0.2cm} \paragraph{Evaluating Instruction-Following Models.}
When evaluating a LLM' responses to a instruction, prior work either directly grade the response with a score, or perform a pairwise comparison with the response form another LLM~\citep{zheng2023judging}. Chatbot Arena~\citep{chiang2024chatbot} prompts the same user who creates the instruction to also do a pairwise comparison between responses from two models (i.e., selecting the better response), and the benchmark's evaluation results are treated as ground-truth and compared against by several other benchmarks~\citep{alpaca_eval, lin2024wildbench}. However, such evaluation requires extensive human feedback, which is expensive to collect for majority of the benchmarks. LLM-as-a-judge, acting as a proxy for human annotators, has been widely adopted by many benchmarks in both single response grading and pairwise comparison. However, prior work use closed API models, which lacks transparency and consistency in their judgment. Our work is uses LLM-as-a-judge with public models and shows the benefits that brings.

\vspace{-0.2cm} \paragraph{Reference Guided Evaluation}
Comparing text embeddings to a human-written reference answer is widely used in traditional NLP tasks, especially summarization~\citep{zhang2019bertscore, lin2004rouge, papineni2002bleu, banerjee2005meteor}, but it is less clear how to properly utilize the reference answer to evaluation more open-ended instruction-following. AlpacaEval~\citep{alpaca_eval} has found that including model-generated responses in the prompt when using LLM-as-a-Judge is beneficial in following instruction related to math. Our work adopt an combination of comparing text embeddings to human-written responses and using human-written responses with LLM-as-a-Judge depending on the task categories. Additionally, we provide insights about when and how these responses are beneficial.

\vspace{-0.2cm} \paragraph{Risk of Contamination}
When the test data of the prior work are public released, they are at a high risks of being contaminated. They can potentially lose the robustness and credibility in their evaluation when the evaluated LLMs are trained on the test data. To migrate such risk, WildBench~\citep{lin2024wildbench} keeps their test set private and only release a development set. However, another implicit source of potential contamination remains unsolved when prompting the closed API models with the test data either when using them as the baseline model or as judges. Although not by directly training on the test data, LLMs can still gain knowledge about the them through either distillation from closed API models or training on synthetic data generated by these models~\citep{dubey2024llama, wang2022self, zhou2024lima, peng2023instruction, xu2024wizardlm, zhao2024wildchat}. \ours\ migrates such risks by using local public open-weight models for both the baseline model and the judge. 

\section*{Limitations}\label{sec:limitation}\vspace{-0.2cm} \paragraph{Multi-turn Evaluation.}
Multi-turn evaluation is not the focus of work, and \ours\ is only suitable for single-turn instruction following evaluation. We suggest using benchmarks like WildBench for multi-turn evaluation.

\vspace{-0.2cm} \paragraph{Absolute Rating.} Our work focuses solely on improving pairwise evaluation, which requires the use of a baseline model. We recognize that there might be circumstances where an independent absolute score can be useful, and we leave the topic of improving the accuracy of absolute rating of an LLM in instruction-following for future work. 


\section*{Acknowledgments}We acknowledge the National Artificial Intelligence Research Resource (NAIRR) Pilot and Microsoft Azure for contributing to the results in this work.

\bibliography{iclr2025_conference}

\begin{thebibliography}{34}
\providecommand{\natexlab}[1]{#1}
\providecommand{\url}[1]{\texttt{#1}}
\expandafter\ifx\csname urlstyle\endcsname\relax
  \providecommand{\doi}[1]{doi: #1}\else
  \providecommand{\doi}{doi: \begingroup \urlstyle{rm}\Url}\fi

\bibitem[Abdin et~al.(2024)Abdin, Jacobs, Awan, Aneja, Awadallah, Awadalla, Bach, Bahree, Bakhtiari, Behl, et~al.]{abdin2024phi}
Marah Abdin, Sam~Ade Jacobs, Ammar~Ahmad Awan, Jyoti Aneja, Ahmed Awadallah, Hany Awadalla, Nguyen Bach, Amit Bahree, Arash Bakhtiari, Harkirat Behl, et~al.
\newblock Phi-3 technical report: A highly capable language model locally on your phone.
\newblock \emph{arXiv preprint arXiv:2404.14219}, 2024.

\bibitem[Achiam et~al.(2023)Achiam, Adler, Agarwal, Ahmad, Akkaya, Aleman, Almeida, Altenschmidt, Altman, Anadkat, et~al.]{achiam2023gpt}
Josh Achiam, Steven Adler, Sandhini Agarwal, Lama Ahmad, Ilge Akkaya, Florencia~Leoni Aleman, Diogo Almeida, Janko Altenschmidt, Sam Altman, Shyamal Anadkat, et~al.
\newblock Gpt-4 technical report.
\newblock \emph{arXiv preprint arXiv:2303.08774}, 2023.

\bibitem[Bai et~al.(2023)Bai, Bai, Chu, Cui, Dang, Deng, Fan, Ge, Han, Huang, et~al.]{bai2023qwen}
Jinze Bai, Shuai Bai, Yunfei Chu, Zeyu Cui, Kai Dang, Xiaodong Deng, Yang Fan, Wenbin Ge, Yu~Han, Fei Huang, et~al.
\newblock Qwen technical report.
\newblock \emph{arXiv preprint arXiv:2309.16609}, 2023.

\bibitem[Banerjee \& Lavie(2005)Banerjee and Lavie]{banerjee2005meteor}
Satanjeev Banerjee and Alon Lavie.
\newblock Meteor: An automatic metric for mt evaluation with improved correlation with human judgments.
\newblock In \emph{Proceedings of the acl workshop on intrinsic and extrinsic evaluation measures for machine translation and/or summarization}, pp.\  65--72, 2005.

\bibitem[Biderman et~al.(2023)Biderman, Schoelkopf, Anthony, Bradley, O’Brien, Hallahan, Khan, Purohit, Prashanth, Raff, et~al.]{biderman2023pythia}
Stella Biderman, Hailey Schoelkopf, Quentin~Gregory Anthony, Herbie Bradley, Kyle O’Brien, Eric Hallahan, Mohammad~Aflah Khan, Shivanshu Purohit, USVSN~Sai Prashanth, Edward Raff, et~al.
\newblock Pythia: A suite for analyzing large language models across training and scaling.
\newblock In \emph{International Conference on Machine Learning}, pp.\  2397--2430. PMLR, 2023.

\bibitem[Brown(2020)]{brown2020language}
Tom~B Brown.
\newblock Language models are few-shot learners.
\newblock \emph{arXiv preprint arXiv:2005.14165}, 2020.

\bibitem[Chiang et~al.(2023)Chiang, Li, Lin, Sheng, Wu, Zhang, Zheng, Zhuang, Zhuang, Gonzalez, Stoica, and Xing]{vicuna2023}
Wei-Lin Chiang, Zhuohan Li, Zi~Lin, Ying Sheng, Zhanghao Wu, Hao Zhang, Lianmin Zheng, Siyuan Zhuang, Yonghao Zhuang, Joseph~E. Gonzalez, Ion Stoica, and Eric~P. Xing.
\newblock Vicuna: An open-source chatbot impressing gpt-4 with 90\%* chatgpt quality, March 2023.
\newblock URL \url{https://lmsys.org/blog/2023-03-30-vicuna/}.

\bibitem[Chiang et~al.(2024)Chiang, Zheng, Sheng, Angelopoulos, Li, Li, Zhang, Zhu, Jordan, Gonzalez, and Stoica]{chiang2024chatbot}
Wei-Lin Chiang, Lianmin Zheng, Ying Sheng, Anastasios~Nikolas Angelopoulos, Tianle Li, Dacheng Li, Hao Zhang, Banghua Zhu, Michael Jordan, Joseph~E. Gonzalez, and Ion Stoica.
\newblock Chatbot arena: An open platform for evaluating llms by human preference, 2024.

\bibitem[Conover et~al.(2023)Conover, Hayes, Mathur, Xie, Wan, Shah, Ghodsi, Wendell, Zaharia, and Xin]{conover2023free}
Mike Conover, Matt Hayes, Ankit Mathur, Jianwei Xie, Jun Wan, Sam Shah, Ali Ghodsi, Patrick Wendell, Matei Zaharia, and Reynold Xin.
\newblock Free dolly: Introducing the world’s first truly open instruction-tuned llm.
\newblock \emph{Company Blog of Databricks}, 2023.

\bibitem[Dubey et~al.(2024)Dubey, Jauhri, Pandey, Kadian, Al-Dahle, Letman, Mathur, Schelten, Yang, Fan, et~al.]{dubey2024llama}
Abhimanyu Dubey, Abhinav Jauhri, Abhinav Pandey, Abhishek Kadian, Ahmad Al-Dahle, Aiesha Letman, Akhil Mathur, Alan Schelten, Amy Yang, Angela Fan, et~al.
\newblock The llama 3 herd of models.
\newblock \emph{arXiv preprint arXiv:2407.21783}, 2024.

\bibitem[Dubois et~al.(2024)Dubois, Galambosi, Liang, and Hashimoto]{dubois2024length}
Yann Dubois, Bal{\'a}zs Galambosi, Percy Liang, and Tatsunori~B Hashimoto.
\newblock Length-controlled alpacaeval: A simple way to debias automatic evaluators.
\newblock \emph{arXiv preprint arXiv:2404.04475}, 2024.

\bibitem[First(2014)]{Prolific_2014}
Prolific First.
\newblock Prolific first, 2014.
\newblock URL \url{https://www.prolific.com/}.

\bibitem[Geng et~al.(2023)Geng, Gudibande, Liu, Wallace, Abbeel, Levine, and Song]{koala_blogpost_2023}
Xinyang Geng, Arnav Gudibande, Hao Liu, Eric Wallace, Pieter Abbeel, Sergey Levine, and Dawn Song.
\newblock Koala: A dialogue model for academic research.
\newblock Blog post, April 2023.
\newblock URL \url{https://bair.berkeley.edu/blog/2023/04/03/koala/}.

\bibitem[Groeneveld et~al.(2024)Groeneveld, Beltagy, Walsh, Bhagia, Kinney, Tafjord, Jha, Ivison, Magnusson, Wang, et~al.]{groeneveld2024olmo}
Dirk Groeneveld, Iz~Beltagy, Pete Walsh, Akshita Bhagia, Rodney Kinney, Oyvind Tafjord, Ananya~Harsh Jha, Hamish Ivison, Ian Magnusson, Yizhong Wang, et~al.
\newblock Olmo: Accelerating the science of language models.
\newblock \emph{arXiv preprint arXiv:2402.00838}, 2024.

\bibitem[Jiang et~al.(2023)Jiang, Sablayrolles, Mensch, Bamford, Chaplot, Casas, Bressand, Lengyel, Lample, Saulnier, et~al.]{jiang2023mistral}
Albert~Q Jiang, Alexandre Sablayrolles, Arthur Mensch, Chris Bamford, Devendra~Singh Chaplot, Diego de~las Casas, Florian Bressand, Gianna Lengyel, Guillaume Lample, Lucile Saulnier, et~al.
\newblock Mistral 7b.
\newblock \emph{arXiv preprint arXiv:2310.06825}, 2023.

\bibitem[Lambert et~al.(2024)Lambert, Morrison, Pyatkin, Huang, Ivison, Brahman, Miranda, Liu, Dziri, Lyu, et~al.]{lambert2024t}
Nathan Lambert, Jacob Morrison, Valentina Pyatkin, Shengyi Huang, Hamish Ivison, Faeze Brahman, Lester James~V Miranda, Alisa Liu, Nouha Dziri, Shane Lyu, et~al.
\newblock T$\backslash$" ulu 3: Pushing frontiers in open language model post-training.
\newblock \emph{arXiv preprint arXiv:2411.15124}, 2024.

\bibitem[Li et~al.(2024)Li, Chiang, Frick, Dunlap, Wu, Zhu, Gonzalez, and Stoica]{li2024crowdsourced}
Tianle Li, Wei-Lin Chiang, Evan Frick, Lisa Dunlap, Tianhao Wu, Banghua Zhu, Joseph~E Gonzalez, and Ion Stoica.
\newblock From crowdsourced data to high-quality benchmarks: Arena-hard and benchbuilder pipeline.
\newblock \emph{arXiv preprint arXiv:2406.11939}, 2024.

\bibitem[Li et~al.(2023)Li, Zhang, Dubois, Taori, Gulrajani, Guestrin, Liang, and Hashimoto]{alpaca_eval}
Xuechen Li, Tianyi Zhang, Yann Dubois, Rohan Taori, Ishaan Gulrajani, Carlos Guestrin, Percy Liang, and Tatsunori~B. Hashimoto.
\newblock Alpacaeval: An automatic evaluator of instruction-following models.
\newblock \url{https://github.com/tatsu-lab/alpaca_eval}, 5 2023.

\bibitem[Lin et~al.(2024)Lin, Deng, Chandu, Brahman, Ravichander, Pyatkin, Dziri, Bras, and Choi]{lin2024wildbench}
Bill~Yuchen Lin, Yuntian Deng, Khyathi Chandu, Faeze Brahman, Abhilasha Ravichander, Valentina Pyatkin, Nouha Dziri, Ronan~Le Bras, and Yejin Choi.
\newblock Wildbench: Benchmarking llms with challenging tasks from real users in the wild.
\newblock \emph{arXiv preprint arXiv:2406.04770}, 2024.

\bibitem[Lin(2004)]{lin2004rouge}
Chin-Yew Lin.
\newblock Rouge: A package for automatic evaluation of summaries.
\newblock In \emph{Text summarization branches out}, pp.\  74--81, 2004.

\bibitem[Liu et~al.(2019)Liu, Ott, Goyal, Du, Joshi, Chen, Levy, Lewis, Zettlemoyer, and Stoyanov]{DBLP:journals/corr/abs-1907-11692}
Yinhan Liu, Myle Ott, Naman Goyal, Jingfei Du, Mandar Joshi, Danqi Chen, Omer Levy, Mike Lewis, Luke Zettlemoyer, and Veselin Stoyanov.
\newblock Roberta: {A} robustly optimized {BERT} pretraining approach.
\newblock \emph{CoRR}, abs/1907.11692, 2019.
\newblock URL \url{http://arxiv.org/abs/1907.11692}.

\bibitem[Miranda et~al.(2024)Miranda, Wang, Elazar, Kumar, Pyatkin, Brahman, Smith, Hajishirzi, and Dasigi]{miranda2024hybrid}
Lester James~V Miranda, Yizhong Wang, Yanai Elazar, Sachin Kumar, Valentina Pyatkin, Faeze Brahman, Noah~A Smith, Hannaneh Hajishirzi, and Pradeep Dasigi.
\newblock Hybrid preferences: Learning to route instances for human vs. ai feedback.
\newblock \emph{arXiv preprint arXiv:2410.19133}, 2024.

\bibitem[Ouyang et~al.(2022)Ouyang, Wu, Jiang, Almeida, Wainwright, Mishkin, Zhang, Agarwal, Slama, Ray, et~al.]{ouyang2022training}
Long Ouyang, Jeffrey Wu, Xu~Jiang, Diogo Almeida, Carroll Wainwright, Pamela Mishkin, Chong Zhang, Sandhini Agarwal, Katarina Slama, Alex Ray, et~al.
\newblock Training language models to follow instructions with human feedback.
\newblock In \emph{NeurIPS}, 2022.

\bibitem[Papineni et~al.(2002)Papineni, Roukos, Ward, and Zhu]{papineni2002bleu}
Kishore Papineni, Salim Roukos, Todd Ward, and Wei-Jing Zhu.
\newblock Bleu: a method for automatic evaluation of machine translation.
\newblock In \emph{Proceedings of the 40th annual meeting of the Association for Computational Linguistics}, pp.\  311--318, 2002.

\bibitem[Peng et~al.(2023)Peng, Li, He, Galley, and Gao]{peng2023instruction}
Baolin Peng, Chunyuan Li, Pengcheng He, Michel Galley, and Jianfeng Gao.
\newblock Instruction tuning with gpt-4.
\newblock \emph{arXiv preprint arXiv:2304.03277}, 2023.

\bibitem[Rajani et~al.(2023)Rajani, Tunstall, Beeching, Lambert, Rush, and Wolf]{no_robots}
Nazneen Rajani, Lewis Tunstall, Edward Beeching, Nathan Lambert, Alexander~M. Rush, and Thomas Wolf.
\newblock No robots.
\newblock \url{https://huggingface.co/datasets/HuggingFaceH4/no_robots}, 2023.

\bibitem[Touvron et~al.(2023)Touvron, Martin, Stone, Albert, Almahairi, Babaei, Bashlykov, Batra, Bhargava, Bhosale, et~al.]{touvron2023llama}
Hugo Touvron, Louis Martin, Kevin Stone, Peter Albert, Amjad Almahairi, Yasmine Babaei, Nikolay Bashlykov, Soumya Batra, Prajjwal Bhargava, Shruti Bhosale, et~al.
\newblock Llama 2: Open foundation and fine-tuned chat models.
\newblock \emph{arXiv preprint arXiv:2307.09288}, 2023.

\bibitem[Wang et~al.(2022)Wang, Kordi, Mishra, Liu, Smith, Khashabi, and Hajishirzi]{wang2022self}
Yizhong Wang, Yeganeh Kordi, Swaroop Mishra, Alisa Liu, Noah~A Smith, Daniel Khashabi, and Hannaneh Hajishirzi.
\newblock Self-instruct: Aligning language models with self-generated instructions.
\newblock \emph{arXiv preprint arXiv:2212.10560}, 2022.

\bibitem[Xu et~al.(2024)Xu, Sun, Zheng, Geng, Zhao, Feng, Tao, Lin, and Jiang]{xu2024wizardlm}
Can Xu, Qingfeng Sun, Kai Zheng, Xiubo Geng, Pu~Zhao, Jiazhan Feng, Chongyang Tao, Qingwei Lin, and Daxin Jiang.
\newblock Wizard{LM}: Empowering large pre-trained language models to follow complex instructions.
\newblock In \emph{The Twelfth International Conference on Learning Representations}, 2024.
\newblock URL \url{https://openreview.net/forum?id=CfXh93NDgH}.

\bibitem[Young et~al.(2024)Young, Chen, Li, Huang, Zhang, Zhang, Li, Zhu, Chen, Chang, et~al.]{young2024yi}
Alex Young, Bei Chen, Chao Li, Chengen Huang, Ge~Zhang, Guanwei Zhang, Heng Li, Jiangcheng Zhu, Jianqun Chen, Jing Chang, et~al.
\newblock Yi: Open foundation models by 01. ai.
\newblock \emph{arXiv preprint arXiv:2403.04652}, 2024.

\bibitem[Zhang et~al.(2019)Zhang, Kishore, Wu, Weinberger, and Artzi]{zhang2019bertscore}
Tianyi Zhang, Varsha Kishore, Felix Wu, Kilian~Q Weinberger, and Yoav Artzi.
\newblock Bertscore: Evaluating text generation with bert.
\newblock \emph{arXiv preprint arXiv:1904.09675}, 2019.

\bibitem[Zhao et~al.(2024)Zhao, Ren, Hessel, Cardie, Choi, and Deng]{zhao2024wildchat}
Wenting Zhao, Xiang Ren, Jack Hessel, Claire Cardie, Yejin Choi, and Yuntian Deng.
\newblock Wildchat: 1m chatgpt interaction logs in the wild.
\newblock \emph{arXiv preprint arXiv:2405.01470}, 2024.

\bibitem[Zheng et~al.(2023)Zheng, Chiang, Sheng, Zhuang, Wu, Zhuang, Lin, Li, Li, Xing, et~al.]{zheng2023judging}
Lianmin Zheng, Wei-Lin Chiang, Ying Sheng, Siyuan Zhuang, Zhanghao Wu, Yonghao Zhuang, Zi~Lin, Zhuohan Li, Dacheng Li, Eric Xing, et~al.
\newblock Judging llm-as-a-judge with mt-bench and chatbot arena.
\newblock \emph{Advances in Neural Information Processing Systems}, 36:\penalty0 46595--46623, 2023.

\bibitem[Zhou et~al.(2024)Zhou, Liu, Xu, Iyer, Sun, Mao, Ma, Efrat, Yu, Yu, et~al.]{zhou2024lima}
Chunting Zhou, Pengfei Liu, Puxin Xu, Srinivasan Iyer, Jiao Sun, Yuning Mao, Xuezhe Ma, Avia Efrat, Ping Yu, Lili Yu, et~al.
\newblock Lima: Less is more for alignment.
\newblock \emph{Advances in Neural Information Processing Systems}, 36, 2024.

\end{thebibliography}
\bibliographystyle{iclr2025_conference}

\newpage
\appendix

\begin{table*}[t!]
    \centering  \small
    \setlength{\tabcolsep}{4.5pt}
    \begin{tabular}{c p{0.30\linewidth} | c c c c c c c c | c}
    \toprule
        \multirow{2}{*}{Rank} &
        \multirow{2}{*}{Model} & \multicolumn{8}{c}{General Category}  & \multirow{2}{*}{All} \\
        \cmidrule(lr){3-10} &
         & Brn & OQA & CQA & Ext & Gen & Rew & Sum & Cls &  \\
    \midrule
        1 & Mistral-Large-Instruct-2407 & \mistrallargeinstructceahbd & \mistrallargeinstructceahoqd & \mistrallargeinstructceahcqd & \mistrallargeinstructceahed & \mistrallargeinstructceahgd & \mistrallargeinstructceahrd & \mistrallargeinstructceahsd & \mistrallargeinstructceahcd & \mistrallargeinstructceahod \\
        1 & gpt-4-turbo-2024-04-09 & \gpteturbocaceaeajbd & \gpteturbocaceaeajoqd & \gpteturbocaceaeajcqd & \gpteturbocaceaeajed & \gpteturbocaceaeajgd & \gpteturbocaceaeajrd & \gpteturbocaceaeajsd & \gpteturbocaceaeajcd & \gpteturbocaceaeajod \\
        1 & gpt-4o-2024-05-13 & \gpteocaceafbdbd & \gpteocaceafbdoqd & \gpteocaceafbdcqd & \gpteocaceafbded & \gpteocaceafbdgd & \gpteocaceafbdrd & \gpteocaceafbdsd & \gpteocaceafbdcd & \gpteocaceafbdod \\
        1 & gpt-4-1106-preview & \gptebbagpreviewbd & \gptebbagpreviewoqd & \gptebbagpreviewcqd & \gptebbagpreviewed & \gptebbagpreviewgd & \gptebbagpreviewrd & \gptebbagpreviewsd & \gptebbagpreviewcd & \gptebbagpreviewod \\
        5 & Llama-3.1-70B-Instruct & \llamadbhabinstructbd & \llamadbhabinstructoqd & \llamadbhabinstructcqd & \llamadbhabinstructed & \llamadbhabinstructgd & \llamadbhabinstructrd & \llamadbhabinstructsd & \llamadbhabinstructcd & \llamadbhabinstructod \\
        6 & Qwen2.5-72B-Instruct & \qwencfhcbinstructbd & \qwencfhcbinstructoqd & \qwencfhcbinstructcqd & \qwencfhcbinstructed & \qwencfhcbinstructgd & \qwencfhcbinstructrd & \qwencfhcbinstructsd & \qwencfhcbinstructcd & \qwencfhcbinstructod \\
        6 & Mistral-Small-Instruct-2409 & \mistralsmallinstructceajbd & \mistralsmallinstructceajoqd & \mistralsmallinstructceajcqd & \mistralsmallinstructceajed & \mistralsmallinstructceajgd & \mistralsmallinstructceajrd & \mistralsmallinstructceajsd & \mistralsmallinstructceajcd & \mistralsmallinstructceajod \\
        8 & Qwen1.5-110B-Chat & \qwenbfbbabchatbd & \qwenbfbbabchatoqd & \qwenbfbbabchatcqd & \qwenbfbbabchated & \qwenbfbbabchatgd & \qwenbfbbabchatrd & \qwenbfbbabchatsd & \qwenbfbbabchatcd & \qwenbfbbabchatod \\
        9 & Llama-3.1-8B-Instruct & \llamadbibinstructbd & \llamadbibinstructoqd & \llamadbibinstructcqd & \llamadbibinstructed & \llamadbibinstructgd & \llamadbibinstructrd & \llamadbibinstructsd & \llamadbibinstructcd & \llamadbibinstructod \\
        9 & Qwen2-72B-Instruct & \qwenchcbinstructbd & \qwenchcbinstructoqd & \qwenchcbinstructcqd & \qwenchcbinstructed & \qwenchcbinstructgd & \qwenchcbinstructrd & \qwenchcbinstructsd & \qwenchcbinstructcd & \qwenchcbinstructod \\
        9 & Yi-1.5-34B-Chat & \yibfdebchatbd & \yibfdebchatoqd & \yibfdebchatcqd & \yibfdebchated & \yibfdebchatgd & \yibfdebchatrd & \yibfdebchatsd & \yibfdebchatcd & \yibfdebchatod \\
        9 & Phi-3-medium-4k-instruct & \phidmediumekinstructbd & \phidmediumekinstructoqd & \phidmediumekinstructcqd & \phidmediumekinstructed & \phidmediumekinstructgd & \phidmediumekinstructrd & \phidmediumekinstructsd & \phidmediumekinstructcd & \phidmediumekinstructod \\
        13 & gpt-3.5-turbo & \gptdfturbobd & \gptdfturbooqd & \gptdfturbocqd & \gptdfturboed & \gptdfturbogd & \gptdfturbord & \gptdfturbosd & \gptdfturbocd & \gptdfturbood \\
        13 & tulu-2-dpo-70b & \tulucdpohabbd & \tulucdpohaboqd & \tulucdpohabcqd & \tulucdpohabed & \tulucdpohabgd & \tulucdpohabrd & \tulucdpohabsd & \tulucdpohabcd & \tulucdpohabod \\
        13 & Mistral-7B-Instruct-v0.3 & \mistralhbinstructvadbd & \mistralhbinstructvadoqd & \mistralhbinstructvadcqd & \mistralhbinstructvaded & \mistralhbinstructvadgd & \mistralhbinstructvadrd & \mistralhbinstructvadsd & \mistralhbinstructvadcd & \mistralhbinstructvadod \\
        16 & Llama-2-70b-chat-hf & \llamachabchathfbd & \llamachabchathfoqd & \llamachabchathfcqd & \llamachabchathfed & \llamachabchathfgd & \llamachabchathfrd & \llamachabchathfsd & \llamachabchathfcd & \llamachabchathfod \\
        16 & WizardLM-13B-V1.2 & \wizardlmbdbvbcbd & \wizardlmbdbvbcoqd & \wizardlmbdbvbccqd & \wizardlmbdbvbced & \wizardlmbdbvbcgd & \wizardlmbdbvbcrd & \wizardlmbdbvbcsd & \wizardlmbdbvbccd & \wizardlmbdbvbcod \\
        16 & tulu-v2.5-ppo-13b & \tuluvcfppobdbufmeanhabufrmbd & \tuluvcfppobdbufmeanhabufrmoqd & \tuluvcfppobdbufmeanhabufrmcqd & \tuluvcfppobdbufmeanhabufrmed & \tuluvcfppobdbufmeanhabufrmgd & \tuluvcfppobdbufmeanhabufrmrd & \tuluvcfppobdbufmeanhabufrmsd & \tuluvcfppobdbufmeanhabufrmcd & \tuluvcfppobdbufmeanhabufrmod \\
        16 & vicuna-13b-v1.5 & \vicunabdbvbfbd & \vicunabdbvbfoqd & \vicunabdbvbfcqd & \vicunabdbvbfed & \vicunabdbvbfgd & \vicunabdbvbfrd & \vicunabdbvbfsd & \vicunabdbvbfcd & \vicunabdbvbfod \\
        16 & tulu-2-dpo-13b & \tulucdpobdbbd & \tulucdpobdboqd & \tulucdpobdbcqd & \tulucdpobdbed & \tulucdpobdbgd & \tulucdpobdbrd & \tulucdpobdbsd & \tulucdpobdbcd & \tulucdpobdbod \\
        21 & Llama-2-13b-chat-hf & \llamacbdbchathfbd & \llamacbdbchathfoqd & \llamacbdbchathfcqd & \llamacbdbchathfed & \llamacbdbchathfgd & \llamacbdbchathfrd & \llamacbdbchathfsd & \llamacbdbchathfcd & \llamacbdbchathfod \\
        21 & Llama-2-7b-chat-hf & \llamachbchathfbd & \llamachbchathfoqd & \llamachbchathfcqd & \llamachbchathfed & \llamachbchathfgd & \llamachbchathfrd & \llamachbchathfsd & \llamachbchathfcd & \llamachbchathfod \\
        21 & tulu-2-dpo-7b & \tulucdpohbbd & \tulucdpohboqd & \tulucdpohbcqd & \tulucdpohbed & \tulucdpohbgd & \tulucdpohbrd & \tulucdpohbsd & \tulucdpohbcd & \tulucdpohbod \\
        21 & vicuna-7b-v1.5 & \vicunahbvbfbd & \vicunahbvbfoqd & \vicunahbvbfcqd & \vicunahbvbfed & \vicunahbvbfgd & \vicunahbvbfrd & \vicunahbvbfsd & \vicunahbvbfcd & \vicunahbvbfod \\
        21 & gpt4all-13b-snoozy & \gpteallbdbsnoozybd & \gpteallbdbsnoozyoqd & \gpteallbdbsnoozycqd & \gpteallbdbsnoozyed & \gpteallbdbsnoozygd & \gpteallbdbsnoozyrd & \gpteallbdbsnoozysd & \gpteallbdbsnoozycd & \gpteallbdbsnoozyod \\
        26 & OLMo-7B-SFT-hf & \olmohbsfthfbd & \olmohbsfthfoqd & \olmohbsfthfcqd & \olmohbsfthfed & \olmohbsfthfgd & \olmohbsfthfrd & \olmohbsfthfsd & \olmohbsfthfcd & \olmohbsfthfod \\
        26 & OLMo-7B-0724-Instruct-hf & \olmohbahceinstructhfbd & \olmohbahceinstructhfoqd & \olmohbahceinstructhfcqd & \olmohbahceinstructhfed & \olmohbahceinstructhfgd & \olmohbahceinstructhfrd & \olmohbahceinstructhfsd & \olmohbahceinstructhfcd & \olmohbahceinstructhfod \\
        26 & dolly-v2-7b & \dollyvchbbd & \dollyvchboqd & \dollyvchbcqd & \dollyvchbed & \dollyvchbgd & \dollyvchbrd & \dollyvchbsd & \dollyvchbcd & \dollyvchbod \\
        26 & koala-13B-HF & \koalabdbhfbd & \koalabdbhfoqd & \koalabdbhfcqd & \koalabdbhfed & \koalabdbhfgd & \koalabdbhfrd & \koalabdbhfsd & \koalabdbhfcd & \koalabdbhfod \\
        26 & mpt-7b-chat & \mpthbchatbd & \mpthbchatoqd & \mpthbchatcqd & \mpthbchated & \mpthbchatgd & \mpthbchatrd & \mpthbchatsd & \mpthbchatcd & \mpthbchatod \\
        26 & dolly-v2-12b & \dollyvcbcbbd & \dollyvcbcboqd & \dollyvcbcbcqd & \dollyvcbcbed & \dollyvcbcbgd & \dollyvcbcbrd & \dollyvcbcbsd & \dollyvcbcbcd & \dollyvcbcbod \\
        32 & koala-7B-HF & \koalahbhfbd & \koalahbhfoqd & \koalahbhfcqd & \koalahbhfed & \koalahbhfgd & \koalahbhfrd & \koalahbhfsd & \koalahbhfcd & \koalahbhfod \\
        32 & oasst-sft-1-pythia-12b & \oasstsftbpythiabcbbd & \oasstsftbpythiabcboqd & \oasstsftbpythiabcbcqd & \oasstsftbpythiabcbed & \oasstsftbpythiabcbgd & \oasstsftbpythiabcbrd & \oasstsftbpythiabcbsd & \oasstsftbpythiabcbcd & \oasstsftbpythiabcbod \\
    \bottomrule
    \end{tabular}\vspace{-.1em}
    \caption{
    \textbf{\winrate s of all 33 starting models evaluated on the validation set of \ours.} All numbers are in \%. ($i$) indicates the ranking. Brn $\rightarrow$ Brainstorm; OQA $\rightarrow$ Open QA; CQA $\rightarrow$ Closed QA; Ext $\rightarrow$ Extraction; Gen $\rightarrow$ Generation; Rew $\rightarrow$ Rewriting; Sum $\rightarrow$ Summarization; Cls $\rightarrow$ Classification; FC $\rightarrow$ Fact Checking / Attributed QA; MDS $\rightarrow$ Multi-Document Synthesis; RND $\rightarrow$ Reasoning Over Numerical Data; 
 }\label{tab:dev-results-full}
\end{table*}

\section{Full Validation Set Results}\label{app:full-eval-results}
See the \winrate\ of all of the starting 29 models evaluated on validation set of \ours\ in Table~\ref{tab:dev-results-full}. 

\begin{table*}[t!]
    \centering
    \small
    \setlength{\tabcolsep}{4.5pt}
    \begin{tabular}{l | l}
    \toprule
        Model Family & Model Name \\
    \midrule
        \multirow{2}{*}{Dolly} & dolly-v2-12b \\
        & dolly-v2-7b \\
    \midrule
        \multirow{2}{*}{Koala} & koala-7B-HF \\
        & koala-13B-HF \\
    \midrule
        \multirow{3}{*}{Llama-2} & Llama-2-7b-chat-hf \\
        & Llama-2-13b-chat-hf \\
        & Llama-2-70b-chat-hf \\
    \midrule
        \multirow{4}{*}{Llama-3} & Meta-Llama-3-8B-Instruct \\
        & Meta-Llama-3.1-8B-Instruct \\
        & Meta-Llama-3-70B-Instruct \\
        & Meta-Llama-3.1-70B-Instruct \\
    \midrule
        \multirow{3}{*}{Mistral} & Mistral-7B-Instruct-v0.3 \\
        & Mistral-Small-Instruct-2409 \\
        & Mistral-Large-Instruct-2407 \\
    \midrule
        MPT & mpt-7b-chat  \\
    \midrule
        OpenAssistant & oasst-sft-1-pythia-12b  \\
    \midrule
        \multirow{2}{*}{OLMo} & OLMo-7B-SFT-hf \\
        & OLMo-7B-0724-Instruct-hf \\
    \midrule
        Phi & Phi-3-medium-4k-instruct  \\
    \midrule
        \multirow{2}{*}{Qwen} & Qwen2-72B-Instruct \\
        & Qwen1.5-110B-Chat \\
    \midrule
        \multirow{2}{*}{Vicuna} & vicuna-7b-v1.5 \\
        & vicuna-13b-v1.5 \\
    \midrule
        WizardLM & WizardLM-13B-V1.2  \\
    \midrule
        Yi & Yi-1.5-34B-Chat  \\
    \midrule
        GPT-3 & gpt-3.5-turbo  \\
    \midrule
        \multirow{3}{*}{GPT-4} & gpt-4-1106-preview \\
        & gpt-4-turbo-2024-04-09 \\
        & gpt-4o-2024-05-13 \\
    \midrule
        \multirow{2}{*}{O-1} & o1-mini \\
        & o1-preview \\
    \bottomrule
    \end{tabular}\vspace{-.1em}
    \caption{
    Full list of model family and names that we use to construct the model pool where we sample the responses for the human agreement set.
 }\label{tab:human-set-model-pool}
\end{table*}

\section{Formulation}\label{app:formulation}

We formally define the research problem and our proposed evaluation method, \ourmethod.

\subsection{Problem Definition}
We denote \ours's evaluation dataset as ${D}$, with each element being $(in, o_\mathcal{B}, o_\mathcal{H})$, denoting the instruction, the baseline model response, and the human written reference response respectively. 

Given a target LLM $\mathcal{T}$, \ours\ aims to estimate the rate that human would consider the responses from $\mathcal{T}$ are at least as good as the baseline model $\mathcal{B}$ in following instructions, which we formally defined as: 
\begin{equation*}
    \winrate(\mathcal{T}, \mathcal{B}) = \frac{1}{|D|} \sum_{(in, o_\mathcal{B}, o_\mathcal{H}) \in D}p(in, o_\mathcal{T}, o_\mathcal{B}, o_\mathcal{H})
\end{equation*}
where $o_{\mathcal{T}} = \mathcal{T}(in)$ represents the response of $\mathcal{T}$ given the instruction as the input, and $p(in, o_\mathcal{T}, o_\mathcal{B}, o_\mathcal{H})$ is a binary function representing the pairwise preference ($0$ if the baseline model is preferred and $1$ otherwise). 

\subsection{LLM-as-a-Judge with Optional Human Reference}
We proposes the evaluation method, LLM-as-a-judge with human reference, as one of the methods to estimates $p(in, o_\mathcal{T}, o_\mathcal{B})$. Specifically, we embed $in, o_\mathcal{T}, o_\mathcal{B}, o_\mathcal{H}$ into a prompt template as the input to a separate judge model $\mathcal{J}$ formally: 
\begin{equation*}
    p(in, o_\mathcal{T}, o_\mathcal{B}, o_\mathcal{H}) = \mathcal{J}(in, o_\mathcal{T}, o_\mathcal{B}, o_\mathcal{H})
\end{equation*}
Note that when not using a reference, the defination is the same except that $o_\mathcal{H}$ will not be an input to $\mathcal{J}$.

\subsection{\ourrb: Comparing Text Embeddings with Human Reference}
We also proposes to compare the cosine similarity between the text embeddings of $o_\mathcal{T}$ and  $o_\mathcal{H}$ against $o_\mathcal{B}$ and  $o_\mathcal{H}$. Formally, 
\begin{equation*}
    p(in, o_\mathcal{T}, o_\mathcal{B}, o_\mathcal{H}) = \begin{cases} 
      0 & \text{if } \text{sim}(o_\mathcal{T}, o_\mathcal{H}) < \text{sim}(o_\mathcal{B}, o_\mathcal{H}) \\
      1 & \text{otherwise}.
   \end{cases}
\end{equation*}
where
\begin{equation*}
    \text{sim}(o_\mathcal{X}, o_\mathcal{Y}) = \frac{\text{Embed}(o_\mathcal{X}) \cdot \text{Embed}(o_\mathcal{Y})}{\|\text{Embed}(o_\mathcal{X})\| \|\text{Embed}(o_\mathcal{Y})\|}
\end{equation*}
with Embed$(o_\mathcal{Y})$ represents some embeddings of $o_\mathcal{Y}$.

\begin{figure}
\centering \footnotesize
\resizebox{\columnwidth}{!}{\includegraphics[trim={1.5cm 0cm 1.5cm 2cm},clip]{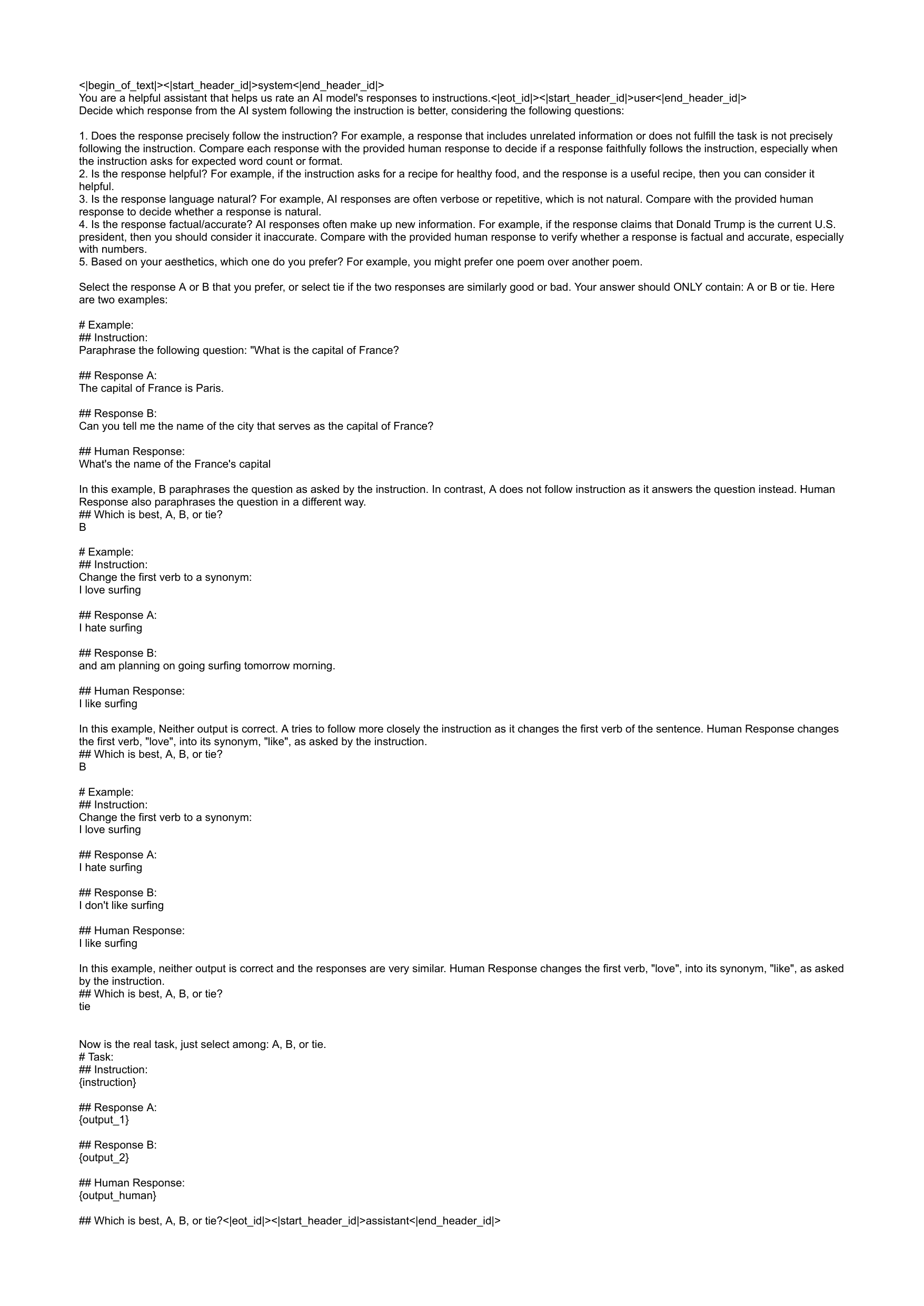}}
\caption{
\textbf{Prompt Template For LLM-as-a-Judge with Human Response. }
The prompt template we use to prompt our judge model Llama-3.1-70B-Instruct to give the preference between two model responses along with human reference. Note that we intentionally transform the guidelines we give to the human annotators into this prompt to maximize the fairness in comparison.
}\label{fig:model-prompt-reference}
\end{figure}

\begin{figure}
\centering \footnotesize
\resizebox{\columnwidth}{!}{\includegraphics[trim={1.5cm 3.5cm 1.5cm 2cm},clip]{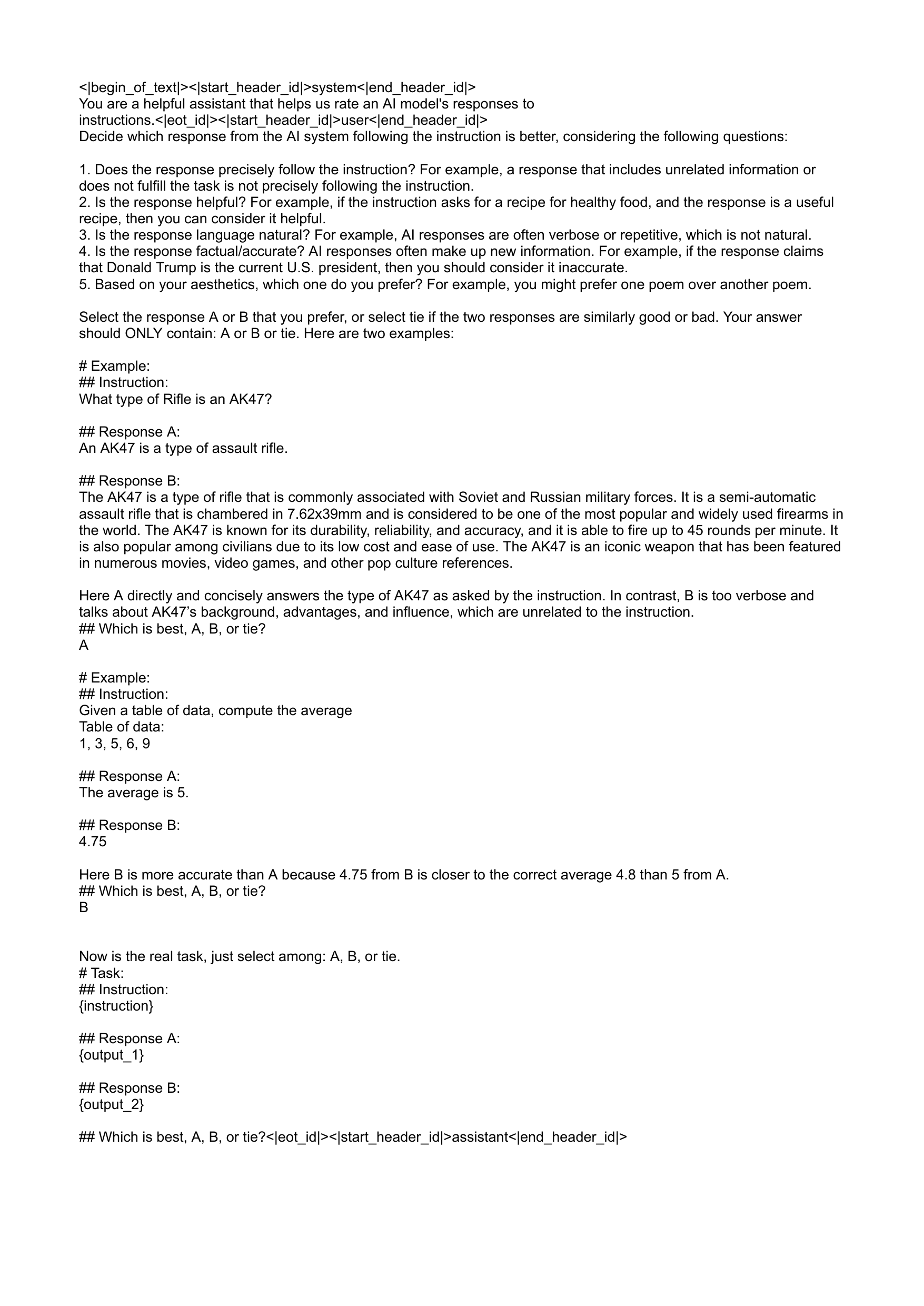}}
\caption{
\textbf{Prompt Template For LLM-as-a-Judge.}
The prompt template we use to prompt our judge model Llama-3.1-70B-Instruct to give the preference between two model responses without a reference. Note that we intentionally transform the guidelines we give to the human annotators into this prompt to maximize the fairness in comparison.
}\label{fig:model-prompt}
\end{figure}

\begin{figure}
\centering \footnotesize
\resizebox{\columnwidth}{!}{\includegraphics[trim={1.5cm 0cm 1.5cm 2cm},clip]{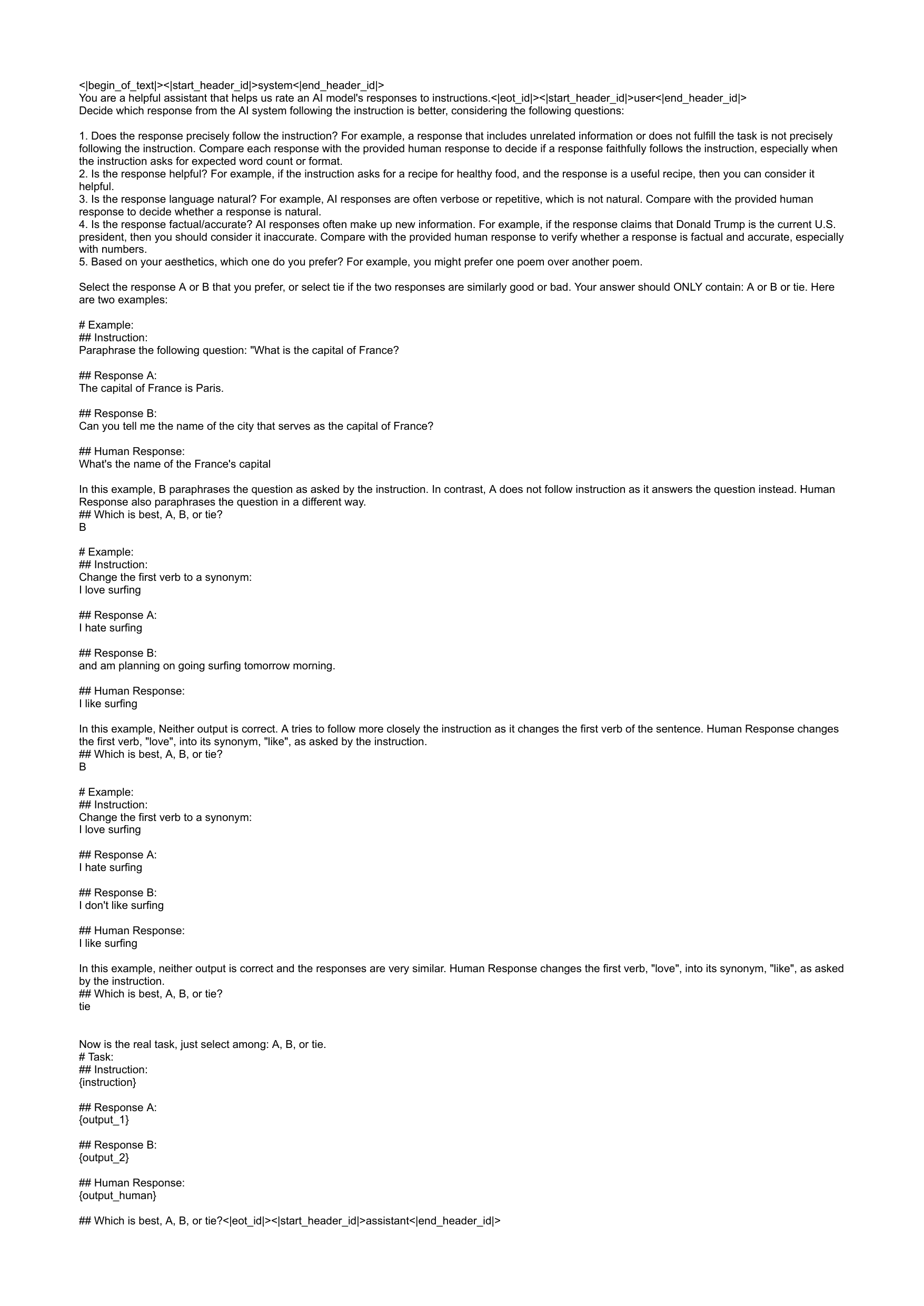}}
\caption{
\textbf{Prompt Template with demonstration examples replaced.}
A modified version of the prompt template we use to prompt our judge model Llama-3.1-70B-Instruct to give the preference between two model responses with a reference. We replace the demonstrations examples with the ones adopted from the examples given to the human annotators by AlpacaEval. 
}\label{fig:model-prompt-alpacaeval-examples}
\end{figure}

\begin{figure}
\centering \footnotesize
\resizebox{\columnwidth}{!}{\includegraphics[trim={1.5cm 18cm 1.5cm 2cm},clip]{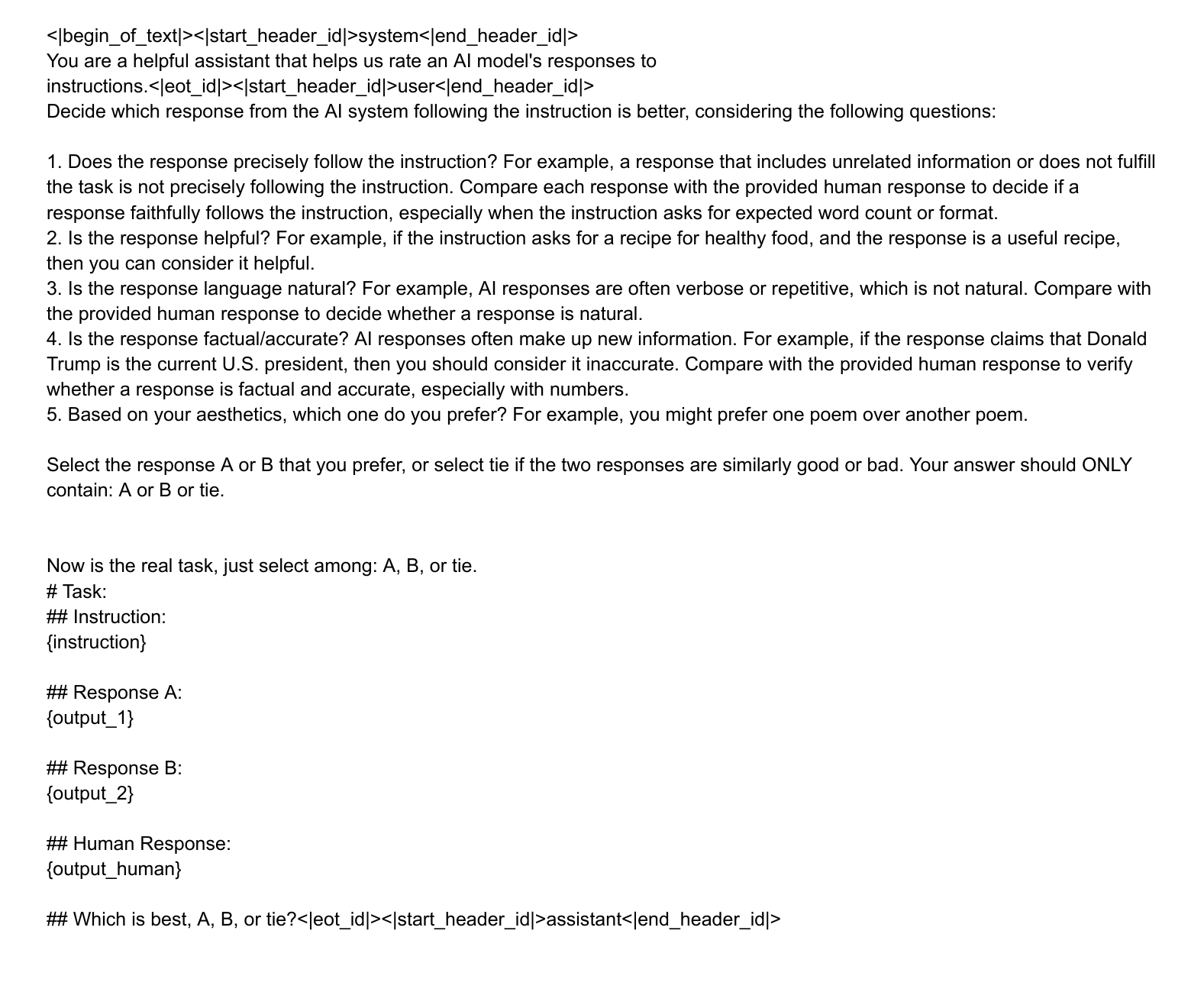}}
\caption{
\textbf{Prompt Template with demonstration examples removed.}
A modified version of the prompt template we use to prompt our judge model Llama-3.1-70B-Instruct to give the preference between two model responses with a reference. We removes the demonstration examples. 
}\label{fig:model-prompt-no-example}
\end{figure}

\begin{figure}
\centering \footnotesize
\resizebox{\columnwidth}{!}{\includegraphics[trim={1.5cm 18cm 1.5cm 2cm},clip]{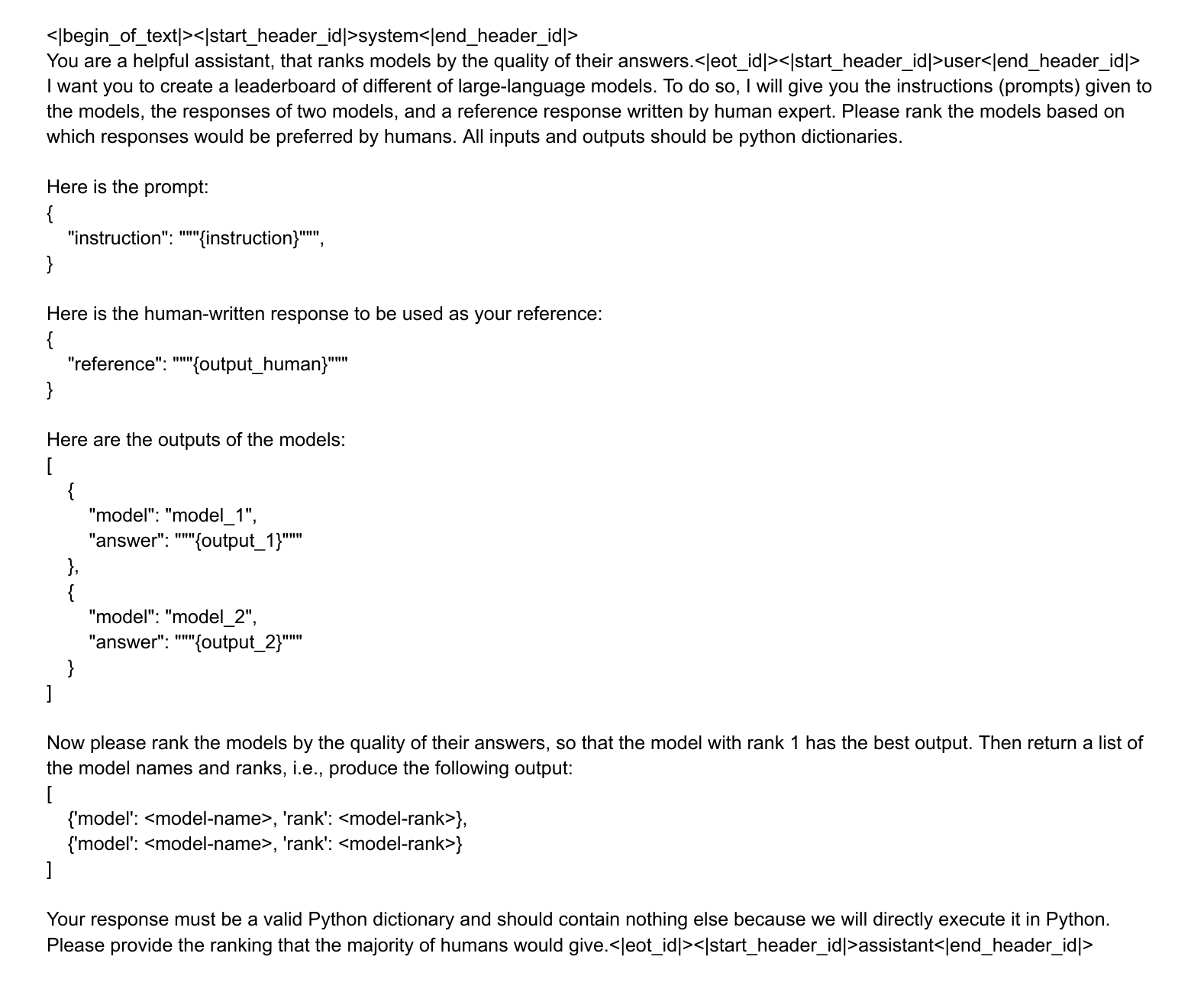}}
\caption{
\textbf{Prompt Template from AlpacaEval.}
A modified version of the prompt template we use to prompt our judge model Llama-3.1-70B-Instruct to give the preference between two model responses with a reference. We adopt the exactly prompt that AlpacaEval uses for their judge LLMs. 
}\label{fig:model-prompt-alpacaeval-all}
\end{figure}

\section{LLM-as-a-Judge Prompt and Parsing }\label{app:llm-template}
Figure~\ref{fig:model-prompt-reference} shows the prompt template for LLM-as-a-Judge where we embed the instruction, the target and reference model responses, and the human written reference into to construct the final prompt for the judge LLM as mentioned in Section~\ref{subsec:methods}. Figures~\ref{fig:model-prompt} shows the one without including human reference. We design the template to match the guideline we give to human annotators in Section~\ref{app:human-guideline}, resulting in a 2-shot prompting. Note that we randomly swap the target and reference model response to avoid potential label bias.

During parsing, we strip and normalized the generated output, and map the exact match of "a" into $0$, and "b" or "tie" into $1$. We optionally reverse the preference if the embedded responses are swapped. Note that when the parsing fails, we ignore the current data point in the calculate of the \winrate s.

Figure~\ref{fig:model-prompt-alpacaeval-examples}, Figure~\ref{fig:model-prompt-no-example}, and Figure~\ref{fig:model-prompt-alpacaeval-all} shows the other modified version of prompt templates that we compare our prompt template against in Section~\ref{subsec:choice-of-prompt}.

\section{Human Agreement Analysis Details}
\subsection{Model Pool}\label{app:human-set-model-pool}
The full model pool from which we sample the responses to construct our human agreement dataset in Section~\ref{subsec:human-agreement-instruction-collection} and Section~\ref{sec:dataset} includes Dolly~\citep{conover2023free}, Koala~\citep{koala_blogpost_2023}, Llama-2~\citep{touvron2023llama}, Llama-3.1~\citep{dubey2024llama}, Mistral~\citep{jiang2023mistral}, MPT~\citep{dubey2024llama}, Pythia~\citep{biderman2023pythia}, OLMo~\citep{groeneveld2024olmo}, Phi~\citep{abdin2024phi}, Qwen~\citep{bai2023qwen}, Vicuna~\citep{vicuna2023}, WizardLM~\citep{xu2024wizardlm}, Yi~\citep{young2024yi}, GPT-3~\citep{brown2020language}, GPT-4~\citep{achiam2023gpt}, and O1\footnote{https://openai.com/o1/}. See Table~\ref{tab:human-set-model-pool} for the full list of model names.

\begin{figure}
\centering \footnotesize
\resizebox{\columnwidth}{!}{\includegraphics[trim={1.5cm 14cm 1.5cm 1.5cm},clip]{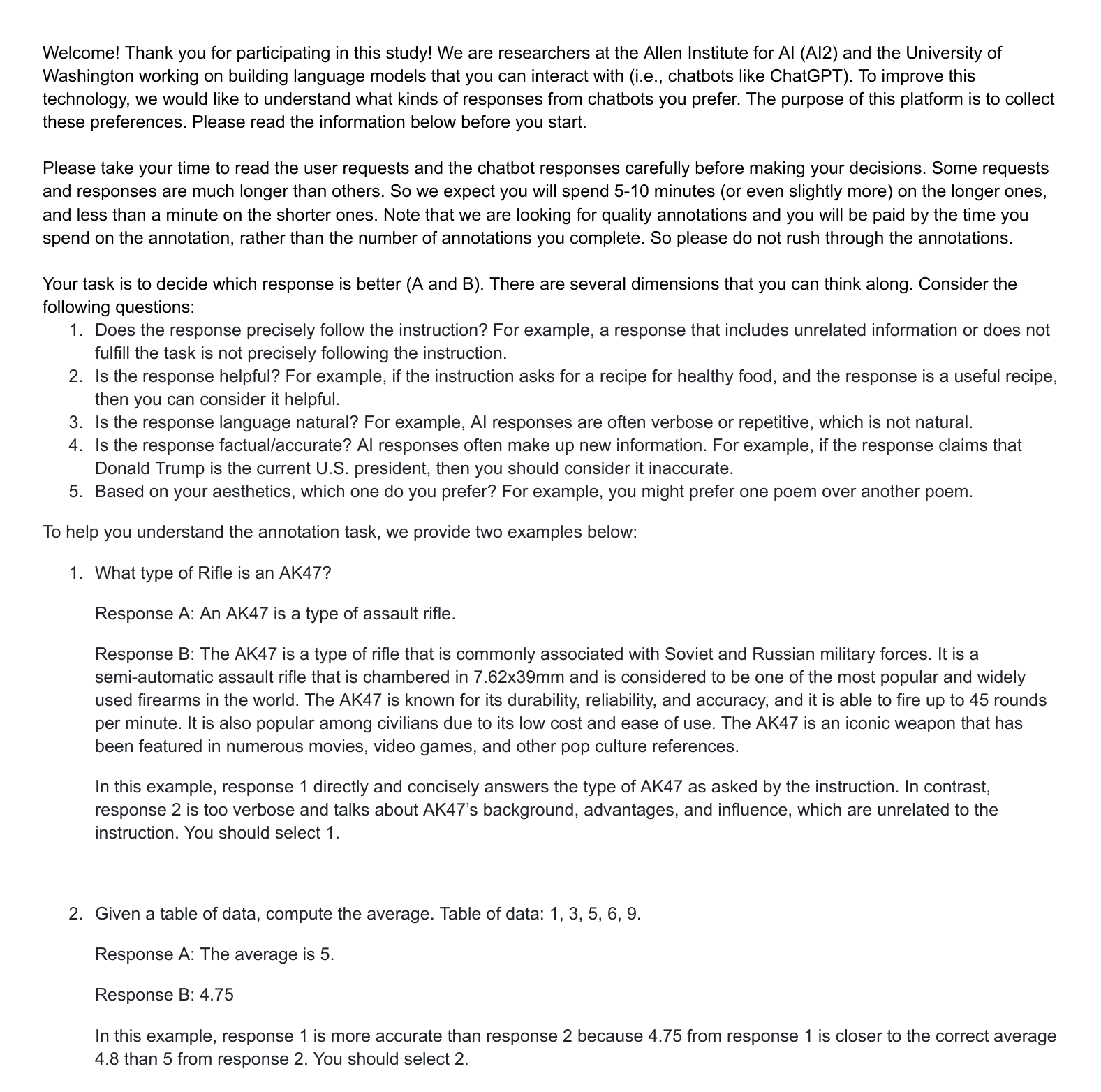}}
\caption{
\textbf{Guideline for Human Annotator. }
The guideline we provide for the human annotators. A modified version from \citet{alpaca_eval}. 
}\label{fig:human-guideline}
\end{figure}

\subsection{Human Annotation Guideline}\label{app:human-guideline}
Figure~\ref{fig:human-guideline} shows the full guideline we provide to the annotators during preference collection. We adopt the guideline from \citet{alpaca_eval} with some modifications.

\begin{figure}
\centering \footnotesize
\resizebox{\columnwidth}{!}{\includegraphics[trim={0 0 0 0},clip]{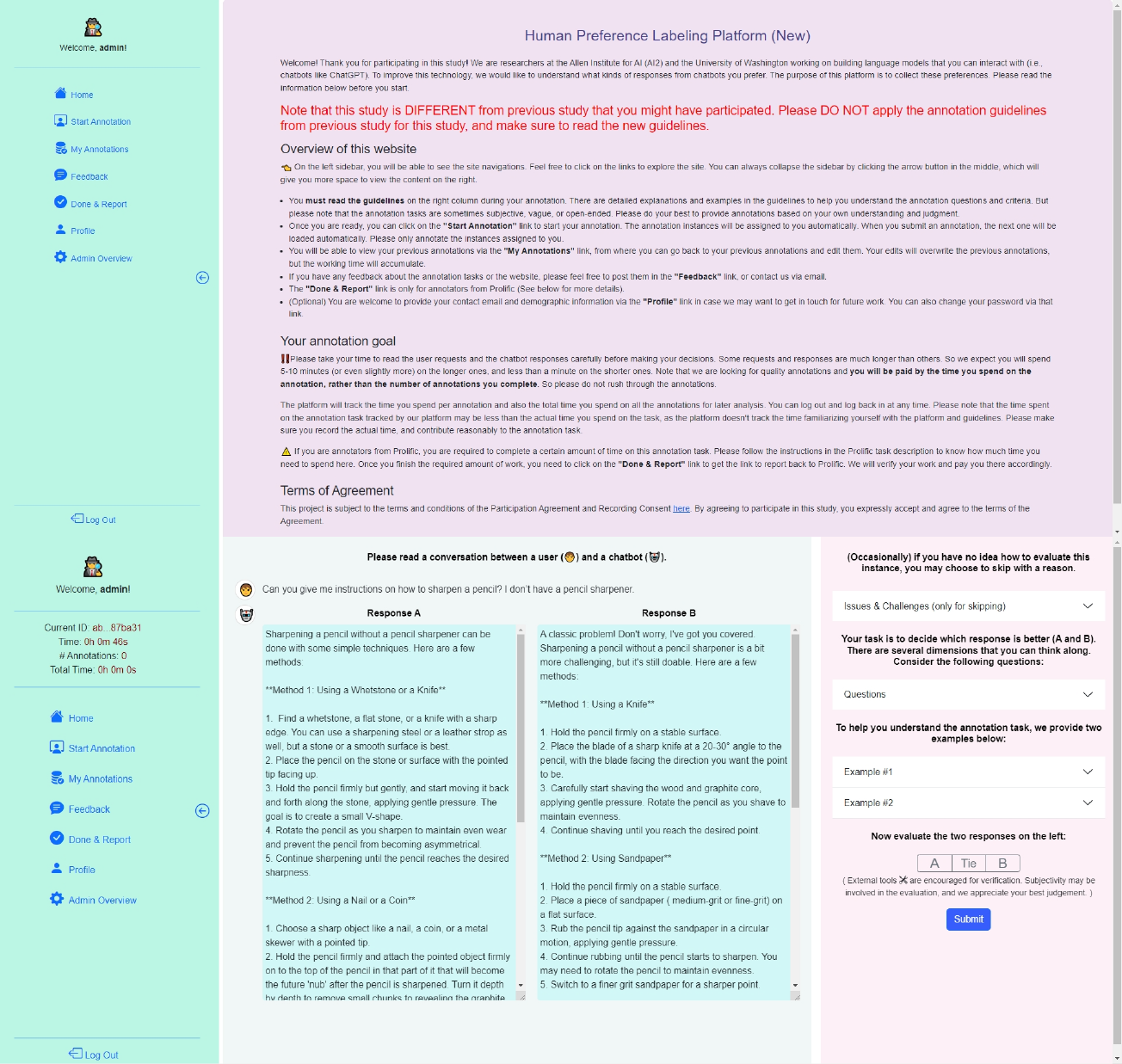}}
\caption{
\textbf{Annotation Website. }
The main pages of the website we build for collecting human annotations. The website framework is adopted from \citet{miranda2024hybrid}. 
}\label{fig:website}
\end{figure}

\subsection{Annotation Website}
See Figure~\ref{fig:website} for an overview of the website that we direct our human annotator to. We ask them to spend time in getting familiar with website before annotations.

\begin{algorithm}
\caption{Algorithms to calculate Leave-One-Out (LOO) agreement rate either within the set of annotation of annotations (inner) or against a evaluator prediction (outer).}\label{alg:loo}
\begin{algorithmic}

\Function{get\_mode}{annotations}
    \State modes $\gets$ list of annotations with highest occurrence frequency 
    \If{length of modes $>$ 1}
        \State \Return randomly chosen annotation from modes
    \Else
        \State \Return modes[0]
    \EndIf
\EndFunction
\newline

\Function{leave\_one\_out\_agreement\_inner}{annotations}
    \State n\_annotations $\gets$ length of annotations
    \State n\_correct\_predictions $\gets$ 0

    \For{each $i$ from 1 to n\_annotations}
        \State target\_annotations $\gets$ annotations without $i$-th element
        \State mode $\gets$ get\_mode(target\_annotations)
        
        \If{annotations[$i$] = mode}
            \State correct\_predictions $\gets$ correct\_predictions + 1
        \EndIf
    \EndFor

    \State \Return n\_correct\_predictions / n\_annotations
\EndFunction
\newline

\Function{leave\_one\_out\_agreement\_outer}{annotations, prediction}
    \State n\_annotations $\gets$ length of annotations
    \State n\_correct\_predictions $\gets$ 0

    \For{each $i$ from 1 to n\_annotations}
        \State target\_annotations $\gets$ annotations without $i$-th element
        \State mode $\gets$ get\_mode(target\_annotations)
        
        \If{prediction = mode}
            \State correct\_predictions $\gets$ correct\_predictions + 1
        \EndIf
    \EndFor

    \State \Return n\_correct\_predictions / n\_annotations
\EndFunction
\end{algorithmic}
\end{algorithm}\

\subsection{Leave-One-Out Agreement Rate Calculation}\label{app:loo}
Algorithm~\ref{alg:loo} provides a detailed overview of the metric Leave-One-Out Agreement Rate used in human agreement analysis (Section~\ref{sec:experimental-setup}).

\end{document}